\documentclass[10pt,journal]{IEEEtran}
\usepackage[noadjust]{cite}
\usepackage[bookmarks=false]{hyperref}
\hypersetup{
    colorlinks,
    linkcolor={blue!20!black},
    citecolor={blue!20!black},
    urlcolor={blue!50!black}
}
\usepackage[font=footnotesize]{subcaption}
\usepackage[font=footnotesize]{caption}
\usepackage{caption}
\usepackage[utf8]{inputenc}
\usepackage{graphicx,url}

\usepackage{booktabs}
\graphicspath{ {./images/} }
\usepackage{multirow}
\usepackage{hhline}
\usepackage{xspace}
\usepackage{graphicx}
\usepackage{amssymb,amsmath,mathtools}
\usepackage{amsthm}
\newtheorem{proposition}{Proposition} %
\newtheorem{remark}{Remark}

\usepackage{bm}                 %
\usepackage{pifont}
\newtheorem{definition}{Definition}

\theoremstyle{plain}

\usepackage{color,soul}
\usepackage[dvipsnames]{xcolor}
\usepackage{bbm}
\usepackage{stmaryrd}
\usepackage{textcomp}
\usepackage{lipsum,lineno}
\usepackage{float}
\makeatletter
\newif\if@restonecol
\makeatother
\usepackage{amsmath}
\usepackage{kbordermatrix}
\usepackage{mathrsfs}
\usepackage[shortlabels]{enumitem}
\newcommand\numeq[1]%
{\stackrel{\scriptscriptstyle(\mkern-1.5mu#1\mkern-1.5mu)}{=}}
\newcommand\numeqq[1]%
{\stackrel{\scriptscriptstyle(\mkern-1.5mu#1\mkern-1.5mu)}{\triangleq}}
\newcommand\numleq[1]%
{\stackrel{\scriptscriptstyle(\mkern-1.5mu#1\mkern-1.5mu)}{\leq}}
\newcommand\numgeq[1]%
{\stackrel{\scriptscriptstyle(\mkern-1.5mu#1\mkern-1.5mu)}{\geq}}
\newcommand\numimp[1]%
{\stackrel{\scriptscriptstyle(\mkern-1.5mu#1\mkern-1.5mu)}{\implies}}
\newcommand\norm[1]{\lVert#1\rVert}

\usepackage[titlenumbered,ruled,linesnumbered]{algorithm2e}
\SetAlCapNameFnt{\footnotesize}
\SetAlCapFnt{\footnotesize}
\usepackage[shortlabels]{enumitem}

\let\oldnl\nl
\newcommand{\nonl}{\renewcommand{\nl}{\let\nl\oldnl}}
\usepackage{algpseudocode}

\usepackage{fancyhdr}
\usepackage{tabularx}
\usepackage{dsfont}
\usepackage{listings}
\usepackage{tcolorbox}
\lstdefinestyle{mystyle}{
    backgroundcolor=\color{backcolour},
    commentstyle=\color{codegreen},
    keywordstyle=\color{magenta},
    numberstyle=\tiny\color{codegray},
    stringstyle=\color{codepurple},
    basicstyle=\ttfamily\footnotesize,
    breakatwhitespace=false,
    breaklines=true,
    captionpos=b,
    keepspaces=true,
    showspaces=false,
    showstringspaces=false,
    showtabs=false,
    tabsize=2,
    xleftmargin=50pt,
    xrightmargin=50pt
  }
\usepackage{tikz,pgfplots}
\usepackage{pgfplots}
\usepackage{pgfplotstable}
\definecolor{gray2}{HTML}{ededed}
\definecolor{gray3}{HTML}{F5F5F5}
\definecolor{RoyalAzure}{rgb}{0.0, 0.22, 0.66}
\definecolor{lightgray}{gray}{0.9}
\definecolor{lightgray}{gray}{0.9}
\definecolor{lightgreen}{rgb}{0.88, 1, 0.88}
\definecolor{lightred}{rgb}{1, 0.88, 0.88}
\definecolor{lightblue}{rgb}{0.88, 0.94, 1}
\usepackage{colortbl}
\usetikzlibrary{shapes.geometric,backgrounds,patterns, trees}
\usetikzlibrary{3d,decorations.text,shapes.arrows,positioning,fit,backgrounds}
\usetikzlibrary{positioning, decorations.pathmorphing, shapes}
\usetikzlibrary{decorations.pathreplacing}
\usetikzlibrary{shapes.geometric,backgrounds,patterns, trees}
\usetikzlibrary{spy}
\usetikzlibrary{arrows.meta,
                bending,
                intersections,
                quotes,
                shapes.geometric}
              \usetikzlibrary{automata, positioning}
              \usepgfplotslibrary{fillbetween}
\usetikzlibrary{shapes,arrows}
\usetikzlibrary{arrows.meta}
\usetikzlibrary{positioning}
\tikzset{set/.style={draw,circle,inner sep=0pt,align=center}}
\usetikzlibrary{automata, positioning}
  \usetikzlibrary{shapes,shadows}
  \tikzstyle{abstractbox} = [draw=black, fill=white, rectangle,
  inner sep=10pt, style=rounded corners, drop shadow={fill=black,
  opacity=1}]
\tikzstyle{abstracttitle} =[fill=white]
\usetikzlibrary{calc,positioning,shapes.geometric}
\usetikzlibrary{arrows.meta,arrows}

\DeclareMathOperator*{\argmin}{arg\,min}

\DeclareMathOperator*{\minimize}{minimize}

\usetikzlibrary{matrix}
\tikzstyle{cblue}=[circle, draw, thin,fill=cyan!20, scale=0.8]
\tikzstyle{qgre}=[rectangle, draw, thin,fill=green!20, scale=0.8]
\tikzstyle{rpath}=[ultra thick, red, opacity=0.4]
\tikzstyle{legend_isps}=[rectangle, rounded corners, thin,
                       fill=gray!20, text=blue, draw]

\tikzstyle{legend_overlay}=[rectangle, rounded corners, thin,
                           top color= white,bottom color=green!25,
                           minimum width=2.5cm, minimum height=0.8cm,
                           pinegreen]
\tikzstyle{legend_phytop}=[rectangle, rounded corners, thin,
                          top color= white,bottom color=cyan!25,
                          minimum width=2.5cm, minimum height=0.8cm,
                          royalblue]
\tikzstyle{legend_general}=[rectangle, rounded corners, thin,
                          top color= white,bottom color=lavander!25,
                          minimum width=2.5cm, minimum height=0.8cm,
                          violet]
                          \usetikzlibrary{matrix}

\colorlet{myRed}{red!20}
\tikzset{
  rows/.style 2 args={/utils/temp/.style={row ##1/.append style={nodes={#2}}},
    /utils/temp/.list={#1}},
  columns/.style 2 args={/utils/temp/.style={column ##1/.append style={nodes={#2}}},
    /utils/temp/.list={#1}}}
\usetikzlibrary{backgrounds,calc,shadings,shapes.arrows,shapes.symbols,shadows}
\definecolor{switch}{HTML}{006996}

\makeatletter
\pgfkeys{/pgf/.cd,
  parallelepiped offset x/.initial=2mm,
  parallelepiped offset y/.initial=2mm
}
\pgfdeclareshape{parallelepiped}
{
  \inheritsavedanchors[from=rectangle] 
  \inheritanchorborder[from=rectangle]
  \inheritanchor[from=rectangle]{north}
  \inheritanchor[from=rectangle]{north west}
  \inheritanchor[from=rectangle]{north east}
  \inheritanchor[from=rectangle]{center}
  \inheritanchor[from=rectangle]{west}
  \inheritanchor[from=rectangle]{east}
  \inheritanchor[from=rectangle]{mid}
  \inheritanchor[from=rectangle]{mid west}
  \inheritanchor[from=rectangle]{mid east}
  \inheritanchor[from=rectangle]{base}
  \inheritanchor[from=rectangle]{base west}
  \inheritanchor[from=rectangle]{base east}
  \inheritanchor[from=rectangle]{south}
  \inheritanchor[from=rectangle]{south west}
  \inheritanchor[from=rectangle]{south east}
  \backgroundpath{
    \southwest \pgf@xa=\pgf@x \pgf@ya=\pgf@y
    \northeast \pgf@xb=\pgf@x \pgf@yb=\pgf@y
    \pgfmathsetlength\pgfutil@tempdima{\pgfkeysvalueof{/pgf/parallelepiped
      offset x}}
    \pgfmathsetlength\pgfutil@tempdimb{\pgfkeysvalueof{/pgf/parallelepiped
      offset y}}
    \def\ppd@offset{\pgfpoint{\pgfutil@tempdima}{\pgfutil@tempdimb}}
    \pgfpathmoveto{\pgfqpoint{\pgf@xa}{\pgf@ya}}
    \pgfpathlineto{\pgfqpoint{\pgf@xb}{\pgf@ya}}
    \pgfpathlineto{\pgfqpoint{\pgf@xb}{\pgf@yb}}
    \pgfpathlineto{\pgfqpoint{\pgf@xa}{\pgf@yb}}
    \pgfpathclose
    \pgfpathmoveto{\pgfqpoint{\pgf@xb}{\pgf@ya}}
    \pgfpathlineto{\pgfpointadd{\pgfpoint{\pgf@xb}{\pgf@ya}}{\ppd@offset}}
    \pgfpathlineto{\pgfpointadd{\pgfpoint{\pgf@xb}{\pgf@yb}}{\ppd@offset}}
    \pgfpathlineto{\pgfpointadd{\pgfpoint{\pgf@xa}{\pgf@yb}}{\ppd@offset}}
    \pgfpathlineto{\pgfqpoint{\pgf@xa}{\pgf@yb}}
    \pgfpathmoveto{\pgfqpoint{\pgf@xb}{\pgf@yb}}
    \pgfpathlineto{\pgfpointadd{\pgfpoint{\pgf@xb}{\pgf@yb}}{\ppd@offset}}
  }
}

\makeatletter
\tikzset{anchor/.append code=\let\tikz@auto@anchor\relax,
  add font/.code=%
    \expandafter\def\expandafter\tikz@textfont\expandafter{\tikz@textfont#1},
  left delimiter/.style 2 args={append after command={\tikz@delimiter{south east}
    {south west}{every delimiter,every left delimiter,#2}{south}{north}{#1}{.}{\pgf@y}}}}
\tikzstyle{sms} = [rectangle callout, draw,very thick, rounded corners, minimum height=20pt]
\makeatletter
\tikzset{anchor/.append code=\let\tikz@auto@anchor\relax,
  add font/.code=%
    \expandafter\def\expandafter\tikz@textfont\expandafter{\tikz@textfont#1},
  left delimiter/.style 2 args={append after command={\tikz@delimiter{south east}
    {south west}{every delimiter,every left delimiter,#2}{south}{north}{#1}{.}{\pgf@y}}}}
\tikzstyle{sms} = [rectangle callout, draw,very thick, rounded corners, minimum height=20pt]
\usetikzlibrary{positioning,calc}
\tikzstyle{block} = [rectangle, draw,
text width=10.5em, text centered, rounded corners, minimum height=4em]
\tikzstyle{line} = [draw, -latex]
\tikzset{
  mybackground51/.style={execute at end picture={
      \begin{scope}[on background layer]
        \draw[black, rounded corners=2ex, fill=gray2] (current bounding box.south west)
        rectangle (current bounding box.north east);
        \node[draw,fill=white,ellipse,anchor=west,inner sep=1pt,minimum width=1ex] at (current bounding box.north
        west){#1};
      \end{scope}
    }},
}
\tikzset{
  mybackground9/.style={execute at end picture={
        \begin{scope}[on background layer]
          \draw[black,fill=black!5,rounded corners=6ex] (current bounding box.south west)
                    rectangle (current bounding box.north east);
          \node[draw,fill=white,ellipse,anchor=west,inner sep=1pt,minimum width=4ex] at (current bounding box.north
                   west){#1};
        \end{scope}
    }},
}

\tikzset{
  mybackground13/.style={execute at end picture={
        \begin{scope}[on background layer]
          \draw[black, fill=gray2, rounded corners=4ex] (current bounding box.south west)
                    rectangle (current bounding box.north east);
          \node[draw,fill=white,ellipse,anchor=west,inner sep=1pt,minimum width=4ex] at (current bounding box.north
                   west){#1};
        \end{scope}
    }},
}
\tikzset{
  mybackground14/.style={execute at end picture={
        \begin{scope}[on background layer]
          \draw[black, rounded corners=2ex] (current bounding box.south west)
                    rectangle (current bounding box.north east);
          \node[draw,fill=white,ellipse,anchor=west,inner sep=1pt,minimum width=4ex] at (current bounding box.north
                   west){#1};
        \end{scope}
    }},
}

\tikzset{
  mybackground6/.style={execute at end picture={
        \begin{scope}[on background layer]
          \draw[black,rounded corners=1ex, line width=0.15mm] (current bounding box.south west)
                    rectangle (current bounding box.north east);
          \node[draw,fill=white,ellipse,anchor=west,inner sep=1pt,minimum width=4ex] at (current bounding box.north
                   west){#1};
        \end{scope}
    }},
}

\tikzset{
  mybackground11/.style={execute at end picture={
        \begin{scope}[on background layer]
          \draw[black, fill=Black!80!Sepia!9, rounded corners=6ex] (current bounding box.south west)
                    rectangle (current bounding box.north east);
          \node[draw,fill=white,ellipse,anchor=west,inner sep=1pt,minimum width=4ex] at (current bounding box.north
                   west){#1};
        \end{scope}
    }},
}

\tikzset{
  mybackground15/.style={execute at end picture={
        \begin{scope}[on background layer]
          \draw[black, fill=Black!80!Sepia!9, rounded corners=3ex] (current bounding box.south west)
                    rectangle (current bounding box.north east);
          \node[draw,fill=white,ellipse,anchor=west,inner sep=1pt,minimum width=4ex] at (current bounding box.north
                   west){#1};
        \end{scope}
    }},
}

\tikzset{
  mybackground12/.style={execute at end picture={
        \begin{scope}[on background layer]
          \draw[black, fill=Black!40!Emerald!30, rounded corners=3ex, line width=0.3mm] (current bounding box.south west)
                    rectangle (current bounding box.north east);
        \end{scope}
    }},
}
\tikzset{
  mybackground18/.style={execute at end picture={
      \begin{scope}[on background layer]
        \draw[black, fill=gray3, rounded corners=3.5ex] (current bounding box.south west)
        rectangle (current bounding box.north east);
        \node[draw,fill=white,ellipse,anchor=west,inner sep=1pt,minimum width=4ex] at (current bounding box.north
        west){#1};
      \end{scope}
    }}
}
\tikzset{
  mybackground58/.style={execute at end picture={
        \begin{scope}[on background layer]
          \draw[black, fill=blue!40!black!5, rounded corners=1ex] (current bounding box.south west)
                    rectangle (current bounding box.north east);
          \node[draw,fill=white,ellipse,anchor=west,inner sep=1pt,minimum width=4ex, rounded corners=1ex] at (current bounding box.north
                   west){#1};
        \end{scope}
    }},
}
\tikzset{l3 switch/.style={
    parallelepiped,fill=switch, draw=white,
    minimum width=0.75cm,
    minimum height=0.75cm,
    parallelepiped offset x=1.75mm,
    parallelepiped offset y=1.25mm,
    path picture={
      \node[fill=white,
        circle,
        minimum size=6pt,
        inner sep=0pt,
        append after command={
          \pgfextra{
            \foreach \angle in {0,45,...,360}
            \draw[-latex,fill=white] (\tikzlastnode.\angle)--++(\angle:2.25mm);
          }
        }
      ]
       at ([xshift=-0.75mm,yshift=-0.5mm]path picture bounding box.center){};
    }
  },
  ports/.style={
    line width=0.3pt,
    top color=gray!20,
    bottom color=gray!80
  },
  rack switch/.style={
    parallelepiped,fill=white, draw,
    minimum width=1.25cm,
    minimum height=0.25cm,
    parallelepiped offset x=2mm,
    parallelepiped offset y=1.25mm,
    xscale=-1,
    path picture={
      \draw[top color=gray!5,bottom color=gray!40]
      (path picture bounding box.south west) rectangle
      (path picture bounding box.north east);
      \coordinate (A-west) at ([xshift=-0.2cm]path picture bounding box.west);
      \coordinate (A-center) at ($(path picture bounding box.center)!0!(path
        picture bounding box.south)$);
      \foreach \x in {0.275,0.525,0.775}{
        \draw[ports]([yshift=-0.05cm]$(A-west)!\x!(A-center)$)
          rectangle +(0.1,0.05);
        \draw[ports]([yshift=-0.125cm]$(A-west)!\x!(A-center)$)
          rectangle +(0.1,0.05);
       }
      \coordinate (A-east) at (path picture bounding box.east);
      \foreach \x in {0.085,0.21,0.335,0.455,0.635,0.755,0.875,1}{
        \draw[ports]([yshift=-0.1125cm]$(A-east)!\x!(A-center)$)
          rectangle +(0.05,0.1);
      }
    }
  },
  server/.style={
    parallelepiped,
    fill=white, draw,
    minimum width=0.35cm,
    minimum height=0.75cm,
    parallelepiped offset x=3mm,
    parallelepiped offset y=2mm,
    xscale=-1,
    path picture={
      \draw[top color=gray!5,bottom color=gray!40]
      (path picture bounding box.south west) rectangle
      (path picture bounding box.north east);
      \coordinate (A-center) at ($(path picture bounding box.center)!0!(path
        picture bounding box.south)$);
      \coordinate (A-west) at ([xshift=-0.575cm]path picture bounding box.west);
      \draw[ports]([yshift=0.1cm]$(A-west)!0!(A-center)$)
        rectangle +(0.2,0.065);
      \draw[ports]([yshift=0.01cm]$(A-west)!0.085!(A-center)$)
        rectangle +(0.15,0.05);
      \fill[black]([yshift=-0.35cm]$(A-west)!-0.1!(A-center)$)
        rectangle +(0.235,0.0175);
      \fill[black]([yshift=-0.385cm]$(A-west)!-0.1!(A-center)$)
        rectangle +(0.235,0.0175);
      \fill[black]([yshift=-0.42cm]$(A-west)!-0.1!(A-center)$)
        rectangle +(0.235,0.0175);
    }
  },
}

\usetikzlibrary{calc, shadings, shadows, shapes.arrows}
\tikzset{cross/.style={cross out, draw=black, minimum size=2*(#1-\pgflinewidth), inner sep=0pt, outer sep=0pt},
cross/.default={1pt}}
\tikzset{%
  interface/.style={draw, rectangle, rounded corners, font=\LARGE\sffamily},
  ethernet/.style={interface, fill=yellow!50},
  serial/.style={interface, fill=green!70},
  speed/.style={sloped, anchor=south, font=\large\sffamily},
  route/.style={draw, shape=single arrow, single arrow head extend=4mm,
    minimum height=1.7cm, minimum width=3mm, white, fill=switch!20,
    drop shadow={opacity=.8, fill=switch}, font=\tiny}
}
\newcommand*{\shift}{1.3cm}
\newcommand*{\router}[1]{
\begin{tikzpicture}
  \coordinate (ll) at (-3,0.5);
  \coordinate (lr) at (3,0.5);
  \coordinate (ul) at (-3,2);
  \coordinate (ur) at (3,2);
  \shade [shading angle=90, left color=switch, right color=white] (ll)
    arc (-180:-60:3cm and .75cm) -- +(0,1.5) arc (-60:-180:3cm and .75cm)
    -- cycle;
  \shade [shading angle=270, right color=switch, left color=white!50] (lr)
    arc (0:-60:3cm and .75cm) -- +(0,1.5) arc (-60:0:3cm and .75cm) -- cycle;
  \draw [thick] (ll) arc (-180:0:3cm and .75cm)
    -- (ur) arc (0:-180:3cm and .75cm) -- cycle;
  \draw [thick, shade, upper left=switch, lower left=switch,
    upper right=switch, lower right=white] (ul)
    arc (-180:180:3cm and .75cm);
  \node at (0,0.5){\color{blue!60!black}\Huge #1};
  \begin{scope}[yshift=2cm, yscale=0.28, transform shape]
    \node[route, rotate=45, xshift=\shift] {\strut};
    \node[route, rotate=-45, xshift=-\shift] {\strut};
    \node[route, rotate=-135, xshift=\shift] {\strut};
    \node[route, rotate=135, xshift=-\shift] {\strut};
  \end{scope}
\end{tikzpicture}}

\makeatletter
\pgfdeclareradialshading[tikz@ball]{cloud}{\pgfpoint{-0.275cm}{0.4cm}}{%
  color(0cm)=(tikz@ball!75!white);
  color(0.1cm)=(tikz@ball!85!white);
  color(0.2cm)=(tikz@ball!95!white);
  color(0.7cm)=(tikz@ball!89!black);
  color(1cm)=(tikz@ball!75!black)
}
\tikzoption{cloud color}{\pgfutil@colorlet{tikz@ball}{#1}%
  \def\tikz@shading{cloud}\tikz@addmode{\tikz@mode@shadetrue}}
\makeatother

\tikzset{my cloud/.style={
     cloud, draw, aspect=2,
     cloud color={gray!5!white}
  }
}

\usetikzlibrary{backgrounds}
\usetikzlibrary{patterns}
\pgfmathdeclarefunction{gauss}{3}{%
  \pgfmathparse{1/(#3*sqrt(2*pi))*exp(-((#1-#2)^2)/(2*#3^2))}%
}
\pgfmathdeclarefunction{cdf}{3}{%
  \pgfmathparse{1/(1+exp(-0.07056*((#1-#2)/#3)^3 - 1.5976*(#1-#2)/#3))}%
}
\pgfmathdeclarefunction{fq}{3}{%
  \pgfmathparse{1/(sqrt(2*pi*#1))*exp(-(sqrt(#1)-#2/#3)^2/2)}%
}
\pgfmathdeclarefunction{fq0}{1}{%
  \pgfmathparse{1/(sqrt(2*pi*#1))*exp(-#1/2)}%
}
\renewcommand\thetable{\arabic{table}}
\usepackage{etoolbox}

\usepackage{wrapfig}

\newcommand{\myboxxx}[4]{
    \begin{figure}[H]
        \centering
    \begin{tikzpicture}
      \node[anchor=text,text width=\columnwidth-1.2cm, draw, rounded corners, line width=1pt, fill=#3, inner sep=3.5mm] (big) {\\#4};
        \node[draw, rounded corners, line width=.5pt, fill=#2, anchor=west, xshift=5mm] (small) at (big.north west) {#1};
    \end{tikzpicture}
    \end{figure}

  }

\makeatletter
\newcommand{\setword}[2]{%
  \phantomsection
  #1\def\@currentlabel{\unexpanded{#1}}\label{#2}%
}
\makeatother
\usepackage{lettrine}
 \definecolor{DBrown}{HTML}{9B8879}
 \definecolor{LBrown}{HTML}{C5B99F}
 \definecolor{backg}{HTML}{BCC534}
 \definecolor{latCol}{HTML}{E8B041}
 \definecolor{pot}{HTML}{185BD9}

\tikzset{%
  wireless/.pic={
      \draw [->] (0,0) -| (.5,#1);
    \foreach \r in {.1,.2,.3}
      \draw (.6,#1) ++ (60:\r) arc (60:-60:\r);
  },
  vdots/.pic={
    \foreach \i in {-.1,0,.1}
      \fill (.25,\i) circle [radius=.75pt];
  },
  block/.style={
    shape=rectangle,
    minimum width=2cm,
    minimum height=1cm,
    draw
  },
  Tx/.style 2 args={
    block,
    node contents=Tx,
    append after command={
      \pgfextra{\pgfnodealias{@}{\tikzlastnode}}
      (@.north #1) [yshift=-.125cm] pic [#2] {wireless=.5}
      (@.#1)                        pic [#2] {vdots}
      (@.south #1) [yshift= .125cm] pic [#2] {wireless=.5}
    }
  },
  MIMO Tx east/.style={Tx={east}{xscale=1}},
  MIMO Tx west/.style={Tx={west}{xscale=-1}},
  Tx2/.style 2 args={
    node contents=,
    append after command={
      \pgfextra{\pgfnodealias{@}{\tikzlastnode}}
      (@.north #1) [yshift=-.125cm] pic [#2] {wireless=.5}
      (@.south #1) [yshift= .125cm] pic [#2] {wireless=.5}
    }
  },
  MIMO2 Tx east/.style={Tx2={east}{xscale=1}},
  MIMO2 Tx west/.style={Tx2={west}{xscale=-1}}
}

\newcommand{\figref}[1]{\hyperref[#1]{Fig.~\ref*{#1}}}
\newcommand{\Figref}[1]{\hyperref[#1]{Figure~\ref*{#1}}}
\newcommand{\Figsref}[1]{\hyperref[#1]{Figures~\ref*{#1}}}
\newcommand{\figsref}[1]{\hyperref[#1]{Figs.~\ref*{#1}}}
\newcommand{\tableref}[1]{\hyperref[#1]{Table~\ref*{#1}}}
\newcommand{\tablesref}[1]{\hyperref[#1]{Tables~\ref*{#1}}}
\newcommand{\appendixref}[1]{\hyperref[#1]{Appendix~\ref*{#1}}}
\newcommand{\theoremref}[1]{\hyperref[#1]{Thm.~\ref*{#1}}}
\newcommand{\Theoremref}[1]{\hyperref[#1]{Theorem~\ref*{#1}}}
\newcommand{\lemmaref}[1]{\hyperref[#1]{Lemma~\ref*{#1}}}
\newcommand{\propref}[1]{\hyperref[#1]{Prop.~\ref*{#1}}}
\newcommand{\propsref}[1]{\hyperref[#1]{Props.~\ref*{#1}}}
\newcommand{\Propref}[1]{\hyperref[#1]{Proposition~\ref*{#1}}}
\newcommand{\corref}[1]{\hyperref[#1]{Cor.~\ref*{#1}}}
\newcommand{\Corref}[1]{\hyperref[#1]{Corollary~\ref*{#1}}}
\newcommand{\scenarioref}[1]{\hyperref[#1]{Scenario~\ref*{#1}}}
\newcommand{\Scenarioref}[1]{\hyperref[#1]{\textsc{scenario}~\ref*{#1}}}
\newcommand{\probref}[1]{\hyperref[#1]{Prob.~\ref*{#1}}}
\newcommand{\Probref}[1]{\hyperref[#1]{Problem~\ref*{#1}}}
\newcommand{\gameref}[1]{\hyperref[#1]{Game~\ref*{#1}}}
\newcommand{\chapterref}[1]{\hyperref[#1]{Chapter~\ref*{#1}}}
\newcommand{\sectionref}[1]{\hyperref[#1]{\S\ref*{#1}}}
\newcommand{\Algref}[1]{\hyperref[#1]{Algorithm ~\ref*{#1}}}
\newcommand{\myalgref}[1]{\hyperref[#1]{Alg.~\ref*{#1}}}
\newcommand{\Myalgref}[1]{\hyperref[#1]{Algorithm~\ref*{#1}}}
\newcommand{\defref}[1]{\hyperref[#1]{Def.~\ref*{#1}}}
\newcommand{\Defref}[1]{\hyperref[#1]{Definition~\ref*{#1}}}
\newcommand{\assumptionref}[1]{\hyperref[#1]{Assumption~\ref*{#1}}}
\newcommand{\remarkref}[1]{\hyperref[#1]{Remark~\ref*{#1}}}
\newcommand{\exampleref}[1]{\hyperref[#1]{Ex.~\ref*{#1}}}

\usepackage{pifont}

\newtcolorbox{problemone}{
  colback=black!5!white,
  colframe=black!30!black!70,
  title=Problem (Online identification of an \textsc{it} system),
  fonttitle=\bfseries,
  sharp corners
}

\newtcolorbox{summary}{
  float,
  colback=black!5!white,
  colframe=black!30!black!70,
  title=Summary of our identification method ({\hypersetup{linkcolor=white}\figref{fig:framework}}),
  fonttitle=\bfseries,
  sharp corners
}

\tikzstyle{mynode}=[thick,draw=blue,fill=blue!50!black,circle,minimum size=22]

\usepackage{stfloats}

\makeatletter
\patchcmd{\@makecaption}
  {\scshape}
  {}
  {}
  {}
\makeatother

\allowdisplaybreaks
\begin{document}
\bstctlcite{MyBSTcontrol}

\title{Online Identification of IT Systems\\through Active Causal Learning}

\author{\IEEEauthorblockN{Kim Hammar and Rolf Stadler}\\
 \IEEEauthorblockA{
   KTH Royal Institute of Technology, Sweden}\\
 Email: kimham@kth.se and stadler@kth.se\\
\today
}
\maketitle
\begin{abstract}
Identifying a causal model of an \textsc{it} system is fundamental to many branches of systems engineering and operation. Such a model can be used to predict the effects of control actions, optimize operations, diagnose failures, detect intrusions, etc., which is central to achieving the longstanding goal of automating network and system management tasks. Traditionally, causal models have been designed and maintained by domain experts. This, however, proves increasingly challenging with the growing complexity and dynamism of modern \textsc{it} systems. In this paper, we present the first principled method for online, data-driven identification of an \textsc{it} system in the form of a causal model. The method, which we call \textit{active causal learning}, estimates causal functions that capture the dependencies among system variables in an iterative fashion using Gaussian process regression based on system measurements, which are collected through a rollout-based intervention policy. We prove that this method is optimal in the Bayesian sense and that it produces effective interventions. Experimental validation on a testbed shows that our method enables accurate identification of a causal system model while inducing low interference with system operations.
\end{abstract}
\section{Introduction}
\lettrine[lines=2]{\textbf{S}}{ignificant} progress in autonomous management of \textsc{it} systems is required to achieve reliable operation and predictable service quality as these systems are becoming increasingly complex and dynamic. Efforts towards automating the management of networks and \textsc{it} systems have been undertaken over the last 30 years, initially motivated by the lack of experts who could reliably configure and maintain these increasingly capable systems and technologies. Starting with policy-based management in the 1990s (e.g., \cite{sloman1994policy}), a series of paradigms have been proposed and developed, often initiated by industry and then studied in collaboration with academia. These efforts include autonomic management (e.g., \cite{kephart2003vision}), self-organizing networks (e.g., \cite{prehofer2005self}), intent-based network management (e.g., \cite{clemm2022rfc}), and zero-touch management (e.g., \cite{9738820}). A comprehensive overview of these developments from a networking perspective is provided by Coronado et al. \cite{coronado2022zero}.

We advocate for a principled approach to autonomous management that is based on a formal foundation. Specifically, we propose to construct and maintain a \textit{causal model} of an \textsc{it} system under consideration. The formal concept of causality and the theory of causal models that we use in this paper have been established in the seminal work by Pearl and collaborators \cite{pearl2000causality, shpitser2008complete}, and the connection to machine learning has been investigated more recently by Peters et al. \cite{peters2017elements}.
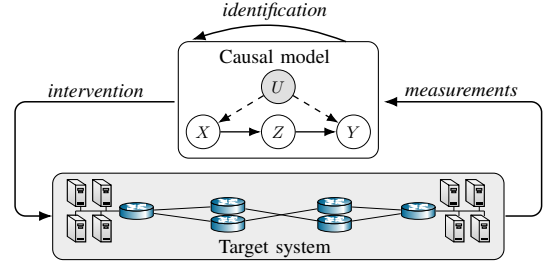
\begin{figure}
  \centering
  \scalebox{1}{
          \begin{tikzpicture}[fill=white, >=stealth,
    node distance=3cm,
    database/.style={
      cylinder,
      cylinder uses custom fill,
      shape border rotate=90,
      aspect=0.25,
      draw}]

    \tikzset{
node distance = 9em and 4em,
sloped,
   box/.style = {%
    shape=rectangle,
    rounded corners,
    draw=blue!40,
    fill=blue!15,
    align=center,
    font=\fontsize{12}{12}\selectfont},
 arrow/.style = {%
    line width=0.1mm,
    -{Triangle[length=5mm,width=2mm]},
    shorten >=1mm, shorten <=1mm,
    font=\fontsize{8}{8}\selectfont},
}

\draw[-, rounded corners] (0.67,3) rectangle node (m1){} (3.33,1.45);

\node[scale=0.8] (gi) at (2,2.1)
{
  \begin{tikzpicture}
\node[draw,circle, minimum width=0.8cm, scale=0.7](x) at (0,0) {};
\node[draw,circle, minimum width=0.8cm, scale=0.7](z) at (1.25,0) {};
\node[draw,circle, minimum width=0.8cm, scale=0.7](y) at (2.5,0) {};
\node[draw,circle, minimum width=0.8cm, scale=0.7, fill=black!12](u) at (1.25,0.75) {};
\draw[-{Latex[length=2mm]}, line width=0.22mm, color=black] (x) to (z);
\draw[-{Latex[length=2mm]}, line width=0.22mm, color=black] (z) to (y);
\draw[-{Latex[length=2mm]}, line width=0.22mm, color=black,dashed] (u) to (y);
\draw[-{Latex[length=2mm]}, line width=0.22mm, color=black,dashed] (u) to (x);
\node[inner sep=0pt,align=center, scale=0.8] (dots4) at (1.25,0.75) {$U$};
\node[inner sep=0pt,align=center, scale=0.8] (dots4) at (0,0) {$X$};
\node[inner sep=0pt,align=center, scale=0.8] (dots4) at (1.25,0) {$Z$};
\node[inner sep=0pt,align=center, scale=0.8] (dots4) at (2.5,0) {$Y$};
\end{tikzpicture}
};

\draw[fill=gray2, rounded corners] (-1,0.1) rectangle node (m1){} (5,1.25);

\node [scale=0.75] (node1) at (2,0.75) {
\begin{tikzpicture}[fill=white, >=stealth,
    node distance=3cm,
    database/.style={
      cylinder,
      cylinder uses custom fill,
      shape border rotate=90,
      aspect=0.25,
      draw}]
    \tikzset{
node distance = 9em and 4em,
sloped,
   box/.style = {%
    shape=rectangle,
    rounded corners,
    draw=blue!40,
    fill=blue!15,
    align=center,
    font=\fontsize{12}{12}\selectfont},
 arrow/.style = {%
    line width=0.1mm,
    shorten >=1mm, shorten <=1mm,
    font=\fontsize{8}{8}\selectfont},
}

\node[scale=1] (emulation_system) at (-0.07,-2.7)
{
\begin{tikzpicture}

\node[scale=0.9] (cloud_1) at (3,0.5)
{
  \begin{tikzpicture}

 \node[scale=0.11](r1) at (2.4,-5.75) {\router{}};
 \node[scale=0.11](r2) at (7.95,-5.75) {\router{}};
 \node[scale=0.11](r3) at (4.2,-5.65) {\router{}};
 \node[scale=0.11](r4) at (6.3,-5.65) {\router{}};
 \node[scale=0.11](r5) at (4.2,-6) {\router{}};
\node[scale=0.11](r6) at (6.3,-6) {\router{}};
  \draw[-, color=black] (r1) to (r3);
  \draw[-, color=black] (r1) to (r5);
  \draw[-, color=black] (r5) to (r4);
  \draw[-, color=black] (r3) to (r6);
  \draw[-, color=black] (r2) to (r6);
  \draw[-, color=black] (r2) to (r4);
  \draw[-, color=black] (r1) to (1.2, -5.75);
  \draw[-, color=black] (r2) to (9.25, -5.75);

  \draw[-, color=black] (8.65, -5.75) to (8.65, -5.65);
  \draw[-, color=black] (8.75, -5.75) to (8.75, -5.85);

  \draw[-, color=black] (9.15, -5.75) to (9.15, -5.65);
  \draw[-, color=black] (9.25, -5.75) to (9.25, -5.85);

  \node[server, scale=0.55](n1) at (8.65,-5.45) {};
  \node[server, scale=0.55](n1) at (8.75,-6.15) {};

  \node[server, scale=0.55](n1) at (9.15,-5.45) {};
  \node[server, scale=0.55](n1) at (9.25,-6.15) {};

  \node[server, scale=0.55](n1) at (1.8,-5.45) {};
  \node[server, scale=0.55](n1) at (1.8,-6.15) {};
  \draw[-, color=black] (1.8, -5.75) to (1.8, -5.65);
  \draw[-, color=black] (1.7, -5.75) to (1.7, -5.85);

  \node[server, scale=0.55](n1) at (1.3,-5.45) {};
  \node[server, scale=0.55](n1) at (1.3,-6.15) {};

  \draw[-, color=black] (1.3, -5.75) to (1.3, -5.65);
  \draw[-, color=black] (1.2, -5.75) to (1.2, -5.85);


    \end{tikzpicture}
  };
    \end{tikzpicture}
  };
\end{tikzpicture}
};

\node[inner sep=0pt,align=center, scale=0.75, color=black] (hacker) at (2,3.4) {
  \textit{identification}
};

\node[inner sep=0pt,align=center, scale=0.75, color=black] (hacker) at (4.48,2.35) {
  \textit{measurements}
};

\node[inner sep=0pt,align=center, scale=0.75, color=black] (hacker) at (-0.38,2.35) {
  \textit{intervention}
};

\node[inner sep=0pt,align=center, scale=0.75, color=black] (hacker) at (2,0.25) {
  Target system
};
\node[inner sep=0pt,align=center, scale=0.75, color=black] (hacker) at (2,2.8) {
  Causal model
};

\draw[-{Latex[length=1.8mm]}, black, thick, line width=0.2mm, rounded corners] (5.02,0.68) to (5.5,0.68) to (5.5, 2.2) to (3.4, 2.2);

\draw[-{Latex[length=1.8mm]}, black, thick, line width=0.2mm, rounded corners] (0.62, 2.2) to (-1.5, 2.2) to (-1.5, 0.68) to (-1, 0.68);

\draw[-{Latex[length=1.8mm]}, black, thick, line width=0.2mm, rounded corners, bend right=20] (2.9, 3) to (0.85, 3.025);

\end{tikzpicture}       
  }
\caption{Online identification of \textsc{it} systems through active causal learning.}
\label{fig:system}
\end{figure}

A causal model of an \textsc{it} system captures the causal relations between key variables that characterize the system's infrastructure (e.g., the available memory), its services (e.g., the response time of a service request), and external factors (e.g., the load generated by users of the services). Knowing the causal model of an \textsc{it} system allows to predict how the system will react to a control action, such as scaling the \textsc{cpu} allocation, or to a change in an external factor, such as the service load. It allows building autonomous resource allocation functions that achieve management objectives in a changing environment. Also, it provides a formal understanding of the system dynamics and their relation to management objectives.

Such a model is defined by a directed graph, called the \textit{causal graph}, which expresses the causal dependencies among system variables, and by a set of \textit{causal functions}, which capture the functional dependencies among the variables. Since causal dependencies can often be deduced from the hardware and software architecture of the system and since they seldom change during operation, we assume in this paper that the causal graph is known, and we focus on identifying the causal functions of a causal model that represents the \textsc{it} system. Specifically, we present an \textit{online} method for identifying the causal functions of an \textsc{it} system and for updating them over time. We call this method \textit{active causal learning}. 

\begin{figure}
  \centering
  \scalebox{0.93}{
    \input{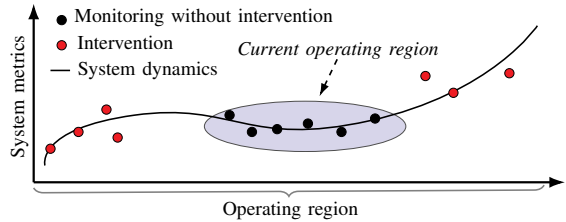}    
  }
\caption{Comparison between data collection with and without interventions. Monitoring the system without interventions yields measurements only within the system’s \textit{current} operating region \cite{JOHANSEN1995321}, while interventions allow for collecting measurement samples across the (complete) operating region.}
\label{fig:operating_region}
\end{figure}

In our method, we learn the causal model in an iterative manner by combining continuous monitoring with a sequence of interventions on the \textsc{it} system; see \figref{fig:system}. During an intervention, we set one or more control variables of the system to new values and measure the resulting change in the system variables. For example, an intervention in this context may be to temporarily adjust the \textsc{cpu} limit of a service and observe how this change affects response times of service requests. By combining interventions with continuous monitoring, we can obtain measurement samples across the system’s (complete) \textit{operating region}, i.e., the combinations of workloads and configurations of control variables under which the system is designed to operate \cite{feedback_control_computing_systems}; see \figref{fig:operating_region}.

Using the measurements up to the current time, we produce a new estimate of the causal model after each monitoring interval. Specifically, we estimate the causal functions of the \textsc{it} system through Gaussian process (\textsc{gp}) regression. This approach allows us to quantify the uncertainty in the current estimates, which we use to guide the selection of interventions. We show that the problem of selecting interventions that reduce model uncertainty while having a low operational cost can be formulated as a dynamic programming problem. Solving this problem is computationally challenging as the number of possible interventions grows exponentially with the number of system variables. Moreover, evaluating the expected effect of an intervention involves computing high-dimensional integrals. We address these challenges by designing an efficient rollout algorithm for approximating optimal intervention policies through lookahead optimization.

We prove that our method is optimal in the Bayesian sense and establish conditions under which our rollout algorithm produces effective interventions. We evaluate the method in a testbed where we set up an \textsc{it} system with two web services based on a microservice architecture. During operation of this system, we continually estimate its causal model and use it to predict service response times in function of control variable settings. The experimental results are consistent with our theoretical claims and show that our method can closely track the evolving model of a dynamic system.

The main contributions of this paper are:
\begin{itemize}
\item We present the first principled method for online, data-driven identification of an \textsc{it} system, which we call \textit{active causal learning}. The method estimates a causal model of the system in an iterative fashion, involving \textsc{gp} regression, rollout, and lookahead optimization.    
\item We prove that our method is optimal in the Bayesian sense and produces effective intervention policies.
\item We experimentally validate the method on a testbed \cite{noms24_rollout}. The results show that it enables efficient and accurate identification of a causal system model while inducing low interference with system operations.
\end{itemize}

\section{Example Use Case: Performance Objectives}

Consider an \textsc{it} system that provides network services to a client population. Continuously meeting performance objectives for these services requires the system to periodically take control actions, such as scaling resources when the load increases, or relocating network functions in response to failures of system components. To automate the selection of such actions, a model is required that captures how changes to certain system variables (e.g., \textsc{cpu} allocations) affect changes in others (e.g., response times). In other words, the model must convey cause-and-effect relationships among system variables.

Such causal relationships are governed by complex interactions among system components and further depend on external factors. For example, relocating a network function to mitigate a failure may improve service availability while, at the same time, increasing latency or congestion in other parts of the network. Moreover, these causal relationships are time-varying and dependent on underlying hardware architectures.

\section{Problem Statement}
We consider the problem of identifying a causal model of an \textsc{it} system based on measurement data from the system. We assume that causal dependencies among the system variables are known and expressed in the form of a \textit{causal graph} \cite[Def. 2.2.1]{pearl2000causality}. This graph encodes structural properties, which define a set of candidate models. Our goal is to identify the most suitable model within this set based on the available data. This data can be obtained either through monitoring or through interventions on specific system variables.

Since the system and its operating conditions may change over time, the identification must be performed \textit{online} and involves two interconnected tasks: (\textit{i}) estimating a causal model based on the available data; and (\textit{ii}) selecting interventions for collecting new measurements. These tasks are carried out in an iterative process, where each new measurement informs the next model update. When designing this process, our goal is to track the evolution of the system while keeping intervention costs low. The next section formalizes this problem.
\section{The Online Causal Identification Problem}\label{sec:online_prob_def}
To define a causal model for an \textsc{it} system, we adopt the formalism of structural causal models (\textsc{scm}) \cite[Def 7.1.1]{pearl2000causality}. Following this formalism, we define the system model as
\begin{align}
\mathscr{M}_t \triangleq \left\langle \mathbf{U},\mathbf{V}, \mathbf{F}_t, \mathbb{P}[\mathbf{U}] \right\rangle, && t=1,2,\hdots, \label{eq:scm_def}
\end{align}
where $\mathbf{U}$ and $\mathbf{V}$ are finite sets of (real-valued) \textit{exogenous} and \textit{endogenous} random variables, respectively. The exogenous variables represent external factors such as the service load, while the endogenous ones describe internal system properties like response time. Among these variables, we distinguish between those that can be directly controlled (e.g., \textsc{cpu} allocation), denoted by $\mathbf{X}$, and those that cannot (e.g., the end-to-end response time of a service request), denoted by $\mathbf{N}$.

We assume that the sets $\mathbf{U}$ and $\mathbf{V}$ are chosen so that all possible configurations of the variables in $\mathbf{U} \cup \mathbf{V}$ define the system’s \textit{operating region} \cite{JOHANSEN1995321}, i.e., the set of configurations for which the system is intended to function. Formally,
\begin{definition}[Operating region]\label{def:or}
Given a structural causal model $\mathscr{M}_t$ [cf.~(\ref{eq:scm_def})] of an \textsc{it} system, the system's (complete) operating region is given by
\begin{align*}
\mathcal{O} \triangleq \mathcal{R}(\mathbf{U}) \times \mathcal{R}(\mathbf{V}),
\end{align*}
where $\mathcal{R}(\cdot)$ denotes the range of a set of random variables. 
\end{definition}  

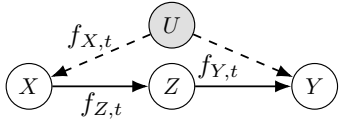
\begin{figure}[H]
  \centering
  \scalebox{1.55}{
   \begin{tikzpicture}[fill=white, >=stealth,
    node distance=3cm,
    database/.style={
      cylinder,
      cylinder uses custom fill,
      shape border rotate=90,
      aspect=0.25,
      draw}]

    \tikzset{
node distance = 9em and 4em,
sloped,
   box/.style = {%
    shape=rectangle,
    rounded corners,
    draw=blue!40,
    fill=blue!15,
    align=center,
    font=\fontsize{12}{12}\selectfont},
 arrow/.style = {%
    line width=0.1mm,
    -{Triangle[length=5mm,width=2mm]},
    shorten >=1mm, shorten <=1mm,
    font=\fontsize{8}{8}\selectfont},
}


\node[scale=0.7] (gi) at (0,0)
{
  \begin{tikzpicture}
\node[draw,circle, minimum width=0.8cm, scale=0.7](x) at (0,0) {};
\node[draw,circle, minimum width=0.8cm, scale=0.7](z) at (1.75,0) {};
\node[draw,circle, minimum width=0.8cm, scale=0.7](y) at (3.5,0) {};
\node[draw,circle, minimum width=0.8cm, scale=0.7, fill=black!12](u) at (1.75,0.75) {};
\draw[-{Latex[length=2mm]}, line width=0.22mm, color=black] (x) to (z);
\draw[-{Latex[length=2mm]}, line width=0.22mm, color=black] (z) to (y);
\draw[-{Latex[length=2mm]}, line width=0.22mm, color=black,dashed] (u) to (y);
\draw[-{Latex[length=2mm]}, line width=0.22mm, color=black,dashed] (u) to (x);
\node[inner sep=0pt,align=center, scale=0.8] (dots4) at (1.75,0.75) {$U$};
\node[inner sep=0pt,align=center, scale=0.8] (dots4) at (0,0) {$X$};
\node[inner sep=0pt,align=center, scale=0.8] (dots4) at (1.75,0) {$Z$};
\node[inner sep=0pt,align=center, scale=0.8] (dots4) at (3.5,0) {$Y$};
\end{tikzpicture}
};


\node[inner sep=0pt,align=center, scale=0.65] (dots4) at (-0.6,-0.43) {$f_{Z,t}$};
\node[inner sep=0pt,align=center, scale=0.65] (dots4) at (-0.7,0.155) {$f_{X,t}$};
\node[inner sep=0pt,align=center, scale=0.65] (dots4) at (0.4,-0.1) {$f_{Y,t}$};


\end{tikzpicture} 
  }
  \caption{A causal graph \cite[Def. 2.2.1]{pearl2000causality}; circles represent variables in an \textsc{scm}; cf.~(\ref{eq:scm_def}); solid arrows represent causal dependencies between endogenous variables; dashed arrows represent causal dependencies on exogenous variables.}
  \label{fig:causal_graphs_intro}
\end{figure}

Dependencies among variables are encoded in a (directed and acyclic) \textit{causal graph} $\mathcal{G}$, whose nodes correspond to elements of $\mathbf{U} \cup \mathbf{V}$ and edges represent causal functions; see \figref{fig:causal_graphs_intro}. Specifically, each endogenous variable $V_i$ is determined by a \textit{causal function} $f_{V_{i,t}}$, which maps its parent variables in the graph to its output value. For example, a causal function may take the form $R = f_{R,t}(L)$, where $R$ represents response time and $L$ represents system load. The collection of all such functions at time $t$ is denoted by $\mathbf{F}_t \triangleq \{f_{V_{i},t}\}_{V_{i} \in \mathbf{V}}$. We consider that these functions may evolve over time and that they are unknown. Further, we assume that the causal graph $\mathcal{G}$ and the probability distribution $\mathbb{P}[\mathbf{U}]$ are fixed and known.

In the context of \textsc{it} systems, samples of the variables $\mathbf{V} \cup \mathbf{U}$ correspond to logs and metrics measured on the target system. Following the formalism of \textsc{scm}s, such samples can be drawn from any probability distribution from the set
\begin{align}
\{\mathbb{P}[\mathbf{V},\mathbf{U} \mid \mathrm{do}(\mathbf{X}^{\prime}=\mathbf{x}^{\prime})] \mid \mathbf{X}^{\prime} \in 2^{\mathbf{X}}, \mathbf{x}^{\prime} \in \mathcal{R}(\mathbf{X}^{\prime})\},\label{eq:distributions}
\end{align}
where $2^{\mathbf{X}}$ is the powerset of $\mathbf{X}$, $\mathcal{R}(\mathbf{X}^{\prime})$ is the range of $\mathbf{X}^{\prime}$ (i.e., the set of values $\mathbf{X}^{\prime}$ can take on), and $\mathrm{do}(\mathbf{X}^{\prime}=\mathbf{x}^{\prime})$ represents an \textit{atomic intervention} that temporarily fixes a set of variable(s) $\mathbf{X}^{\prime}$ to constant value(s) $\mathbf{x}^{\prime}$ irrespective of the functions $\mathbf{F}_t$ \cite[Def. 3.2.1]{pearl2000causality}. We use the $\mathrm{do}$ operator as a mathematical representation of an intervention on the system.

In such an intervention, the system is configured according to $\mathbf{X}^{\prime}=\mathbf{x}^{\prime}$ and measurement samples are collected while the system operates under these imposed conditions. We assume that the samples are collected after the system has reached a steady state under these conditions, i.e., after the transient effects of the intervention have settled. For example, suppose the intervention modifies the routing configuration. In this case, the system is in a steady state when the changes in the routing tables have propagated through the network. Once sufficient data has been gathered in the steady state, the intervention is terminated and the system variables in the set $\mathbf{X}^{\prime}$ are restored to their pre-intervention values.

The special case $\mathrm{do}(\emptyset)$ corresponds to the \textit{passive intervention} without changing any control variables, i.e., observing the system in its \textit{current operating region}, as defined next.
\begin{definition}[Current operating region]\label{def:cor}
Given a structural causal model $\mathscr{M}_t$ [cf.~(\ref{eq:scm_def})] of an \textsc{it} system, the current operating region at time $t$ is the set of feasible system configurations given the distribution $\mathbb{P}[\mathbf{U}]$ and the causal functions $\mathbf{F}_t$, i.e.,
\begin{align*}
\mathcal{O}_t &\triangleq \Big\{ (\mathbf{u}, \mathbf{v}) \in \mathcal{O} \mid \mathbb{P}[\mathbf{U}=\mathbf{u}] > 0, \mathbb{P}[\mathbf{V}=\mathbf{v}\mid \mathbf{u},\mathbf{F}_t] > 0\Big\},
\end{align*}
where $\mathcal{O}$ is the (complete) operating region; cf.~\defref{def:or}.
\end{definition}
The current operating region is generally a strict subset of the (complete) operating region, i.e., $\mathcal{O}_t \subset \mathcal{O}$. Consequently, relying solely on samples from $\mathcal{O}_t$ is generally insufficient for constructing an accurate model across the complete operating region, which is the goal of system identification.

We model \textit{online system identification} as a discrete-time process in which interventions and model updates occur at discrete points in time. Specifically, at each time $t=1,2,\hdots$, one intervention can be performed, which yields $M$ samples from one of the distributions in (\ref{eq:distributions}), where $M$ is a configurable parameter. Let $\mathcal{D}_{\mathrm{do}(\mathbf{X}^{\prime}=\mathbf{x}^{\prime}), t}$ denote the set of samples from the distribution $\mathbb{P}[\mathbf{V},\mathbf{U} \mid \mathrm{do}(\mathbf{X}^{\prime}=\mathbf{x}^{\prime})]$ from time $1$ up to time $t$. The total dataset of samples up to time $t$ is then the union
\begin{align}
\mathcal{D}_t \triangleq \bigcup_{\mathbf{X}^{\prime} \in 2^{\mathbf{X}},\mathbf{x}^{\prime} \in \mathcal{R}(\mathbf{X}^{\prime})} \mathcal{D}_{\mathrm{do}(\mathbf{X}^{\prime}=\mathbf{x}^{\prime}), t}. \label{eq:dataset}
\end{align}
Given the evolving dataset $\mathcal{D}_t$, our goal is to sequentially estimate the causal functions $\mathbf{F}_t$; cf.~(\ref{eq:scm_def}). That is, we aim to estimate a sequence of functions $\hat{\mathbf{F}}_1, \hat{\mathbf{F}}_2, \hdots$ that are as close as possible to the true sequence $\mathbf{F}_1, \mathbf{F}_2, \hdots$ while minimizing the cost of interventions. We formalize this objective as
\begin{align}
\minimize_{(\mathrm{do}(\mathbf{X}_t^{\prime}=\mathbf{x}_t^{\prime}),\hat{\mathbf{F}}_t)_{t\geq 1}} \sum_{t=1}^{\infty}\gamma^{t-1}\bigg(\mathscr{L}(\hat{\mathbf{F}}_t, \mathbf{F}_t) + c\big(\mathrm{do}(\mathbf{X}_t^{\prime}=\mathbf{x}_t^{\prime})\big)\bigg), \label{eq:objective}
\end{align}
where $\gamma \in (0,1)$ is a discount factor, $\mathrm{do}(\mathbf{X}_t^{\prime}=\mathbf{x}_t^{\prime})$ is the intervention at time $t$, $c$ is a cost function that encodes the cost of interventions, and $\mathscr{L}$ is a loss function that quantifies the accuracy of the estimated causal functions. This bi-objective captures a trade-off between estimation accuracy and intervention cost, which can be controlled by tuning the cost function $c$ and the loss function $\mathscr{L}$. In this paper, we consider the cost function $c$ to be specific to the application use case, and we define the loss function $\mathscr{L}$ as
\begin{equation}
\mathscr{L}(\hat{\mathbf{F}}_t, \mathbf{F}_t) \triangleq \!\!\sum_{V_i \in \mathbf{V}} \!\! \int_{\mathcal{R}(\mathrm{pa}_{\mathcal{G}}(V_i))} \!\! \left(f_{V_i,t}(\mathbf{x}) - \hat{f}_{V_i,t}(\mathbf{x})\right)^2\!\! \mathbb{P}[\mathrm{d}\mathbf{x}], \label{eq:loss_fun}
\end{equation}
where $\mathrm{pa}_{\mathcal{G}}(V_i)$ is the set of parents of the variable $V_i$ in the graph $\mathcal{G}$, $\hat{f}_{V_i,t} \in \hat{\mathbf{F}}_t$ is the estimated causal function of $V_i$, and $\mathbb{P}[\mathrm{d}\mathbf{x}]$ represents the probability measure with respect to which the integral is calculated. For example, consider the variables $X$ and $U$ in \figref{fig:causal_graphs_intro}. Suppose the distribution of the exogenous variable $U$ admits a probability density function $p(u)$. Then the integral related to the endogenous variable $X$ becomes
$$\int_{\mathcal{R}(U)}\left(f_{X,t}(u)-\hat{f}_{X,t}(u)\right)^2p(u)\mathrm{d}u.$$

The loss function $\mathscr{L}$ [cf.~(\ref{eq:loss_fun})] quantifies the difference between the true causal functions $\mathbf{F}_t=\{f_{V_i,t}\}_{V_i\in \mathbf{V}}$ and the estimated causal functions $\hat{\mathbf{F}}_t = \{\hat{f}_{V_i,t}\}_{V_i\in \mathbf{V}}$, weighted by the probability distribution determined by the causal graph $\mathcal{G}$ and the distribution $\mathbb{P}[\mathbf{U}]$. (Recall that we assume that both the graph $\mathcal{G}$ and the distribution $\mathbb{P}[\mathbf{U}]$ are known.)

Given these definitions and assumptions, we formally define the online causal identification problem as follows.

\vspace{0.2mm}

\begin{problemone}
\setword{Consider}{problem} an \textsc{it} system modeled by an \textsc{scm}. The causal graph and the distribution $\mathbb{P}[\mathbf{U}]$ of this \textsc{scm} are fixed and known, but the causal functions $\mathbf{F}_t$ are unknown and may vary over time; cf.~(\ref{eq:scm_def}). The problem is to design an \textit{estimator} $\varphi(\mathcal{D}_t)$ and an \textit{intervention policy} $\pi(\mathcal{D}_t)$ for accurately tracking the causal functions $\mathbf{F}_t$, while keeping intervention costs low, as defined in (\ref{eq:objective}).
\end{problemone}

\begin{figure*}
  \centering
  \scalebox{0.77}{
   \begin{tikzpicture}

\definecolor{grad1}{RGB}{0,150,150}
\definecolor{grad2}{RGB}{0,100,200}
\definecolor{grad3}{RGB}{80,0,150}

\shade[left color=grad1!8, right color=white] (0,0) rectangle (5,-4);
\shade[left color=grad2!8, right color=white] (7,0) rectangle (12,-4);
\shade[left color=grad3!5, right color=white] (14,0) rectangle (19,-4);

\fill[grad1!60!black!80] (0,0) rectangle (5,0.8);
\fill[grad2!60!black!80] (7,0) rectangle (12,0.8);
\fill[grad3!60!black!80] (14,0) rectangle (19,0.8);

\node at (2.5,0.4) {\textbf{\textcolor{white}{Model structure}}};
\node at (9.5,0.4) {\textbf{\textcolor{white}{Model estimation}}};
\node at (16.5,0.4) {\textbf{\textcolor{white}{Active learning}}};

\draw[thick] (0,0.8) rectangle (5,-4);
\draw[thick] (7,0.8) rectangle (12,-4);
\draw[thick] (14,0.8) rectangle (19,-4);

\node[scale=1.25] (gi) at (2.5,-1.4)
{
  \begin{tikzpicture}
\node[draw,circle, minimum width=0.8cm, scale=0.7](x) at (0,0) {};
\node[draw,circle, minimum width=0.8cm, scale=0.7](z) at (1.25,0) {};
\node[draw,circle, minimum width=0.8cm, scale=0.7](y) at (2.5,0) {};
\node[draw,circle, minimum width=0.8cm, scale=0.7, fill=black!12](u) at (1.25,1.1) {};
\draw[-{Latex[length=2mm]}, line width=0.22mm, color=black] (x) to (z);
\draw[-{Latex[length=2mm]}, line width=0.22mm, color=black] (z) to (y);
\draw[-{Latex[length=2mm]}, line width=0.22mm, color=black,dashed] (u) to (y);
\draw[-{Latex[length=2mm]}, line width=0.22mm, color=black,dashed] (u) to (x);
\node[inner sep=0pt,align=center, scale=0.8] (dots4) at (1.25,1.1) {$U$};
\node[inner sep=0pt,align=center, scale=0.8] (dots4) at (0,0) {$X$};
\node[inner sep=0pt,align=center, scale=0.8] (dots4) at (1.25,0) {$Z$};
\node[inner sep=0pt,align=center, scale=0.8] (dots4) at (2.5,0) {$Y$};
\end{tikzpicture}
};

\node[inner sep=0pt,align=center, scale=0.95, color=black] (hacker) at (2.5,-3.2) {
  The causal graph $\mathcal{G}$\\ encodes structural relationships\\ among system components.
};

\node[inner sep=0pt,align=center, scale=0.95, color=black] (hacker) at (9.5,-3) {
  Estimate the functions $\mathbf{F}_t$ [cf. (\ref{eq:scm_def})]\\via Gaussian process regression,\\as defined in (\ref{eq:estimator}).
};

\node[scale=1.45] (gi) at (16.5,-2)
{
\begin{tikzpicture}
  \node[scale=0.85, rotate=0](test) at (-0.35,-1.95) {
    \begin{tikzpicture}
\node[scale=0.5, rotate=10](c1) at (2.13,0.72) {
    \begin{tikzpicture}
\draw[->, x=0.15cm,y=0.15cm, line width=0.7mm, Black!40!Red!90, dashed]
        (3,0) sin (5,1) cos (7,0) sin (9,-1) cos (11,0)
        sin (13,1) cos (15,0) sin (17,-1)
        ;
    \end{tikzpicture}
};

\node[scale=0.5, rotate=2](c1) at (2.17,0.5) {
    \begin{tikzpicture}
\draw[->, x=0.15cm,y=0.15cm, line width=0.7mm, Black!40!Red!90, dashed]
        (3,0) sin (5,1) cos (7,0) sin (9,-1) cos (11,0)
        sin (13,1) cos (15,0) sin (17,-1)
        ;
    \end{tikzpicture}
  };

\node[scale=0.5, rotate=-10](c1) at (2.17,0.3) {
    \begin{tikzpicture}
\draw[->, x=0.15cm,y=0.15cm, line width=0.7mm, Black!40!Red!90, dashed]
        (3,0) sin (5,1) cos (7,0) sin (9,-1) cos (11,0)
        sin (13,1) cos (15,0) sin (17,-1)
        ;
    \end{tikzpicture}
  };
  \node[draw,circle, minimum width=1mm, scale=0.4, fill=black](start) at (1.6,0.55) {};
\node[scale=0.6](c1) at (2.62,0.5) {
  \begin{tikzpicture}
\draw (3,0) ellipse (0.25cm and 0.85cm);
    \end{tikzpicture}
  };
\end{tikzpicture}
};

\node[inner sep=0pt,align=center, scale=0.65, color=black] (hacker) at (-0.25,-1.15) {
  (\textit{i}) Evaluate interventions\\
  through rollouts; cf. (\ref{eq:rollout_approximation}).
};

\node[inner sep=0pt,align=center, scale=0.65, color=black] (hacker) at (-0.75,-2.1) {
$\mathbf{b}_t$
};

\node[inner sep=0pt,align=center, scale=0.65, color=black] (hacker) at (0.4,-1.95) {
$\tilde{J}$
};
\node[inner sep=0pt,align=center, scale=0.65, color=black] (hacker) at (-0.25,-2.87) {
  (\textit{ii}) Select an informative\\
 intervention based on\\lookahead optimization; cf. (\ref{eq:rollout}).
};

\end{tikzpicture}
};

\node[draw=none] at (9.5,-1.2) {
\begin{tikzpicture}[scale=0.7]

        \fill[blue!20, opacity=0.5]
            plot[smooth, samples=100, domain=0:6]
            (\x, {0.5*sin(deg(\x)) + 0.7})
            -- plot[smooth, samples=100, domain=6:0]
            (\x, {0.5*sin(deg(\x)) - 0.7})
            -- cycle;

        \draw[thick, black]
            plot[smooth, samples=100, domain=0:6]
            (\x, {0.5*sin(deg(\x))});

        \draw[dashed, thick]
            plot[smooth, samples=100, domain=0:6]
            (\x, {0.8*sin(deg(\x + 0.5)) + 0.4});

        \foreach \x/\y in {0.5/0.2, 1.2/0.8, 2.0/0.4, 3.0/-0.6, 4.0/-0.8, 5.2/0.1} {
            \fill[red] (\x, \y) circle (3pt);
        }

    \end{tikzpicture}
  };

\draw[-{Latex[length=2.5mm]}, line width=0.4mm, color=black] (5,-2) to (6.9,-2);
\node[inner sep=0pt,align=center, scale=1, color=black] (hacker) at (6,-2.5) {
\textit{Candidate}\\\textit{models}
};

\draw[-{Latex[length=2.5mm]}, line width=0.4mm, color=black] (12,-2) to (13.9,-2);
\node[inner sep=0pt,align=center, scale=1, color=black] (hacker) at (13,-2.8) {
\textit{Model}\\\textit{distribution}\\
$\mathbb{P}[\mathbf{F}_t \mid \mathcal{D}_t]$.
};

\draw[-{Latex[length=2.5mm]}, line width=0.4mm, color=black] (19,-2) to (19.7,-2);
\draw[-{Latex[length=2.5mm]}, line width=0.4mm, color=black] (20.8,-2.5) to (20.8, -5) to (11, -5);


\draw[-{Latex[length=2.5mm]}, line width=0.4mm, color=black] (2.5,-4.8) to (2.5,-4);
\node[inner sep=0pt,align=center, scale=1, color=black] (hacker) at (2.5,-5) {
\textit{System configuration}
};
\draw[-{Latex[length=2.5mm]}, line width=0.4mm, color=black] (9.5,-4.8) to (9.5,-4);
\node[inner sep=0pt,align=center, scale=1, color=black] (hacker) at (9.5,-5) {
\textit{Dataset $\mathcal{D}_t$ [cf.~(\ref{eq:dataset})]}
};

\node[inner sep=0pt,align=center, scale=1, color=black] (hacker) at (16.1,-4.7) {
\textit{Measurement samples of the system variables $\mathbf{V}, \mathbf{U}$ [cf.~(\ref{eq:distributions})]}
};

\node[inner sep=0pt,align=center, scale=1, color=black] (hacker) at (20.7,-2) {
\textit{Intervention}\\
$\mathrm{do}(\mathbf{X}^{\prime}=\mathbf{x}^{\prime})$
};

\end{tikzpicture}        
  }
  \caption{Our iterative method for online identification of a causal model of an \textsc{it} system. Such a model consists of a set of causal functions $\mathbf{F}_t$; cf.~(\ref{eq:scm_def}). The set of candidate models is defined by a causal graph that encodes structural properties of the system. During an iteration, we fit a distribution over this set using system measurements and Gaussian processes. This distribution then guides the selection of the next intervention aimed at refining the distribution while keeping the intervention cost low. We implement this selection using rollout and lookahead optimization.}
    \label{fig:framework}
\end{figure*}
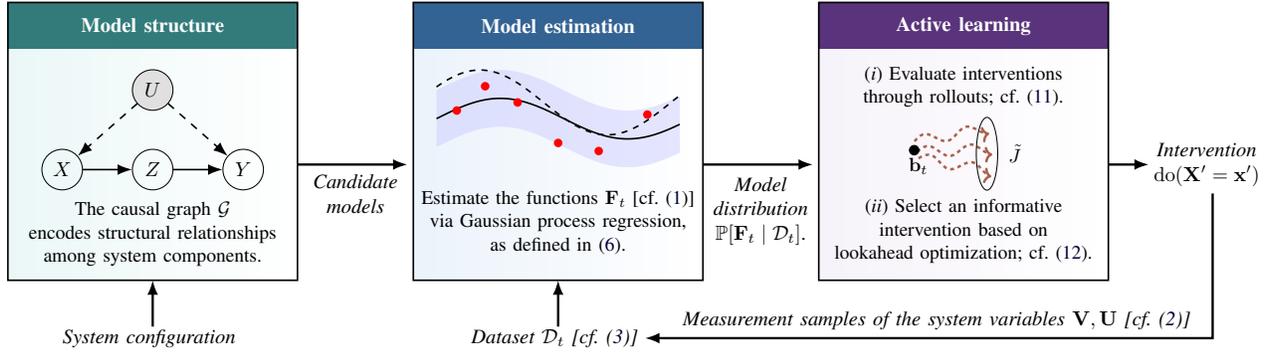
\section{Our Method for Online System Identification}
Our method for solving the \hyperref[problem]{problem} in the preceding section consists of two parts: (\textit{i}) Bayesian learning of the causal functions $\mathbf{F}_t$ [cf.~(\ref{eq:scm_def})] via Gaussian process regression; and (\textit{ii}) selection of interventions for collecting measurement data via rollout and lookahead optimization; see \figref{fig:framework}.

\subsection{Model Estimation through Gaussian Process Regression}\label{sec:gps_reg}
In our method, we use Gaussian process (\textsc{gp}) regression to estimate the unknown causal functions of the target system \cite[Def. 2.1]{Rasmussen2006Gaussian}. Although the functions in $\mathbf{F}_t$ are deterministic at each time $t$, we represent our uncertainty about them through a probability distribution over possible functions. Specifically, we place an independent \textsc{gp} prior on each of the causal functions $f_{V_i,t}$ in $\mathbf{F}_t$; cf.~(\ref{eq:scm_def}). That is, before collecting any measurement data, we express our uncertainty about the causal function $f_{V_i,t}$ for every endogenous variable $V_{i} \in \mathbf{V}$ through the \textsc{gp} $\hat{f}_{V_i,t} \sim \mathcal{GP}(m_i, k_i)$, where $m_i(\mathbf{x}_i)$ is the mean function and $k_i(\mathbf{x}_i,\mathbf{x}_i')$ is the covariance function. Here $\hat{f}_{V_i,t}(\mathbf{x}_i)$ is the estimated value of $V_i$ given that its parents in the causal graph $\mathcal{G}$ take on the values in the vector $\mathbf{x}_i$. Similarly, the variance $k_i(\mathbf{x}_i,\mathbf{x}_i)$ represents the uncertainty about the estimated function $\hat{f}_{V_i,t}$ at the input vector $\mathbf{x}_i$. Finally, $k_i(\mathbf{x}_i,\mathbf{x}_i^{\prime})$ encodes the correlation between the function values at two different inputs: $\mathbf{x}_i$ and $\mathbf{x}_i^{\prime}$.

As the dataset $\mathcal{D}_t$ [cf.~(\ref{eq:dataset})] is updated over time with new measurements, we update the \textsc{gp} prior of each causal function $\hat{f}_{V_i,t}$ via Bayes' rule; see \appendixref{appendix:gps} for detailed formulas of these updates. We denote the resulting posterior as $\hat{f}_{V_i,t} \mid \mathcal{D}_t \sim \mathcal{GP}(m_{i|\mathcal{D}_t}, k_{i|\mathcal{D}_t})$, where $m_{i\mid \mathcal{D}_t}$ and $k_{i\mid\mathcal{D}_t}$ denote the posterior mean and covariance functions given the dataset $\mathcal{D}_t$. This posterior allows us to predict $f_{V_i,t}(\mathbf{x}_i)$ using the mean $m_{i|\mathcal{D}_t}(\mathbf{x}_i)$, whose uncertainty is quantified by the variance $k_{i|\mathcal{D}_t}(\mathbf{x}_i, \mathbf{x}_i)$. Since we can make such predictions for any causal function $f_{V_i,t} \in \mathbf{F}_t$ and input $\mathbf{x}_i \in \mathcal{R}(\mathrm{pa}_{\mathcal{G}}(V_i))$, the collection of all the posterior \textsc{gp}s $\{(\hat{f}_{V_i,t} | \mathcal{D}_t) \mid V_i \in \mathbf{V}\}$ allows to estimate the causal functions $\mathbf{F}_t$; cf.~(\ref{eq:scm_def}). In particular, since the \textsc{gp}s are independent, we can construct a probability distribution over the causal functions $\mathbf{F}_t$ according to
\begin{equation}
\begin{split}
&\varphi(\mathcal{D}_t) \triangleq \mathbb{P}[\mathbf{F}_t \mid \mathcal{D}_t] = \prod_{V_i \in \mathbf{V}}\mathbb{P}[f_{V_i,t} \mid \mathcal{D}_t], \label{eq:estimator}
\end{split}
\end{equation}
where each distribution $\mathbb{P}[f_{V_i,t} | \mathcal{D}_t]$ is represented by a \textsc{gp} $\mathcal{GP}(m_{i \mid \mathcal{D}_t}, k_{i \mid \mathcal{D}_t})$. This distribution quantifies the uncertainty about the causal functions $\mathbf{F}_t$ based on the dataset $\mathcal{D}_t$. To obtain a point estimate of the functions, we take the expectation with respect to this distribution, i.e., $\hat{\mathbf{F}}_t = \mathbb{E}_{\mathbf{F}_t \sim \varphi(\mathcal{D}_t)}\{\mathbf{F}_t\}$. This estimate of $\mathbf{F}_t$ is optimal in the following sense.
\begin{proposition}\label{prop:bayes_optimal}
The expectation $\mathbb{E}_{\mathbf{F}_t \sim \varphi(\mathcal{D}_t)}\{\mathbf{F}_t\}$ minimizes the expected value of the loss function $\mathscr{L}$ [cf.~(\ref{eq:loss_fun})], i.e.,
\begin{align*}
\mathbb{E}_{\mathbf{F}_t \sim \varphi(\mathcal{D}_t)}\{\mathbf{F}_t\} \in \argmin_{\hat{\mathbf{F}}_t}\mathbb{E}_{\mathbf{F}_t \sim \varphi(\mathcal{D}_t)}\left\{\mathscr{L}(\hat{\mathbf{F}}_t, \mathbf{F}_t)\right\},
\end{align*}
where the minimization is over all sets of causal functions $\hat{\mathbf{F}}_t$ compatible with the causal graph $\mathcal{G}$.
\end{proposition}
We present the proof of \propref{prop:bayes_optimal} in \appendixref{appendix:proof_prop1}. This proposition expresses that the expectation $\mathbb{E}_{\mathbf{F}_t \sim \varphi(\mathcal{D}_t)}\{\mathbf{F}_t\}$ based on the estimator $\varphi$ [cf.~(\ref{eq:estimator})] is Bayes-optimal. That is, among all \textsc{scm}s that respect the structure encoded in the causal graph $\mathcal{G}$, this expectation yields the lowest expected value of the loss function $\mathscr{L}$; cf.~(\ref{eq:loss_fun}). In other words, it is the best prediction we can make for the causal functions $\mathbf{F}_t$ given the data up to time $t$. In addition to the Bayes-optimality, the expectation $\mathbb{E}_{\mathbf{F}_t \sim \varphi(\mathcal{D}_t)}\{\mathbf{F}_t\}$ converges to the true causal functions given sufficient data and regularity, as stated below.
\begin{proposition}\label{prop:consistency}
Assume a) that the causal functions $\mathbf{F}$ [cf.~(\ref{eq:scm_def})] are fixed; and b) that the difference between each causal function $f_{V_i}(\mathbf{x}_i)$ and the (prior) mean function $m_i$ lies in the reproducing kernel Hilbert space of the covariance function $k_i$. If each input $\mathbf{x}_i \in \mathcal{R}(\mathrm{pa}_{\mathcal{G}}(V_i))$ is sampled infinitely often with independent zero-mean Gaussian noise, then
\begin{align*}
\lim_{|\mathcal{D}_t|\to \infty}\mathbb{E}\left\{\left(\hat{f}_{V_i}(\mathbf{x}_i)-f_{V_i}(\mathbf{x}_i)\right)^2\right\} = 0,
\end{align*}
where $\hat{f}_{V_i} = \mathbb{E}_{\mathbf{F}_t \sim \varphi(\mathcal{D}_t)}\{f_{V_i}\}$ [cf.~(\ref{eq:estimator})] and the expectation is with respect to the sampling noise and variability. 
\end{proposition}
\Propref{prop:consistency} implies that the estimator $\varphi$ [cf.~(\ref{eq:estimator})] recovers the true causal functions in the limit of infinite data, given certain regularity conditions. In other words, the estimator $\varphi$ is \textit{consistent}. This is a well-known result in \textsc{gp} theory, see e.g., \cite[Thm. 3]{JMLR:v22:21-0853} for a detailed analysis and proof.

\begin{remark}
While the \textsc{gp} estimator $\varphi$ [cf.~(\ref{eq:estimator})] is designed for continuous functions, it can also estimate functions over discrete domains by adapting the covariance function and restricting predictions to the discrete set; see \cite{discrete_gp} for details.
\end{remark}
\subsection{Active Learning through Rollout}
Given the estimator $\varphi$ of the causal functions $\mathbf{F}_t$ [cf.~(\ref{eq:estimator})], the problem of active learning is to select a sequence of interventions that generate samples from the distributions (\ref{eq:distributions}) to update the dataset $\mathcal{D}_t$ and achieve the objective (\ref{eq:objective}). This problem can be formulated as a dynamic programming problem where the (belief) state is $\mathbf{b}_t \triangleq \varphi(\mathcal{D}_t)$, the control at time $t$ is the intervention $u_t \triangleq \mathrm{do}(\mathbf{X}_t^{\prime} = \mathbf{x}_t^{\prime})$, the intervention policy is $\pi(\mathbf{b}_t)$, and the dynamics are defined as
\begin{align}
\mathbf{b}_{t+1} \triangleq \varphi(\mathcal{D}_t \cup \{\mathbf{z}_t\}), && \text{for all }t\geq 1, \label{eq:belief_dynamics}
\end{align}
where $\mathbf{z}_t$ is the measurement obtained after the intervention $u_t$; cf.~(\ref{eq:distributions}). The goal when selecting interventions is to improve the accuracy of the estimated functions $\hat{\mathbf{F}}_t$ while keeping the intervention costs $c(u)$ low; cf.~(\ref{eq:objective}). The accuracy of $\hat{\mathbf{F}}_t$ is quantified by the loss function $\mathscr{L}$, as defined in (\ref{eq:loss_fun}). However, $\mathscr{L}$ cannot be directly computed as it depends on the unknown causal functions $\mathbf{F}$. For this reason, we define a \textit{surrogate loss function} that captures the expected value of $\mathscr{L}$ given the belief state $\mathbf{b}$. We denote this function by $\mathcal{L}$ and define it as
\begin{align}
\mathcal{L}(\mathbf{b}) &\triangleq \int_{\mathcal{F}}\mathbf{b}(\mathbf{F})\bigg(\mathscr{L}(\mathbf{F}, \mathbb{E}_{\mathbf{F}_t \sim \mathbf{b}}\{\mathbf{F}_t\})\bigg)\mathrm{d}\mathbf{F},\label{eq:surrogate_loss}
\end{align}
where the integral is over the space of functions compatible with the causal graph $\mathcal{G}$. Minimizing this surrogate loss function reduces uncertainty in the belief state $\mathbf{b}$. In particular, $\mathcal{L}(\mathbf{b})=0$ if and only if $\mathbf{b}(\mathbf{F})=1$ for some set of causal functions $\mathbf{F}$. Moreover, by the consistency of the \textsc{gp} estimator $\varphi$ [cf.~\propref{prop:consistency}], this condition implies that $\mathbf{F} = \mathbf{F}_t$. Hence, selecting interventions that drive the surrogate loss function $\mathcal{L}(\mathbf{b})$ to zero is equivalent to identifying the true causal functions and thus minimizing the loss function $\mathscr{L}$; cf.~(\ref{eq:loss_fun}).

Given the surrogate loss function $\mathcal{L}$, we define the cost function $g(\mathbf{b}, u)$ of the dynamic programming problem as
\begin{equation}
\begin{split}  
g(\mathbf{b}_{t}, u_t) &\triangleq \mathbb{E}_{\mathbf{b}_{t+1}}\left\{\mathcal{L}(\mathbf{b}_{t+1})-\mathcal{L}(\mathbf{b}_{t})\mid u_t, \mathbf{b}_t\right\} + c(u_t).\label{eq:cost_dynamics}
\end{split}    
\end{equation}
This cost function quantifies the expected change in the surrogate loss $\mathcal{L}(\mathbf{b})$ [cf.~(\ref{eq:surrogate_loss})] after performing the intervention $u_t$, collecting $M$ measurement samples from the corresponding interventional distribution in~(\ref{eq:distributions}), and updating the belief state $\mathbf{b}_t$ using the \textsc{gp} estimator $\varphi$, as defined in~(\ref{eq:estimator}). Hence, the structure of the cost function $g$ aligns with the objective~(\ref{eq:objective}). Specifically, by selecting interventions that minimize the expected cost, we obtain an intervention policy that maximizes the expected reduction in uncertainty of the belief state $\mathbf{b}_t$ [as quantified by the first term in (\ref{eq:cost_dynamics})] while keeping the intervention cost low, as quantified by the second term in (\ref{eq:cost_dynamics}).

The solution to the dynamic program with the dynamics (\ref{eq:belief_dynamics}) and the cost function (\ref{eq:cost_dynamics}) yields an optimal intervention policy $\pi^{\star}$, which minimizes the following cost-to-go function.
\begin{align}
J_{\pi}(\mathbf{b}) \triangleq \lim_{T \rightarrow \infty}\mathbb{E}_{\pi}\left\{\sum_{t=1}^{T}\gamma^{t-1}g(\mathbf{b}_t, u_t) \mid \mathbf{b}_1=\mathbf{b} \right\},\label{eq:bellman_2}
\end{align}
where $\mathbb{E}_{\pi}$ denotes the expectation of $(\mathbf{b}_t)_{t \geq 2}$ when updating the dataset $\mathcal{D}_t$ [cf.~(\ref{eq:dataset})] using the intervention policy $\pi$.

While (\ref{eq:belief_dynamics}) can be efficiently computed, (\ref{eq:cost_dynamics}) involves an integral [cf. (\ref{eq:loss_fun})] that is intractable in general. Another challenge in solving this dynamic programming problem is that the dynamics (\ref{eq:belief_dynamics}) are non-stationary in case the underlying \textsc{it} system evolves. In particular, the distribution of the measurement sample $\mathbf{z}_t$ [cf.~(\ref{eq:distributions})] may become dependent on the time step $t$. For these reasons, the problem of computing an optimal intervention policy $\pi^{\star}$ is intractable in the general case.

To address this computational intractability, we approximate the cost function $g$ [cf.~(\ref{eq:cost_dynamics})] by discretizing the function space $\mathcal{F}$ [cf.~(\ref{eq:surrogate_loss})] and using Monte-Carlo sampling to estimate the expectation in (\ref{eq:cost_dynamics}). Moreover, we approximate an optimal intervention policy using \textit{rollout}, which is an online methodology for approximate dynamic programming developed by Bertsekas; see textbook \cite{bertsekas2021rollout} and paper \cite{bertsekas2022rolloutalgorithmsapproximatedynamic}. Following this methodology, at each time step $t$ of the identification, we simulate the evolution of the dataset $\mathcal{D}_t$ [cf.~(\ref{eq:dataset})] $m$ time steps into the future, whereby interventions are selected according to a \textit{base intervention policy} $\pi$. This lookahead simulation allows us to estimate the cost-to-go of the base policy as
\begin{align}
\tilde{J}_{\pi}(\mathbf{b}_{t})\!\!&=\!\!\frac{1}{L}\sum_{j=1}^{L}\sum_{l=t}^{t+m-1}\!\!\gamma^{l-t}g(\mathbf{b}^{j}_l, \pi(\mathbf{b}^{j}_l)) + \gamma^{m}\tilde{J}(\mathbf{b}^{j}_{t+m}),\label{eq:rollout_approximation}
\end{align}
where $L$ is the number of simulations and $\tilde{J}$ is a function that approximates future costs. Both this function and the base policy $\pi$ can be chosen freely, e.g., based on heuristics or offline optimization \cite{tifs_25_HLALB,hammar2025incidentresponseplanningusing,hammar_aisec25}. For example, they can be defined as $\pi(\mathbf{b})=\mathrm{do}(\emptyset)$ and $\tilde{J}(\mathbf{b})=0$ for all belief states $\mathbf{b}$.

Finally, we use the cost-to-go estimate obtained through (\ref{eq:rollout_approximation}) to transform the base policy to a \textit{rollout policy} $\tilde{\pi}$ as
\begin{equation}
\begin{split}
  &\tilde{\pi}(\mathbf{b}_t) \in \argmin_{u_t} \Bigg[g(\mathbf{b}_t,u_t) + \min_{\pi_{t+1}, \hdots, \pi_{t+\ell-1}}\mathop{\mathbb{E}}_{\mathbf{b}_{t+1},\hdots,\mathbf{b}_{t+\ell}}\bigg\{\label{eq:rollout}\\
  &\quad\quad\quad\quad\sum_{j=t+1}^{t+\ell-1}\gamma^{j-t}g(\mathbf{b}_j,\pi_j(\mathbf{b}_j)) + \gamma^{\ell}\tilde{J}_{\pi}(\mathbf{b}_{t+\ell})\bigg\}\Bigg],
\end{split}
\end{equation}
where $\ell \geq 1$ is the lookahead horizon.

The benefit of this optimization is that it is guaranteed to yield an improved intervention policy (compared to the base policy) under general conditions, as stated below.
\begin{proposition}\label{prop:bertsekas}
If the cost function $g$ [cf.~(\ref{eq:cost_dynamics})] is bounded, the estimation in (\ref{eq:rollout_approximation}) is exact (i.e., $\tilde{J}_{\pi}=J_{\pi}$), the operating region $\mathcal{O}$ [cf.~\defref{def:or}] is a compact subset of a Euclidean space, and the function space $\mathcal{F}$ [cf.~(\ref{eq:surrogate_loss})] is discretized such that the belief $\mathbf{b}$ belongs to a compact subset of a Euclidean space, then the rollout policy $\tilde{\pi}$ obtained through (\ref{eq:rollout}) is guaranteed to improve the base policy $\pi$, i.e., $J_{\tilde{\pi}}(\mathbf{b}) \leq J_{\pi}(\mathbf{b})$ for all $\mathbf{b}$.
\end{proposition}
\Propref{prop:bertsekas} provides a performance guarantee for the rollout policy [cf.~(\ref{eq:rollout})] under certain conditions. It implies that the rollout policy will generally perform at least as well, and typically better than the base policy $\pi$. The proof follows directly from standard results by Bertsekas; see e.g., \cite[Prop. 2.3.1]{bertsekas2021rollout}, \cite[Prop. 5.1.1]{bertsekas2019reinforcement}, and \cite[\S 2.4]{bertsekas2018abstract} for details.

From a computational point of view, the complexity of the minimization (\ref{eq:rollout}) can be adjusted according to available computing resources by tuning the lookahead horizon $\ell$, the rollout horizon $m$, and the number of rollouts $L$. The main computational complexity stems from evaluating the expectations in (\ref{eq:rollout}) and (\ref{eq:loss_fun}). Fortunately, these expectations can be efficiently approximated via Monte-Carlo sampling.

\begin{figure*}
  \centering
  \scalebox{0.75}{
   \includegraphics{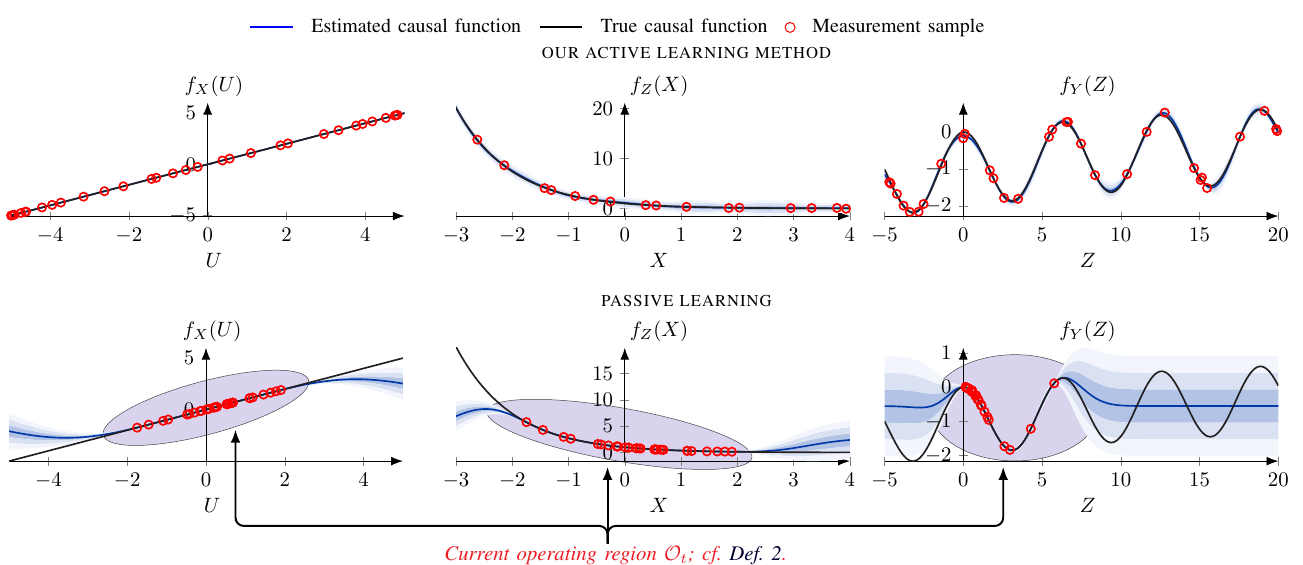}
  }
    \caption{Learned causal functions using the \textsc{gp} estimator $\varphi(\mathcal{D}_t)$ [cf.~(\ref{eq:estimator})] for the example \textsc{scm} in \sectionref{sec:example}. The functions are learned based on $30$ samples, i.e., $|\mathcal{D}_t|=30$; cf.~(\ref{eq:dataset}). The lower plots show the causal functions estimated through passive learning, i.e., the functions estimated from data collected through monitoring the system without influencing its operation through interventions. The upper plots show the functions estimated through our active learning method, i.e., the functions estimated from data collected using the rollout intervention policy; cf.~(\ref{eq:rollout}). Curves show the mean values of the \textsc{gp}s; shaded regions indicate one, two, and three standard deviations from the mean (darker to lighter shades of blue).}
    \label{fig:illustrative_example}
\end{figure*}

\vspace{1mm}

\begin{summary}
Our method for online identification of the causal model of an \textsc{it} system includes the following steps. We assume we know the causal graph $\mathcal{G}$, which encodes the structural relationships within the \textsc{it} system and which does not change over time. This graph defines a set of candidate models, each of which consists of a set of causal functions $\mathbf{F}_t$; cf.~(\ref{eq:scm_def}). Starting at time $t=1$ with initial dataset $\mathcal{D}_1=\emptyset$, we repeat the following:
\begin{enumerate}
\item Estimate the causal functions $\mathbf{F}_t$ [cf.~(\ref{eq:scm_def})] based on the causal graph $\mathcal{G}$ and the current dataset $\mathcal{D}_t$ [cf.~(\ref{eq:dataset})] using the \textsc{gp} estimator $\varphi(\mathcal{D}_t)$; cf.~(\ref{eq:estimator}).
\item Select the next intervention $\mathrm{do}(\mathbf{X}_{t}^{\prime}=\mathbf{x}_{t}^{\prime})$ using the rollout policy $\tilde{\pi}(\varphi(\mathcal{D}_t))$; cf.~(\ref{eq:rollout}).
\item Perform the selected intervention, sample system measurements under the conditions imposed by the intervention according to (\ref{eq:distributions}), update the dataset $\mathcal{D}_t$ [cf.~(\ref{eq:dataset})] to obtain $\mathcal{D}_{t+1}$, and restore the system to the settings before the intervention.
\end{enumerate}
\end{summary}
\section{Illustrative Example}\label{sec:example}
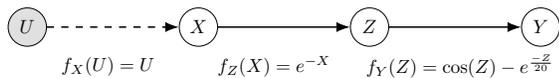
\begin{figure}[H]
  \centering
  \scalebox{1.3}{
   \begin{tikzpicture}[fill=white, >=stealth,
    node distance=3cm,
    database/.style={
      cylinder,
      cylinder uses custom fill,
      shape border rotate=90,
      aspect=0.25,
      draw}]

    \tikzset{
node distance = 9em and 4em,
sloped,
   box/.style = {%
    shape=rectangle,
    rounded corners,
    draw=blue!40,
    fill=blue!15,
    align=center,
    font=\fontsize{12}{12}\selectfont},
 arrow/.style = {%
    line width=0.1mm,
    -{Triangle[length=5mm,width=2mm]},
    shorten >=1mm, shorten <=1mm,
    font=\fontsize{8}{8}\selectfont},
}

\node[scale=0.7] (gi) at (0,0)
{
  \begin{tikzpicture}
\node[draw,circle, minimum width=0.8cm, scale=0.7](x) at (0,0) {};
\node[draw,circle, minimum width=0.8cm, scale=0.7](z) at (2.5,0) {};
\node[draw,circle, minimum width=0.8cm, scale=0.7](y) at (5,0) {};
\node[draw,circle, minimum width=0.8cm, scale=0.7, fill=black!12](u) at (-2.5,0) {};
\draw[-{Latex[length=2mm]}, line width=0.22mm, color=black] (x) to (z);
\draw[-{Latex[length=2mm]}, line width=0.22mm, color=black] (z) to (y);
\draw[-{Latex[length=2mm]}, line width=0.22mm, color=black,dashed] (u) to (x);
\node[inner sep=0pt,align=center, scale=0.8] (dots4) at (-2.5,0) {$U$};
\node[inner sep=0pt,align=center, scale=0.8] (dots4) at (0,0) {$X$};
\node[inner sep=0pt,align=center, scale=0.8] (dots4) at (2.5,0) {$Z$};
\node[inner sep=0pt,align=center, scale=0.8] (dots4) at (5,0) {$Y$};
\end{tikzpicture}
};

\node[inner sep=0pt,align=center, scale=0.55] (dots4) at (-0.1,-0.4) {$f_{Z}(X)=e^{-X}$};
\node[inner sep=0pt,align=center, scale=0.55] (dots4) at (-1.8,-0.4) {$f_{X}(U)=U$};
\node[inner sep=0pt,align=center, scale=0.55] (dots4) at (1.8,-0.4) {$f_Y(Z) = \cos(Z) - e^{\frac{-Z}{20}}$};

\end{tikzpicture}       
  }
  \caption{The causal graph and functions of the \textsc{scm} in the illustrative example.}
  \label{fig:causal_graphs_example}
\end{figure}

To illustrate our method, we apply it to an \textsc{scm} with the causal graph and functions shown in \figref{fig:causal_graphs_example}. The \textsc{scm} is fixed over time and the causal functions are $\mathbf{F}=\{f_X,f_Z,f_Y\}$.
The \textsc{scm} has four variables: one exogenous variable $\mathbf{U}=\{U\}$ and three endogenous variables $\mathbf{V}=\{X,Z,Y\}$. The exogenous variable $U$ follows a Gaussian distribution, $U \sim \mathcal{N}(0, 0.1)$. The operating region [cf.~\defref{def:or}] is given by
\begin{align*}
\mathcal{R}(X)\!\!=\!\!\mathcal{R}(Y)\!\!=\!\![-5, 5], \text{ }\mathcal{R}(Z)= [-5, 20],\text{ }\mathcal{R}(U)= [-\infty, \infty].
\end{align*}
The measurement distributions (\ref{eq:distributions}) can be sampled with additive Gaussian noise $\alpha \sim \mathcal{N}(0, 0.05)$. All variables are controllable (i.e., $\mathbf{X}=\mathbf{V}\cup\mathbf{U}$) and the cost of each intervention except the passive intervention $\mathrm{do}(\emptyset)$ is $1$; cf.~(\ref{eq:cost_dynamics}).

\vspace{2mm}

\noindent\textbf{\textit{Instantiation of our method}.} We collect $M=1$ samples per intervention. We define the cost approximation in (\ref{eq:rollout_approximation}) as $\Tilde{J}(\mathbf{b})=\mathcal{L}(\mathbf{b})$; cf.~(\ref{eq:cost_dynamics}). We configure the base policy $\pi$ [cf.~(\ref{eq:rollout})] to always select the passive intervention $\mathrm{do}(\emptyset)$. We set the lookahead and rollout horizons in (\ref{eq:rollout_approximation})--(\ref{eq:rollout}) as $\ell=1$ and $m=5$, respectively. Finally, we define all \textsc{gp}s [cf.~(\ref{eq:estimator})] to have mean and covariance functions defined as
\begin{equation}\label{eq:gp_prior_def}
\begin{split}
m(\mathbf{x}) &\triangleq 0,\\
k(\mathbf{x}, \mathbf{x}') &\triangleq \left(1 + \sqrt{5}r + \frac{5r^2}{3}\right)\exp\left(-\sqrt{5}r\right),
\end{split}  
\end{equation}
for all input vectors $\mathbf{x}$ and $\mathbf{x}^{\prime}$, where $r \triangleq \norm{\mathbf{x}-\mathbf{x}^{\prime}}_{2}$ and $\norm{\cdot}_{2}$ denotes the Euclidean norm. This covariance function encodes the assumption that the causal functions vary smoothly over the input space. Similarly, the mean function reflects the absence of prior knowledge of the function values. Further details about our experimental setup can be found in \appendixref{appendix:hyperparameters}.

Note that a broad variety of mean and covariance functions can be used to instantiate the \textsc{gp} estimator [cf.~(\ref{eq:estimator})]; see textbook \cite{Rasmussen2006Gaussian} for details. Their design and parameterization offer a principled way to incorporate domain knowledge and structure (e.g., expected smoothness of the causal functions).

\vspace{2mm}

\noindent\textbf{\textit{Baseline method}.} We compare the performance of our method with that of a baseline method that uses the same \textsc{gp} estimator [cf.~(\ref{eq:estimator})] but monitors the system without interventions. In other words, it uses the intervention policy $\pi(\mathbf{b})=\mathrm{do}(\emptyset)$ for all beliefs $\mathbf{b}$. We refer to this baseline as \textsc{passive learning}.

\vspace{2mm}

\noindent\textbf{\textit{Evaluation results}.}
\Figref{fig:illustrative_example} shows the causal functions estimated based on $30$ samples collected through our active learning method and through passive learning. We observe that our method (upper plots) yields more accurate estimates of the true causal functions compared to passive learning (lower plots), especially for the nonlinear functions $f_Z(X)$ and $f_Y(Z)$. In particular, the \textsc{gp} posterior (blue curves) obtained through our method [cf.~(\ref{eq:estimator})] closely follows the causal functions (black lines) with low uncertainty (narrow shaded regions). In contrast, the \textsc{gp}s estimated through passive learning deviate significantly from the causal functions, particularly outside the system's current operating region [cf.~\defref{def:cor}], i.e., configurations of the system variables $(X,Z,Y)$ that occur with low probability under the distribution $\mathbb{P}[U]$.

\Figref{fig:example_loss} shows the value of the loss function $\mathscr{L}$ [cf.~(\ref{eq:loss_fun})], which we approximate through discretization. We observe that as the number of samples $|\mathcal{D}_t|$ increases, the loss of the model estimated through our method (red curve) decreases rapidly. In contrast, the loss of the model estimated through passive learning (blue curve) decreases only slightly (from around $2400$ to $2340$), which is difficult to discern in the figure.
\begin{figure}
  \centering
  \scalebox{0.86}{
    \begin{tikzpicture}

\pgfplotsset{/dummy/workaround/.style={/pgfplots/axis on top}}

\pgfplotstableread{
1 2380.542554165628 2437.1227004212756 2323.962407909981
2 2377.262517769771 2442.9698131609516 2311.5552223785903
3 2375.6897682877943 2446.2011087053966 2305.178427870192
4 2375.1048040658666 2447.35362610362 2302.855982028113
5 2374.9948514772586 2446.903512049806 2303.086190904711
6 2375.4879679112664 2445.4404199358382 2305.5355158866946
7 2375.646209583208 2442.6963349560983 2308.5960842103177
8 2374.932741059622 2440.095491876553 2309.769990242691
9 2374.5998035484963 2438.385847951081 2310.8137591459117
10 2374.2927636098634 2437.1481870094153 2311.4373402103115
11 2373.80474442827 2435.8949249135308 2311.7145639430096
12 2373.1367900208706 2434.5659018766123 2311.707678165129
13 2372.3686356827943 2433.5220564752703 2311.2152148903183
14 2371.4567247915475 2432.228539115606 2310.684910467489
15 2370.717053055111 2431.0559966385786 2310.3781094716437
16 2369.9300903147114 2429.959112656609 2309.901067972814
17 2369.1241398049256 2428.872227421261 2309.37605218859
18 2368.8670869238344 2428.0068864574146 2309.7272873902543
19 2368.945849366901 2427.3949794510677 2310.496719282734
20 2368.6766727351605 2426.6855658440622 2310.667779626259
21 2368.3497794432265 2425.838750482074 2310.860808404379
22 2368.3693533728388 2425.2448642050053 2311.493842540672
23 2368.137841082199 2424.43455086906 2311.841131295338
24 2367.890481381161 2423.575174699252 2312.20578806307
25 2367.607354472174 2422.7570280727314 2312.4576808716165
26 2367.1444751534377 2421.48713809371 2312.8018122131652
27 2360.814560194852 2409.5782924611003 2312.050827928604
28 2354.9086021930584 2400.54102769268 2309.276176693437
29 2349.837636675421 2395.2979702896387 2304.3773030612033
30 2345.1753427125664 2392.4385753764823 2297.9121100486504
}\passive

\pgfplotstableread{
1 2234.903249813632 2509.006853411602 1960.799646215662
2 1576.1324419715725 1723.9582083246885 1428.3066756184564
3 1226.932000661402 1413.4876407468082 1040.376360575996
4 994.6055297471797 1143.139340013803 846.0717194805566
5 827.7200583358426 939.3270661461511 716.1130505255342
6 705.9379337780978 795.8318002482796 616.044067307916
7 610.9724210710143 686.3546852548359 535.5901568871927
8 538.4759933999856 604.091890084615 472.8600967153562
9 480.7917506688424 539.4400141035471 422.14348723413764
10 434.35924492777195 487.16237695167115 381.55611290387276
11 396.30959457040797 444.22784183860483 348.3913473022111
12 364.1422173137647 408.1741180800936 320.11031654743584
13 336.81795304694185 377.492264954025 296.1436411398587
14 313.11142969350806 350.95992955857406 275.26292982844205
15 292.48309861269047 327.897406810208 257.06879041517294
16 274.427187713406 307.74950292770694 241.10487249910506
17 258.4662889678043 289.91950474214707 227.01307319346154
18 244.24589868069262 274.01239831757476 214.47939904381047
19 231.51737814443376 259.770525854456 203.2642304344115
20 220.0510886617435 246.93826870039186 193.16390862309515
21 209.6664584290851 235.311972900863 184.02094395730722
22 200.21713514880184 224.72464201655902 175.70962828104467
23 191.60504406844916 215.03903632522773 168.1710518116706
24 183.69956708684856 206.15175900585015 161.24737516784697
25 176.4105118326043 197.97746959161674 154.84355407359186
26 169.6770949905303 190.4258933660344 148.9282966150262
27 163.44222775656388 183.43336153481576 143.451093978312
28 157.6522594174097 176.93973726129124 138.36478157352818
29 152.2610266227872 170.8932903427234 133.628762902851
30 147.22851954219777 165.24822163286564 129.2088174515299
}\activee

\node[scale=1] (kth_cr) at (0,0)
{
\begin{tikzpicture}
  \begin{axis}
[
        xmin=1,
        xmax=31,
        ymax=3000,
        ymin=100,
        width=10.3cm,
        height=3.5cm,
        axis y line=center,
        axis x line=bottom,
        scaled y ticks=false,
        xlabel style={below right},
        ylabel style={above left},
        axis line style={-{Latex[length=2mm]}},
        smooth,
        legend style={at={(0.95,0.685)}},
        legend columns=1,
        legend style={
          align=left,
          /tikz/every node/.style={anchor=west},
          draw=none,
            /tikz/column 2/.style={
                column sep=5pt,
              }
              }
              ]
              \addplot[RoyalAzure,name path=l1, thick, dashed] table [x index=0, y index=1, domain=0:1] {\passive};
              \addplot[Red,name path=l1, thick] table [x index=0, y index=1, domain=0:1] {\activee};
\legend{\textsc{passive learning}, \textsc{our active learning method}}

              \addplot[draw=none,Black,mark repeat=2, name path=A, thick, domain=0:1] table [x index=0, y index=2] {\activee};
              \addplot[draw=none,Black,mark repeat=2, name path=B, thick, domain=0:1] table [x index=0, y index=3] {\activee};
              \addplot[Red!30!white] fill between [of=A and B];

              \addplot[draw=none,Black,mark repeat=2, name path=A, thick, domain=0:1] table [x index=0, y index=2] {\passive};
              \addplot[draw=none,Black,mark repeat=2, name path=B, thick, domain=0:1] table [x index=0, y index=3] {\passive};
              \addplot[RoyalAzure!30!white] fill between [of=A and B];
  \end{axis}
\node[inner sep=0pt,align=center, scale=1, rotate=0, opacity=1] (obs) at (4.68,-0.75)
{
  Number of samples $|\mathcal{D}_t|$ [cf. (\ref{eq:dataset})]
};
\node[inner sep=0pt,align=center, scale=1, rotate=0, opacity=1] (obs) at (2.7,2.15)
{
  Loss $\mathscr{L}(\hat{\mathbf{F}}_t, \mathbf{F}_t)$ [cf. (\ref{eq:loss_fun})] ($\downarrow$ better)
};
\end{tikzpicture}
};

\end{tikzpicture}       
  }
  \caption{Loss [cf.~(\ref{eq:loss_fun})] of the functions $\hat{\mathbf{F}}_t$ estimated through passive learning (blue curve) and our active learning method (red curve) for the example \textsc{scm} in \sectionref{sec:example}. Curves show the mean value from evaluations with $5$ random seeds; shaded areas indicate standard deviations.}
  \label{fig:example_loss}
\end{figure}

\tableref{tab:rollout_comparison} compares the performance of different configurations of the rollout method; cf.~(\ref{eq:rollout_approximation})--(\ref{eq:rollout}). We find that rollout leads to a significant performance improvement when using a lookahead horizon of $\ell = 1$ and a rollout horizon of $m = 5$. This configuration amounts to around $12$ seconds of computation with our commodity hardware (\textsc{m4 pro}). Further increasing these horizons yields marginal improvements in performance while substantially increasing computation time.
\begin{table}
  \centering
  \scalebox{1}{
    \begin{tabular}{llll} \toprule
\rowcolor{lightgray}
      {\textit{Lookahead $\ell$}} & {\textit{Rollout $m$}} &  {\textit{Loss $\mathscr{L}$ ($\downarrow$ better)}} &  {\textit{Compute time (s)}}\\ \midrule
      - & - & $2345$ & $0.9$\\      
      1 & 5 & $147$ & $12.5$\\
      1 & 20 & $147$ & $17.2$\\
      2 & 5 & $143$ & $47.6$\\
      2 & 20 & $143$ & $71.4$\\
      3 & 5 & $\bm{142}$ & $94.1$\\
      3 & 20 & $\bm{142}$ & $315.8$\\                        
    \bottomrule\\
  \end{tabular}}
  \caption{Performance comparison of different instantiations of the rollout method [cf.~(\ref{eq:rollout_approximation})-(\ref{eq:rollout})] based on the example \textsc{scm} in \sectionref{sec:example} with $|\mathcal{D}_t|=30$ samples; cf.~(\ref{eq:dataset}). The performance is quantified by the loss function $\mathscr{L}$ [cf.~(\ref{eq:loss_fun})] and the compute time to select each intervention. The first row contains the performance of the base policy $\pi$, which is defined as the policy that always selects the passive intervention $\mathrm{do}(\emptyset)$, i.e., the policy of monitoring the system without interventions. We approximate the minimization (\ref{eq:rollout}) using Differential evolution (\textsc{de}) \cite[Fig. 3]{differential_evolution}); see \appendixref{appendix:hyperparameters} for details.}\label{tab:rollout_comparison}
\end{table}
\section{Identifying a Causal Model of an IT System}\label{sec:it_system_case}
In this section, we demonstrate how our method can be applied to identify a causal model of an \textsc{it} system. We begin by describing the system configuration and our testbed implementation. Next, we outline the experimental setup and describe how we applied our method to identify the system. Lastly, we present and discuss the experimental findings.
\subsection{\textsc{IT} System}\label{sec:it_system_description}
\begin{figure}[H]
  \centering
  \scalebox{0.86}{
    \input{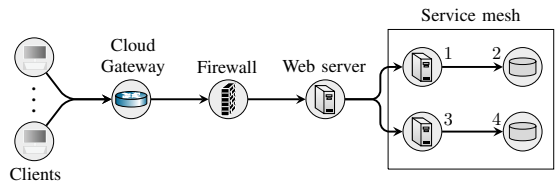}  
  }
  \caption{Architecture of the \textsc{it} system for the experimental evaluation: a cloud-based web application with a service mesh backend.}
  \label{fig:service_chain}
\end{figure}

We consider an \textsc{it} system that involves a cloud-based web application with a backend composed of a web server and a service mesh. Services provided by this mesh are accessed by clients through a cloud gateway; see \figref{fig:service_chain}. The web server is implemented using \textsc{flask} \cite{flask} and the service mesh is implemented using \textsc{kubernetes} \cite{K8} and \textsc{istio} \cite{Istio}.
\vspace{2mm}

\noindent\textbf{\textit{System configuration}.}
The service mesh consists of $4$ physical nodes (labeled $1$-$4$ in \figref{fig:service_chain}), each of which runs two microservices. Specifically, nodes $1$ and $3$ run microservices $(M_1, M_2)$ and nodes $2$ and $4$ run the microservice $M_3$; see \tableref{tab:servers}. Each node is implemented as a \textsc{kubernetes} pod, in which microservices execute within virtual containers.

\begin{table}[H]
  \centering
\scalebox{0.9}{
  \begin{tabular}{ll}
\rowcolor{lightgray}
    {\textit{Microservice}} & {Description}  \\ \midrule
    Web application  & A \textsc{flask} web application \cite{flask}. \\    
    $M_1$  &  A web service that implements an electronic bookstore. \\
    $M_2$  & A \textsc{cpu}-intensive compute service for data processing. \\
    $M_3$  & \textsc{mongodb} database \cite{mongodb}. \\
\end{tabular}
}
\caption{Configurations of microservices in the service mesh.}\label{tab:servers}
\end{table}

\begin{figure}[H]
  \centering
  \scalebox{0.95}{
    \begin{tikzpicture}[fill=white, >=stealth,
    node distance=3cm,
    database/.style={
      cylinder,
      cylinder uses custom fill,
      shape border rotate=90,
      aspect=0.25,
      draw}]

    \tikzset{
node distance = 9em and 4em,
sloped,
   box/.style = {%
    shape=rectangle,
    rounded corners,
    draw=blue!40,
    fill=blue!15,
    align=center,
    font=\fontsize{12}{12}\selectfont},
 arrow/.style = {%
    line width=0.1mm,
    -{Triangle[length=5mm,width=2mm]},
    shorten >=1mm, shorten <=1mm,
    font=\fontsize{8}{8}\selectfont},
}

\node [scale=0.75] (node1) at (0,1.5) {
\begin{tikzpicture}[fill=white, >=stealth,mybackground18={\small \textsc{Node 1}},
    node distance=3cm,    database/.style={
      cylinder,
      cylinder uses custom fill,
      shape border rotate=90,
      aspect=0.25,
      draw}]
    \tikzset{
node distance = 9em and 4em,
sloped,
   box/.style = {%
    shape=rectangle,
    rounded corners,
    draw=blue!40,
    fill=blue!15,
    align=center,
    font=\fontsize{12}{12}\selectfont},
 arrow/.style = {%
    line width=0.1mm,
    shorten >=1mm, shorten <=1mm,
    font=\fontsize{8}{8}\selectfont},
}

\node[scale=0.2] (square) at (0.75,0.9)
{
  \begin{tikzpicture}
    \draw[-, color=black, fill=Red!40] (0,0) to (3,0) to (3, 1.5) to (0,1.5) to (0,0);
    \end{tikzpicture}
  };

\node[scale=0.2] (square) at (2.25,0.9)
{
  \begin{tikzpicture}
    \draw[-, color=black, fill=Cyan!40] (0,0) to (3,0) to (3, 1.5) to (0,1.5) to (0,0);
    \end{tikzpicture}
  };  
\node[inner sep=0pt,align=center, scale=1] (dots4) at (0.75,0.4) {$M_1$};
\node[inner sep=0pt,align=center, scale=1] (dots4) at (2.25,0.4) {$M_2$};

\draw[draw=none,opacity=0] (0,0) rectangle node (m1){} (3,1.5);
\end{tikzpicture}
};

\node [scale=0.75] (node1) at (3,1.5) {
\begin{tikzpicture}[fill=white, >=stealth,mybackground18={\small \textsc{Node 2}},
    node distance=3cm,    database/.style={
      cylinder,
      cylinder uses custom fill,
      shape border rotate=90,
      aspect=0.25,
      draw}]
    \tikzset{
node distance = 9em and 4em,
sloped,
   box/.style = {%
    shape=rectangle,
    rounded corners,
    draw=blue!40,
    fill=blue!15,
    align=center,
    font=\fontsize{12}{12}\selectfont},
 arrow/.style = {%
    line width=0.1mm,
    shorten >=1mm, shorten <=1mm,
    font=\fontsize{8}{8}\selectfont},
}

\node[scale=0.2] (square) at (1.5,0.9)
{
  \begin{tikzpicture}
    \draw[-, color=black, fill=OliveGreen!40] (0,0) to (3,0) to (3, 1.5) to (0,1.5) to (0,0);
    \end{tikzpicture}
  };

\node[inner sep=0pt,align=center, scale=1] (dots4) at (1.5,0.4) {$M_3$};

\draw[draw=none,opacity=0] (0,0) rectangle node (m1){} (3,1.5);
\end{tikzpicture}
};

\node [scale=0.75] (node1) at (-3,0) {
\begin{tikzpicture}[fill=white, >=stealth,mybackground18={\small \textsc{Front node}},
    node distance=3cm,    database/.style={
      cylinder,
      cylinder uses custom fill,
      shape border rotate=90,
      aspect=0.25,
      draw}]
    \tikzset{
node distance = 9em and 4em,
sloped,
   box/.style = {%
    shape=rectangle,
    rounded corners,
    draw=blue!40,
    fill=blue!15,
    align=center,
    font=\fontsize{12}{12}\selectfont},
 arrow/.style = {%
    line width=0.1mm,
    shorten >=1mm, shorten <=1mm,
    font=\fontsize{8}{8}\selectfont},
}

\node[scale=0.2] (square) at (1.5,0.9)
{
  \begin{tikzpicture}
    \draw[-, color=black, fill=Black] (0,0) to (3,0) to (3, 1.5) to (0,1.5) to (0,0);
    \end{tikzpicture}
  };  
\node[inner sep=0pt,align=center, scale=1] (dots4) at (1.5,0.4) {\textit{Web application}};

\draw[draw=none,opacity=0] (0,0) rectangle node (m1){} (3,1.5);
\end{tikzpicture}
};

\node [scale=0.75] (node1) at (0,-1.5) {
\begin{tikzpicture}[fill=white, >=stealth,mybackground18={\small \textsc{Node 3}},
    node distance=3cm,    database/.style={
      cylinder,
      cylinder uses custom fill,
      shape border rotate=90,
      aspect=0.25,
      draw}]
    \tikzset{
node distance = 9em and 4em,
sloped,
   box/.style = {%
    shape=rectangle,
    rounded corners,
    draw=blue!40,
    fill=blue!15,
    align=center,
    font=\fontsize{12}{12}\selectfont},
 arrow/.style = {%
    line width=0.1mm,
    shorten >=1mm, shorten <=1mm,
    font=\fontsize{8}{8}\selectfont},
}

\node[scale=0.2] (square) at (0.75,0.9)
{
  \begin{tikzpicture}
    \draw[-, color=black, fill=Red!40] (0,0) to (3,0) to (3, 1.5) to (0,1.5) to (0,0);
    \end{tikzpicture}
  };

\node[scale=0.2] (square) at (2.25,0.9)
{
  \begin{tikzpicture}
    \draw[-, color=black, fill=Cyan!40] (0,0) to (3,0) to (3, 1.5) to (0,1.5) to (0,0);
    \end{tikzpicture}
  };  
\node[inner sep=0pt,align=center, scale=1] (dots4) at (0.75,0.4) {$M_1$};
\node[inner sep=0pt,align=center, scale=1] (dots4) at (2.25,0.4) {$M_2$};
\draw[draw=none,opacity=0] (0,0) rectangle node (m1){} (3,1.5);
\end{tikzpicture}
};

\node [scale=0.75] (node1) at (3,-1.5) {
\begin{tikzpicture}[fill=white, >=stealth,mybackground18={\small \textsc{Node 4}},
    node distance=3cm,    database/.style={
      cylinder,
      cylinder uses custom fill,
      shape border rotate=90,
      aspect=0.25,
      draw}]
    \tikzset{
node distance = 9em and 4em,
sloped,
   box/.style = {%
    shape=rectangle,
    rounded corners,
    draw=blue!40,
    fill=blue!15,
    align=center,
    font=\fontsize{12}{12}\selectfont},
 arrow/.style = {%
    line width=0.1mm,
    shorten >=1mm, shorten <=1mm,
    font=\fontsize{8}{8}\selectfont},
}

\node[scale=0.2] (square) at (1.5,0.9)
{
  \begin{tikzpicture}
    \draw[-, color=black, fill=OliveGreen!40] (0,0) to (3,0) to (3, 1.5) to (0,1.5) to (0,0);
    \end{tikzpicture}
  };
\node[inner sep=0pt,align=center, scale=1] (dots4) at (1.5,0.4) {$M_3$};

\draw[draw=none,opacity=0] (0,0) rectangle node (m1){} (3,1.5);
\end{tikzpicture}
};

\draw[-, line width=0.4mm, color=Red, rounded corners] (-1.9, 0.25) to (-1.6, 0.25) to (-1.6, 1.6) to (-1.1, 1.6);
\draw[-, line width=0.4mm, color=Blue, rounded corners, dashed] (-1.9, 0.15) to (-1.4, 0.15) to (-1.4, 1.3) to (-1.1, 1.3);
\draw[-, line width=0.4mm, color=Red, rounded corners] (-1.9, -0.4) to (-1.6, -0.4) to (-1.6, -1.9) to (-1.1, -1.9);
\draw[-, line width=0.4mm, color=Blue, rounded corners, dashed] (-1.9, -0.3) to (-1.4, -0.3) to (-1.4, -1.6) to (-1.1, -1.6);

\draw[-, line width=0.4mm, color=Red, rounded corners] (1.1, 1.6) to (1.9, 1.6);

\draw[-, line width=0.4mm, color=Red, rounded corners] (1.1, -1.9) to (1.9, -1.9);

\draw[-, line width=0.4mm, color=Red, rounded corners] (-4.9, 0.1) to (-4.12, 0.1);
\draw[-, line width=0.4mm, color=Blue, rounded corners, dashed] (-4.9, -0.2) to (-4.12, -0.2);

\node[inner sep=0pt,align=center, scale=0.75] (dots4) at (-4.53,-0.65) {\textit{Service}\\\textit{requests}};

\draw[-, line width=0.4mm, color=Red, rounded corners] (0.2, 0.1) to (1, 0.1);
\draw[-, line width=0.4mm, color=Blue, rounded corners, dashed] (0.2, -0.2) to (1, -0.2);

\node[inner sep=0pt,align=center, scale=0.75] (dots4) at (1.7,0.1) {\textit{Service} $S_1$};
\node[inner sep=0pt,align=center, scale=0.75] (dots4) at (1.7,-0.2) {\textit{Service} $S_2$};

\node[inner sep=0pt,align=center, scale=0.75] (dots4) at (0,0.65) {\textsc{cpu}: $C_1$};
\node[inner sep=0pt,align=center, scale=0.75] (dots4) at (0,-0.65) {\textsc{cpu}: $C_3$};

\node[inner sep=0pt,align=center, scale=0.6, rotate=90] (dots4) at (-1.75,0.75) {$P_{1}, B_1$};
\node[inner sep=0pt,align=center, scale=0.6, rotate=90] (dots4) at (-1.25,0.75) {$P_{2}, B_2$};

\node[inner sep=0pt,align=center, scale=0.6, rotate=90] (dots4) at (-1.75,-0.95) {$1-P_{1}, B_1$};
\node[inner sep=0pt,align=center, scale=0.6, rotate=90] (dots4) at (-1.25,-0.95) {$1-P_{2}, B_2$};

\end{tikzpicture}                   
  }
  \caption{Service mesh architecture of the \textsc{it} system. Nodes are \textsc{kubernetes} pods running microservices $M_i$ in containers, which collectively provide services $S_1$ and $S_2$. Service $S_1$ invokes microservices $M_1 \rightarrow M_3$, and service $S_2$ invokes microservice $M_2$. A request for service $S_i$ is blocked with probability $B_i$. A service request can be routed via different subgraphs. The specific subgraph is selected probabilistically via $P_{i}$. All nodes have fixed resources except nodes $1$ and $3$, whose \textsc{cpu} counts ($C_1$, $C_3$) are scalable.}
  \label{fig:microservices_structure}
\end{figure}
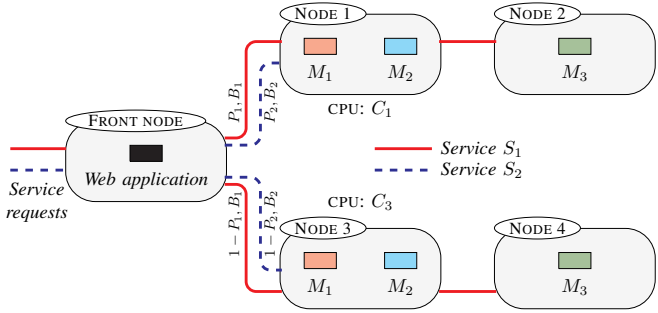

\noindent\textbf{\textit{Network services}.} We deploy two services on the service mesh: an information service and a compute service, which we denote by $S_1$ and $S_2$, respectively. Service $S_1$ invokes the microservices $(M_1, M_3)$ and service $S_2$ invokes microservice $M_2$. These services are accessed by clients that generate service requests, each of which can be handled in two ways, corresponding to different traversals of the service graph in \figref{fig:microservices_structure}. Specifically, a request for service $S_1$ can either traverse the subgraph $\text{\textsc{front node}} \rightarrow 1 \rightarrow 2$ or traverse the subgraph $\text{\textsc{front node}} \rightarrow 3 \rightarrow 4$. Similarly, a request for service $S_2$ can be processed by either node $1$ or $3$.

\vspace{2mm}

\noindent\textbf{\textit{Request routing and blocking}.}
The specific traversal path for a service request is decided by routing probabilities $P_1$ and $P_2$, where $P_{i}$ is the probability that a request for service $S_i$ is routed to node $1$. Apart from these probabilities, each request for service $S_i$ is blocked at the front node with probability $B_i$.

\vspace{2mm}

\noindent\textbf{\textit{Resource allocation}.}
We denote by $C_j$ the \textsc{cpu} allocation to node $j$ in the service mesh; see \figref{fig:microservices_structure}. In our setup, the resource allocations are fixed for all nodes except for nodes $1$ and $3$, whose \textsc{cpu} counts ($C_1$ and $C_3$) are scalable.

\vspace{2mm}

\noindent\textbf{\textit{Implementation}.}
We run the \textsc{it} system on our testbed at \textsc{kth}. This testbed includes a cluster of \textsc{poweredge r715 2u} servers connected through a gigabit Ethernet switch. Each server has $64$ \textsc{gb} \textsc{ram}, two $12$-core \textsc{amd opteron} processors, and four $1$ \textsc{gb} network interfaces. All servers run \textsc{ubuntu server 18.04.6} ($64$ bit) and their clocks are synchronized through the network time protocol \cite{ntp}. The source code of our implementation and a dataset of traces from our testbed are available in \cite{paper_source_code}.

\subsection{Causal Model of the \textsc{IT} System}
\begin{figure}[H]
  \centering
  \scalebox{1.45}{
    \begin{tikzpicture}[fill=white, >=stealth,
    node distance=3cm,
    database/.style={
      cylinder,
      cylinder uses custom fill,
      shape border rotate=90,
      aspect=0.25,
      draw}]

    \tikzset{
node distance = 9em and 4em,
sloped,
   box/.style = {%
    shape=rectangle,
    rounded corners,
    draw=blue!40,
    fill=blue!15,
    align=center,
    font=\fontsize{12}{12}\selectfont},
 arrow/.style = {%
    line width=0.1mm,
    -{Triangle[length=5mm,width=2mm]},
    shorten >=1mm, shorten <=1mm,
    font=\fontsize{8}{8}\selectfont},
}


\node[scale=0.7] (gi) at (0,0)
{
\begin{tikzpicture}
\node[draw,circle, minimum width=0.8cm, scale=0.7,fill=black!12](L2) at (1.5,0) {};
\node[draw,circle, minimum width=0.8cm, scale=0.7,fill=black!12](B2) at (1.5,-1) {};
\node[draw,circle, minimum width=0.8cm, scale=0.7,fill=black!12](L1) at (1.5,1) {};
\node[draw,circle, minimum width=0.8cm, scale=0.7,fill=black!12](B1) at (1.5,2) {};
\node[draw,circle, minimum width=0.8cm, scale=0.7](L1c) at (3,1.5) {};
\node[draw,circle, minimum width=0.8cm, scale=0.7](L2c) at (3,-0.5) {};
\node[draw,circle, minimum width=0.8cm, scale=0.7](R1) at (4.5,1) {};
\node[draw,circle, minimum width=0.8cm, scale=0.7](R2) at (4.5,0) {};
\node[draw,circle, minimum width=0.8cm, scale=0.7,fill=black!12](C2) at (6,0) {};
\node[draw,circle, minimum width=0.8cm, scale=0.7,fill=black!12](C1) at (6,1) {};
\node[draw,circle, minimum width=0.8cm, scale=0.7,fill=black!12](P1) at (4.5,2) {};
\node[draw,circle, minimum width=0.8cm, scale=0.7,fill=black!12](P2) at (4.5,-1) {};

\node[draw,circle, minimum width=0.8cm, scale=0.7,fill=black!12](EpsR1) at (6,2) {};
\node[draw,circle, minimum width=0.8cm, scale=0.7,fill=black!12](EpsR2) at (6,-1) {};

\node[inner sep=0pt,align=center, scale=0.8] (dots4) at (6,2) {$\epsilon_{R_1}$};
\node[inner sep=0pt,align=center, scale=0.8] (dots4) at (6,-1) {$\epsilon_{R_2}$};
\node[inner sep=0pt,align=center, scale=0.8] (dots4) at (1.5,2) {$B_1$};
\node[inner sep=0pt,align=center, scale=0.8] (dots4) at (1.5,1) {$L_1$};
\node[inner sep=0pt,align=center, scale=0.8] (dots4) at (1.5,0) {$L_2$};
\node[inner sep=0pt,align=center, scale=0.8] (dots4) at (1.5,-1) {$B_2$};
\node[inner sep=0pt,align=center, scale=0.8] (dots4) at (3,1.5) {$\Tilde{L}_1$};
\node[inner sep=0pt,align=center, scale=0.8] (dots4) at (3,-0.5) {$\Tilde{L}_2$};
\node[inner sep=0pt,align=center, scale=0.8] (dots4) at (4.5,1) {$R_1$};
\node[inner sep=0pt,align=center, scale=0.8] (dots4) at (4.5,0) {$R_2$};
\node[inner sep=0pt,align=center, scale=0.8] (dots4) at (6,1) {$C_1$};
\node[inner sep=0pt,align=center, scale=0.8] (dots4) at (6,0) {$C_3$};
\node[inner sep=0pt,align=center, scale=0.8] (dots4) at (4.5,2) {$P_1$};
\node[inner sep=0pt,align=center, scale=0.8] (dots4) at (4.5,-1) {$P_2$};
\draw[-{Latex[length=2mm]}, line width=0.22mm, color=black, dashed] (EpsR1) to (R1);
\draw[-{Latex[length=2mm]}, line width=0.22mm, color=black, dashed] (EpsR2) to (R2);
\draw[-{Latex[length=2mm]}, line width=0.22mm, color=black,dashed] (B1) to (L1c);
\draw[-{Latex[length=2mm]}, line width=0.22mm, color=black, dashed] (L1) to (L1c);
\draw[-{Latex[length=2mm]}, line width=0.22mm, color=black,dashed] (B2) to (L2c);
\draw[-{Latex[length=2mm]}, line width=0.22mm, color=black, dashed] (L2) to (L2c);
\draw[-{Latex[length=2mm]}, line width=0.22mm, color=black] (L1c) to (R1);
\draw[-{Latex[length=2mm]}, line width=0.22mm, color=black] (L1c) to (R2);
\draw[-{Latex[length=2mm]}, line width=0.22mm, color=black] (L2c) to (R1);
\draw[-{Latex[length=2mm]}, line width=0.22mm, color=black] (L2c) to (R2);
\draw[-{Latex[length=2mm]}, line width=0.22mm, color=black,dashed] (P1) to (R1);
\draw[-{Latex[length=2mm]}, line width=0.22mm, color=black, bend left=50, dashed] (P1) to (R2);
\draw[-{Latex[length=2mm]}, line width=0.22mm, color=black, bend right=50, dashed] (P2) to (R1);
\draw[-{Latex[length=2mm]}, line width=0.22mm, color=black,dashed] (P2) to (R2);
\draw[-{Latex[length=2mm]}, line width=0.22mm, color=black,dashed] (C1) to (R1);
\draw[-{Latex[length=2mm]}, line width=0.22mm, color=black,dashed] (C1) to (R2);
\draw[-{Latex[length=2mm]}, line width=0.22mm, color=black,dashed] (C2) to (R1);
\draw[-{Latex[length=2mm]}, line width=0.22mm, color=black,dashed] (C2) to (R2);


\end{tikzpicture}
};




\end{tikzpicture}  
  }
  \caption{Causal graph of the \textsc{it} system in \figref{fig:service_chain}.}
  \label{fig:causal_graphs_microservices}
\end{figure}
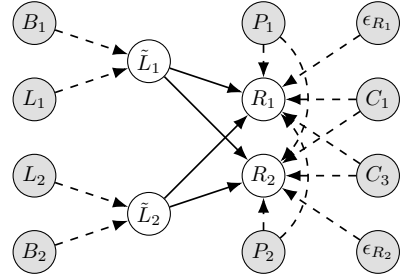

We model the system described in the preceding section as an \textsc{scm} with the causal graph shown in \figref{fig:causal_graphs_microservices}. The system variables of the model are defined as follows.
\begin{itemize}
\item \textit{Exogenous variables $\mathbf{U}$}:
\begin{itemize}
\item $B_i$: blocking probability of service $S_i$;
\item $P_i$: routing probability of service $S_i$ to node $1$;
\item $C_j$: \textsc{cpu} allocation to node $j$;
\item $L_i$: load of service $S_i$ (requests per second); and
\item $\epsilon_{R_1},\epsilon_{R_2}$: random noise variables.
\end{itemize}  
\item \textit{Endogenous variables $\mathbf{V}$}:
\begin{itemize}
\item $\tilde{L}_i$: carried (i.e., non-blocked) load of service $S_i$; and
\item $R_i$: response time (s) of service $S_i$.
\end{itemize}  
\end{itemize}
The operating region [cf.~\defref{def:or}] is defined as follows.
\begin{align*}
&\mathcal{R}(B_i)=\mathcal{R}(P_i) = [0,1]; \text{ }\mathcal{R}(L_i)=\mathcal{R}(\Tilde{L}_i)=[0,50]; \\
&\mathcal{R}(C_j)=\{1,\hdots,5\};\text{ }\mathcal{R}(\epsilon_{R_i}) = [-\infty,\infty];\text{ }\mathcal{R}(R_i)=[0,10].
\end{align*}  

All the exogenous variables except the noise variables are controllable, i.e.,
$$
\mathbf{X}=\{B_1, B_2, L_1, L_2, P_1, P_2, C_1, C_3\}.
$$
These variables can be externally controlled as follows: service loads ($L_i$) can be emulated; \textsc{cpu} allocations ($C_j$) can be configured in \textsc{kubernetes}; and the routing ($P_i$) and blocking probabilities ($B_i$) can be adjusted via \textsc{istio}.

Following the graph in \figref{fig:causal_graphs_microservices}, the causal functions are
\begin{subequations}\label{eq:causal_functions_it_system}
\begin{align}
\Tilde{L}_1 &= f_{\Tilde{L}_1,t}(B_1,L_1),\\
\Tilde{L}_2 &= f_{\Tilde{L}_2,t}(B_2,L_2),\\
R_1 &= f_{R_1,t}(\Tilde{L}_1, \Tilde{L}_2, P_1, P_2, C_1, C_3, \epsilon_{R_1}),\\
R_2 &= f_{R_2,t}(\Tilde{L}_1, \Tilde{L}_2, P_1, P_2, C_1, C_3, \epsilon_{R_2}).
\end{align}
\end{subequations}

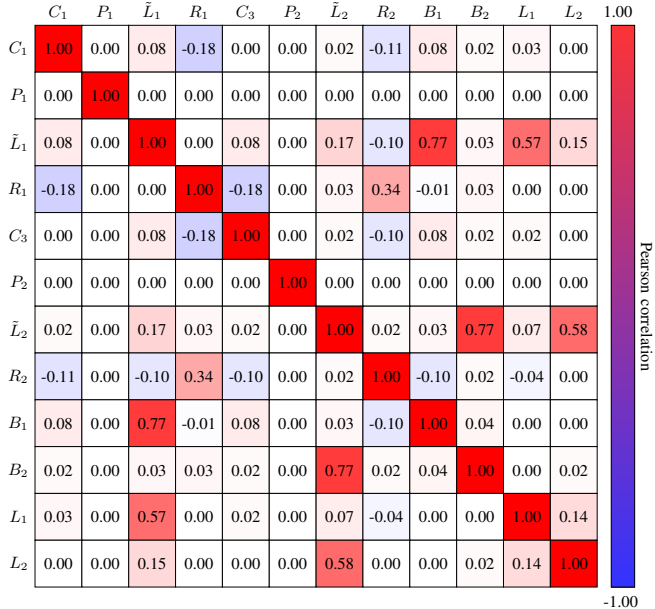
\begin{figure}
  \centering
  \scalebox{0.865}{
    \begin{tikzpicture}

\node[scale=0.8] (kth_cr) at (0,2.15)
{
\begin{tikzpicture}[x=0.9cm,y=0.9cm,font=\small]

\foreach \col [count=\i from 0] in {$C_1$,$P_1$,$\Tilde{L}_1$,$R_1$,$C_3$,$P_2$,$\Tilde{L}_2$,$R_2$,$B_1$,$B_2$,$L_1$,$L_2$}
  \node[anchor=south] at (\i+0.5,0) {\col};

\foreach \row [count=\j from 0] in {$C_1$,$P_1$,$\Tilde{L}_1$,$R_1$,$C_3$,$P_2$,$\Tilde{L}_2$,$R_2$,$B_1$,$B_2$,$L_1$,$L_2$}
  \node[anchor=east] at (0,-\j-0.5) {\row};

\newcommand{\sq}[4]{%
  \path[fill=#1,draw=black] (#2,#3) rectangle ++(1,-1);
  \node at (#2+0.5,#3-0.5) {#4};
}


\sq{red!100}{0}{0}{1.00}\sq{white}{1}{0}{0.00}\sq{red!8}{2}{0}{0.08}\sq{blue!18}{3}{0}{-0.18}
\sq{white}{4}{0}{0.00}\sq{white}{5}{0}{0.00}\sq{red!2}{6}{0}{0.02}\sq{blue!11}{7}{0}{-0.11}
\sq{red!8}{8}{0}{0.08}\sq{red!2}{9}{0}{0.02}\sq{red!3}{10}{0}{0.03}\sq{white}{11}{0}{0.00}

\sq{white}{0}{-1}{0.00}\sq{red!100}{1}{-1}{1.00}\sq{white}{2}{-1}{0.00}\sq{white}{3}{-1}{0.00}
\sq{white}{4}{-1}{0.00}\sq{white}{5}{-1}{0.00}\sq{white}{6}{-1}{0.00}\sq{white}{7}{-1}{0.00}
\sq{white}{8}{-1}{0.00}\sq{white}{9}{-1}{0.00}\sq{white}{10}{-1}{0.00}\sq{white}{11}{-1}{0.00}

\sq{red!8}{0}{-2}{0.08}\sq{white}{1}{-2}{0.00}\sq{red!100}{2}{-2}{1.00}\sq{white}{3}{-2}{0.00}
\sq{red!8}{4}{-2}{0.08}\sq{white}{5}{-2}{0.00}\sq{red!17}{6}{-2}{0.17}\sq{blue!10}{7}{-2}{-0.10}
\sq{red!77}{8}{-2}{0.77}\sq{red!3}{9}{-2}{0.03}\sq{red!57}{10}{-2}{0.57}\sq{red!15}{11}{-2}{0.15}

\sq{blue!18}{0}{-3}{-0.18}\sq{white}{1}{-3}{0.00}\sq{white}{2}{-3}{0.00}\sq{red!100}{3}{-3}{1.00}
\sq{blue!18}{4}{-3}{-0.18}\sq{white}{5}{-3}{0.00}\sq{red!3}{6}{-3}{0.03}\sq{red!34}{7}{-3}{0.34}
\sq{blue!1}{8}{-3}{-0.01}\sq{red!3}{9}{-3}{0.03}\sq{white}{10}{-3}{0.00}\sq{white}{11}{-3}{0.00}

\sq{white}{0}{-4}{0.00}\sq{white}{1}{-4}{0.00}\sq{red!8}{2}{-4}{0.08}\sq{blue!18}{3}{-4}{-0.18}
\sq{red!100}{4}{-4}{1.00}\sq{white}{5}{-4}{0.00}\sq{red!2}{6}{-4}{0.02}\sq{blue!10}{7}{-4}{-0.10}
\sq{red!8}{8}{-4}{0.08}\sq{red!2}{9}{-4}{0.02}\sq{red!2}{10}{-4}{0.02}\sq{white}{11}{-4}{0.00}

\sq{white}{0}{-5}{0.00}\sq{white}{1}{-5}{0.00}\sq{white}{2}{-5}{0.00}\sq{white}{3}{-5}{0.00}
\sq{white}{4}{-5}{0.00}\sq{red!100}{5}{-5}{1.00}\sq{white}{6}{-5}{0.00}\sq{white}{7}{-5}{0.00}
\sq{white}{8}{-5}{0.00}\sq{white}{9}{-5}{0.00}\sq{white}{10}{-5}{0.00}\sq{white}{11}{-5}{0.00}

\sq{red!2}{0}{-6}{0.02}\sq{white}{1}{-6}{0.00}\sq{red!17}{2}{-6}{0.17}\sq{red!3}{3}{-6}{0.03}
\sq{red!2}{4}{-6}{0.02}\sq{white}{5}{-6}{0.00}\sq{red!100}{6}{-6}{1.00}\sq{red!2}{7}{-6}{0.02}
\sq{red!3}{8}{-6}{0.03}\sq{red!77}{9}{-6}{0.77}\sq{red!7}{10}{-6}{0.07}\sq{red!58}{11}{-6}{0.58}

\sq{blue!11}{0}{-7}{-0.11}\sq{white}{1}{-7}{0.00}\sq{blue!10}{2}{-7}{-0.10}\sq{red!34}{3}{-7}{0.34}
\sq{blue!10}{4}{-7}{-0.10}\sq{white}{5}{-7}{0.00}\sq{red!2}{6}{-7}{0.02}\sq{red!100}{7}{-7}{1.00}
\sq{blue!10}{8}{-7}{-0.10}\sq{red!2}{9}{-7}{0.02}\sq{blue!4}{10}{-7}{-0.04}\sq{white}{11}{-7}{0.00}

\sq{red!8}{0}{-8}{0.08}\sq{white}{1}{-8}{0.00}\sq{red!77}{2}{-8}{0.77}\sq{blue!1}{3}{-8}{-0.01}
\sq{red!8}{4}{-8}{0.08}\sq{white}{5}{-8}{0.00}\sq{red!3}{6}{-8}{0.03}\sq{blue!10}{7}{-8}{-0.10}
\sq{red!100}{8}{-8}{1.00}\sq{red!4}{9}{-8}{0.04}\sq{white}{10}{-8}{0.00}\sq{white}{11}{-8}{0.00}

\sq{red!2}{0}{-9}{0.02}\sq{white}{1}{-9}{0.00}\sq{red!3}{2}{-9}{0.03}\sq{red!3}{3}{-9}{0.03}
\sq{red!2}{4}{-9}{0.02}\sq{white}{5}{-9}{0.00}\sq{red!77}{6}{-9}{0.77}\sq{red!2}{7}{-9}{0.02}
\sq{red!4}{8}{-9}{0.04}\sq{red!100}{9}{-9}{1.00}\sq{white}{10}{-9}{0.00}\sq{red!2}{11}{-9}{0.02}

\sq{red!3}{0}{-10}{0.03}\sq{white}{1}{-10}{0.00}\sq{red!57}{2}{-10}{0.57}\sq{white}{3}{-10}{0.00}
\sq{red!2}{4}{-10}{0.02}\sq{white}{5}{-10}{0.00}\sq{red!7}{6}{-10}{0.07}\sq{blue!4}{7}{-10}{-0.04}
\sq{white}{8}{-10}{0.00}\sq{white}{9}{-10}{0.00}\sq{red!100}{10}{-10}{1.00}\sq{red!14}{11}{-10}{0.14}

\sq{white}{0}{-11}{0.00}\sq{white}{1}{-11}{0.00}\sq{red!15}{2}{-11}{0.15}\sq{white}{3}{-11}{0.00}
\sq{white}{4}{-11}{0.00}\sq{white}{5}{-11}{0.00}\sq{red!58}{6}{-11}{0.58}\sq{white}{7}{-11}{0.00}
\sq{white}{8}{-11}{0.00}\sq{red!2}{9}{-11}{0.02}\sq{red!14}{10}{-11}{0.14}\sq{red!100}{11}{-11}{1.00}

\shade[top color=red!80,middle color=white,bottom color=blue!80]
      (12.3,0) rectangle ++(0.5,-12);
\draw[black] (12.3,0) rectangle ++(0.5,-12);

\node[anchor=west] at (12,  0.3)  {1.00};
\node[anchor=west] at (12,-12.3) {-1.00};

\node[rotate=-90,anchor=south] at (12.8,-6) {Pearson correlation};

\end{tikzpicture}
};

\end{tikzpicture}   
  }
\caption{Correlation matrix for the variables in the causal model [cf.~\figref{fig:causal_graphs_microservices}] of the \textsc{it} system in \figref{fig:service_chain}. The numbers in each cell indicate the Pearson correlation coefficient between two variables.}
\label{fig:correlation_matrix}
\end{figure}

\begin{figure*}
  \centering
  \scalebox{0.8}{
    \includegraphics{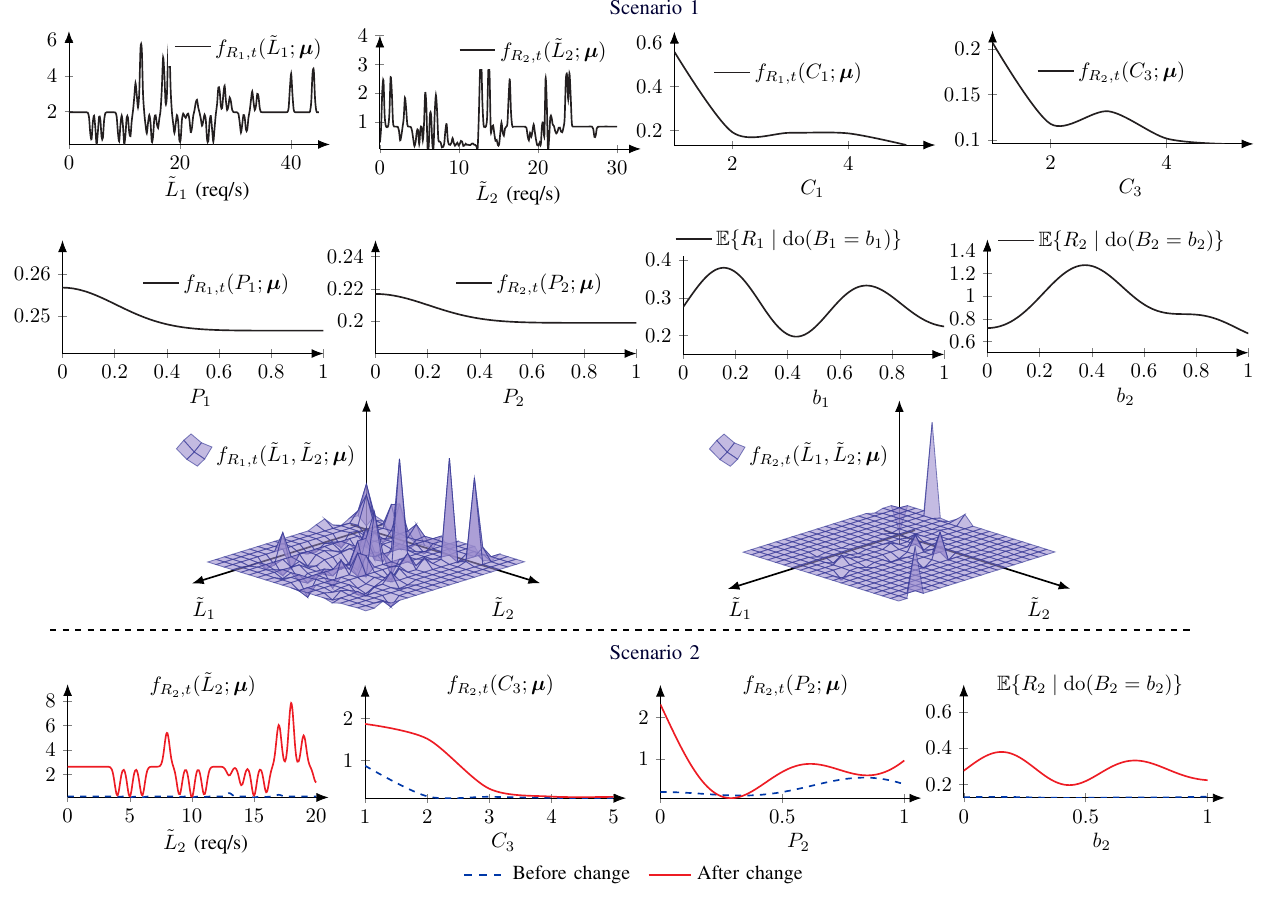}
  }
\caption{Some of the causal functions in the structural causal model (\textsc{scm}) of the \textsc{it} system in \figref{fig:service_chain}. The function $f_{R_i,t}(\Tilde{L}_i; \bm{\mu})$ denotes $f_{R_i,t}(\Tilde{L}_1, \Tilde{L}_2, P_1, P_2, C_1, C_3, \epsilon_{R_i})$ [cf.~(\ref{eq:causal_functions_it_system})] evaluated with all inputs fixed to their mean values except for $\Tilde{L}_i$. The upper plots show the ground truth causal functions for \hyperref[scenario_1]{Scenario 1} and the lower plots show the ground truth causal functions for \hyperref[scenario_2]{Scenario 2}.}
\label{fig:system_model}
\end{figure*}

\subsection{Evaluation Scenarios}
To evaluate our method for identifying the causal functions in (\ref{eq:causal_functions_it_system}) from measurement data, we consider two scenarios: one focuses on identifying the functions during steady-state operation, and the other on tracking the causal functions as they change over time. In both scenarios, the system operates under a nominal configuration: the routing probabilities are set to $P_1 = P_2 = 0.5$, the \textsc{cpu} allocations to $C_1 = C_3 = 1$, and the blocking probabilities to $B_1 = B_2 = 0$. Interventions are required to change this configuration.

\vspace{2mm}

\noindent\textbf{\textit{\setword{Scenario 1}{scenario_1} (Stationary system)}.} In this scenario, we run both the information service $S_1$ and the compute service $S_2$ on the service mesh. The service loads are kept constant with $L_1=4$ requests per second and $L_2=15$ requests per second. This setup defines the current operating region of the system; cf.~\defref{def:cor}. The causal dependencies among the system variables follow the graph shown in \figref{fig:causal_graphs_microservices}.

\vspace{2mm}

\noindent\textbf{\textit{\setword{Scenario 2}{scenario_2} (Non-stationary system)}.} In this scenario, we investigate how the estimates of the causal functions provided by our method adapt to a change in the service offering. The scenario is divided into two time intervals. In the first interval, which starts at $t=1$, the service mesh runs only the compute service $S_2$, which we load with $L_2=1$ requests per second. In the second interval, beginning at $t=11$, we start the information service $S_1$ in the background and load it with $L_1 = 20$ requests per second. This change introduces additional background load on shared resources, such as \textsc{cpu} and bandwidth, which in turn affects the response time of service $S_2$. As a result, the causal function for the response time, i.e., $f_{R_2,t}$, has to be re-estimated after the change.

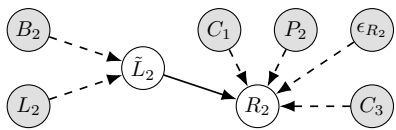
\begin{figure}[H]
  \centering
  \scalebox{1.45}{
    \begin{tikzpicture}[fill=white, >=stealth,
    node distance=3cm,
    database/.style={
      cylinder,
      cylinder uses custom fill,
      shape border rotate=90,
      aspect=0.25,
      draw}]

    \tikzset{
node distance = 9em and 4em,
sloped,
   box/.style = {%
    shape=rectangle,
    rounded corners,
    draw=blue!40,
    fill=blue!15,
    align=center,
    font=\fontsize{12}{12}\selectfont},
 arrow/.style = {%
    line width=0.1mm,
    -{Triangle[length=5mm,width=2mm]},
    shorten >=1mm, shorten <=1mm,
    font=\fontsize{8}{8}\selectfont},
}

\node[scale=0.7] (gi) at (0,0)
{
\begin{tikzpicture}
\node[draw,circle, minimum width=0.8cm, scale=0.7,fill=black!12](L1) at (1.5,1) {};
\node[draw,circle, minimum width=0.8cm, scale=0.7,fill=black!12](B1) at (1.5,2) {};
\node[draw,circle, minimum width=0.8cm, scale=0.7](L1c) at (3,1.5) {};
\node[draw,circle, minimum width=0.8cm, scale=0.7](R1) at (4.5,1) {};
\node[draw,circle, minimum width=0.8cm, scale=0.7,fill=black!12](C2) at (4,2) {};
\node[draw,circle, minimum width=0.8cm, scale=0.7,fill=black!12](C1) at (6,1) {};
\node[draw,circle, minimum width=0.8cm, scale=0.7,fill=black!12](P1) at (5,2) {};

\node[draw,circle, minimum width=0.8cm, scale=0.7,fill=black!12](EpsR1) at (6,2) {};

\node[inner sep=0pt,align=center, scale=0.8] (dots4) at (6,2) {$\epsilon_{R_2}$};

\node[inner sep=0pt,align=center, scale=0.8] (dots4) at (1.5,2) {$B_2$};
\node[inner sep=0pt,align=center, scale=0.8] (dots4) at (1.5,1) {$L_2$};
\node[inner sep=0pt,align=center, scale=0.8] (dots4) at (3,1.5) {$\Tilde{L}_2$};
\node[inner sep=0pt,align=center, scale=0.8] (dots4) at (4.5,1) {$R_2$};
\node[inner sep=0pt,align=center, scale=0.8] (dots4) at (6,1) {$C_3$};
\node[inner sep=0pt,align=center, scale=0.8] (dots4) at (4,2) {$C_1$};
\node[inner sep=0pt,align=center, scale=0.8] (dots4) at (5,2) {$P_2$};

\draw[-{Latex[length=2mm]}, line width=0.22mm, color=black, dashed] (EpsR1) to (R1);;
\draw[-{Latex[length=2mm]}, line width=0.22mm, color=black,dashed] (B1) to (L1c);
\draw[-{Latex[length=2mm]}, line width=0.22mm, color=black, dashed] (L1) to (L1c);
\draw[-{Latex[length=2mm]}, line width=0.22mm, color=black] (L1c) to (R1);
\draw[-{Latex[length=2mm]}, line width=0.22mm, color=black,dashed] (P1) to (R1);
\draw[-{Latex[length=2mm]}, line width=0.22mm, color=black,dashed] (C1) to (R1);
\draw[-{Latex[length=2mm]}, line width=0.22mm, color=black,dashed] (C2) to (R1);

\end{tikzpicture}
};

\end{tikzpicture}    
  }
  \caption{Causal graph for \hyperref[scenario_2]{Scenario 2}.}
  \label{fig:causal_graph_scenario_2}
\end{figure}

To focus the analysis on the identification of the time-varying function $f_{R_2,t}$, we use a simplified \textsc{scm} for this scenario. Specifically, we simplify the \textsc{scm} for \hyperref[scenario_1]{Scenario 1} by only considering variables related to service $S_2$, i.e., we treat the influence of service $S_1$ as part of the environment. This simplification results in the causal graph shown in \figref{fig:causal_graph_scenario_2}.

\begin{figure*}
  \centering
  \scalebox{0.84}{    
    \includegraphics{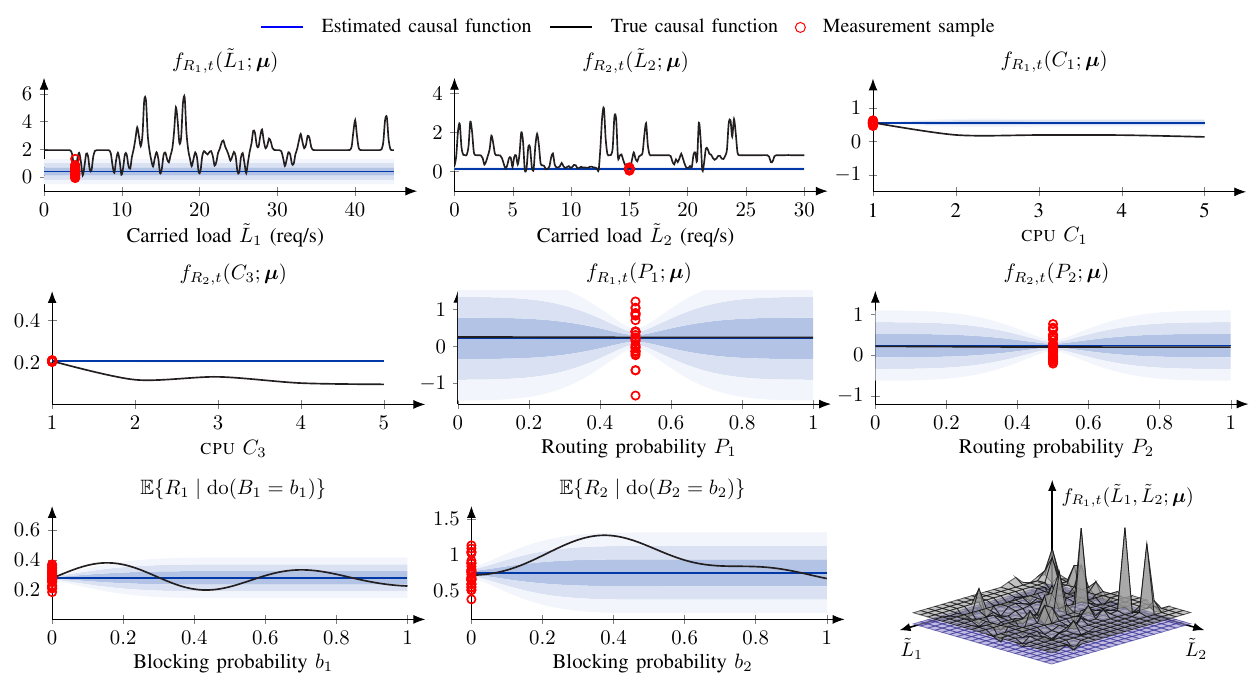}
  }
    \caption{\hyperref[scenario_1]{Scenario 1}: the system is stationary and the causal functions do not change. The figure shows some of the causal functions for the \textsc{it} system described in \sectionref{sec:it_system_description} estimated using the \textsc{gp} estimator $\varphi(\mathcal{D}_t)$ [cf.~(\ref{eq:estimator})] based on $30$ samples (i.e., $|\mathcal{D}_t|=30$) collected through \textbf{passive learning}. Curves show the mean values; shaded regions indicate one, two, and three standard deviations from the mean (darker to lighter shades of blue). The function $f_{R_i,t}(\Tilde{L}_i; \bm{\mu})$ denotes $f_{R_i,t}(\Tilde{L}_1, \Tilde{L}_2, P_1, P_2, C_1, C_3, \epsilon_{R_i})$ [cf.~(\ref{eq:causal_functions_it_system})] evaluated with all inputs fixed to their mean values except for $\Tilde{L}_i$.}
    \label{fig:learned_model_passive}
  \end{figure*}
\subsection{Causal Functions}
To evaluate our method, we need access to the causal functions $\mathbf{F}_t$ [cf.~(\ref{eq:causal_functions_it_system})] to compute the loss function $\mathscr{L}$; cf.~(\ref{eq:loss_fun}). To obtain these functions, we explore the (complete) operating region [cf.~\defref{def:or}] and collect $100$ samples per system variable for each configuration of control variables we consider. This (offline) process takes several days and yields a total dataset of over $30,000$ measurement samples per system variable. We then use this data and the estimator $\varphi$ [cf.~(\ref{eq:estimator})] to learn the causal functions $\mathbf{F}_t$. \Figref{fig:correlation_matrix} shows the correlation matrix of the collected data and \figref{fig:system_model} shows some of the causal functions. The matrix reveals several high correlations, not all of which are causal. For example, the load of service $1$ ($L_1$) is correlated with the load of service $2$ ($L_2$). Moreover, we observe in \figref{fig:system_model} that the causal functions model complex dependencies between the system variables.

\subsection{Instantiation of Our Method}
We apply our method to estimate the causal functions $\mathbf{F}_t$ in \figref{fig:system_model}. To this end, we instantiate our method as described in \sectionref{sec:example}, i.e., we define the base intervention policy in (\ref{eq:rollout}) to be the policy that always selects the passive intervention $\mathrm{do}(\emptyset)$; we define the number of samples per intervention to be $M=1$; we set the lookahead and rollout horizons in (\ref{eq:rollout_approximation})--(\ref{eq:rollout}) as $\ell=1$ and $m=5$, respectively; we define the cost approximation in (\ref{eq:rollout_approximation}) as $\Tilde{J}(\mathbf{b})=\mathcal{L}(\mathbf{b})$; and we set the mean and covariance functions of the \textsc{gp} estimator $\varphi$ [cf.~(\ref{eq:estimator})] according to (\ref{eq:gp_prior_def}). To allow the \textsc{gp} estimator to ``forget'' old data when the system changes in \hyperref[scenario_2]{Scenario 2}, we define the dataset $\mathcal{D}_t$ [cf.~(\ref{eq:dataset})] as a first-in-first-out buffer of size $10$.

\vspace{2mm}

\noindent\textbf{\textit{Intervention costs.}} The cost function $c$ [cf.~(\ref{eq:objective})] depends on the operational impact of different interventions in a specific system. For our experiments, we define this function using the intervention costs listed in \tableref{tab:intervention_costs}. These costs reflect the relative disruption of each intervention type, with high costs assigned to interventions like changing \textsc{cpu} allocations ($C_j$) and lower costs to less disruptive interventions, such as adjusting routing ($R_i$) or blocking ($B_i$) probabilities.
\begin{table}[H]
  \centering
\scalebox{1}{
  \begin{tabular}{ll}
\rowcolor{lightgray}
    {\textit{Intervention}} & {Cost}   \\ \midrule
    Emulating the service load $L_i$ & $3000$\\
    Adjusting the routing probability $P_i$  & $1000$\\
    Adjusting the blocking probability $B_i$  & $2000$\\
    Modifying the \textsc{cpu} allocation $C_j$  & $3000$\\
    Monitoring without intervening, i.e., $\mathrm{do}(\emptyset)$ & $1$\\        
\end{tabular}
}
\caption{Intervention costs for defining the cost function $c$; cf.~(\ref{eq:objective}).}\label{tab:intervention_costs}
\end{table}
\subsection{Evaluation Results}
\begin{figure*}
  \centering
  \scalebox{0.84}{
    \includegraphics{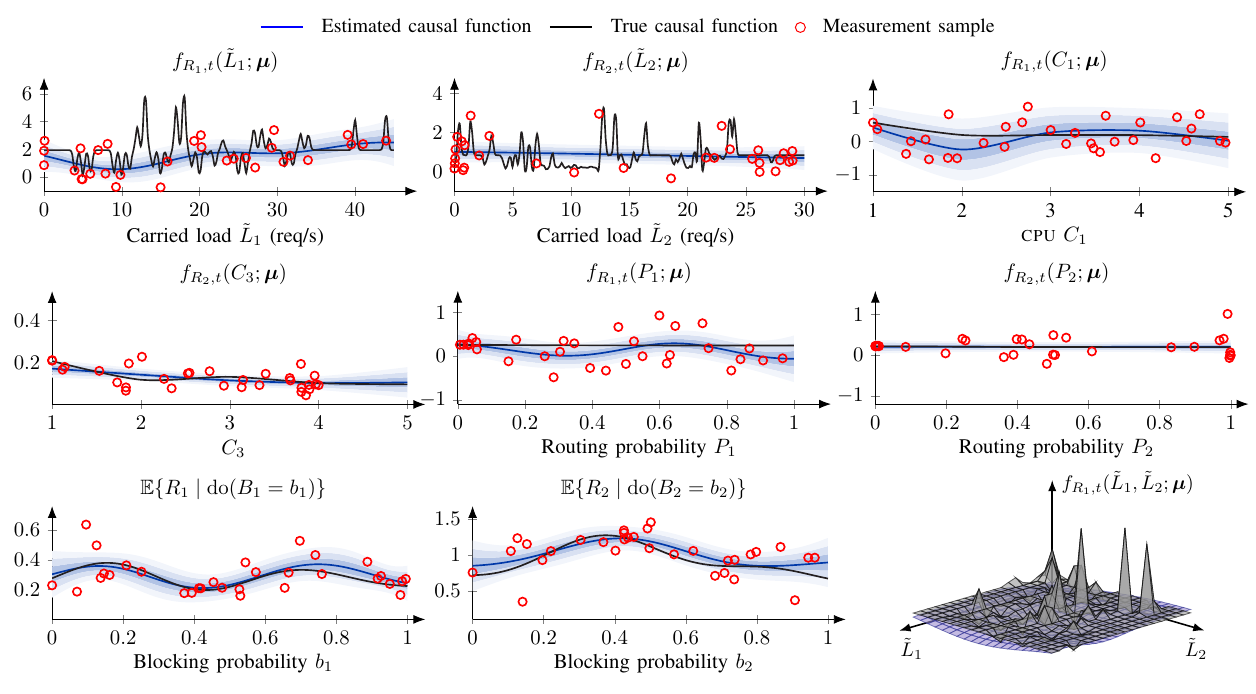}
  }
    \caption{\hyperref[scenario_1]{Scenario 1}: the system is stationary and the causal functions do not change. The figure shows learned causal functions for the \textsc{it} system described in \sectionref{sec:it_system_description}. The functions are learned using the \textsc{gp} estimator $\varphi(\mathcal{D}_t)$ [cf.~(\ref{eq:estimator})] based on $30$ samples (i.e., $|\mathcal{D}_t|=30$) collected using our \textbf{active learning method} with the rollout intervention policy; cf.~(\ref{eq:rollout}). Curves show the mean values; shaded regions indicate one, two, and three standard deviations from the mean (darker to lighter shades of blue). The function $f_{R_i,t}(\Tilde{L}_i; \bm{\mu})$ denotes $f_{R_i,t}(\Tilde{L}_1, \Tilde{L}_2, P_1, P_2, C_1, C_3, \epsilon_{R_i})$ [cf.~(\ref{eq:causal_functions_it_system})] evaluated with all inputs fixed to their mean values except for $\Tilde{L}_i$.}
    \label{fig:learned_model_causal}
  \end{figure*}
\begin{figure*}
  \centering
  \scalebox{0.86}{
    \includegraphics{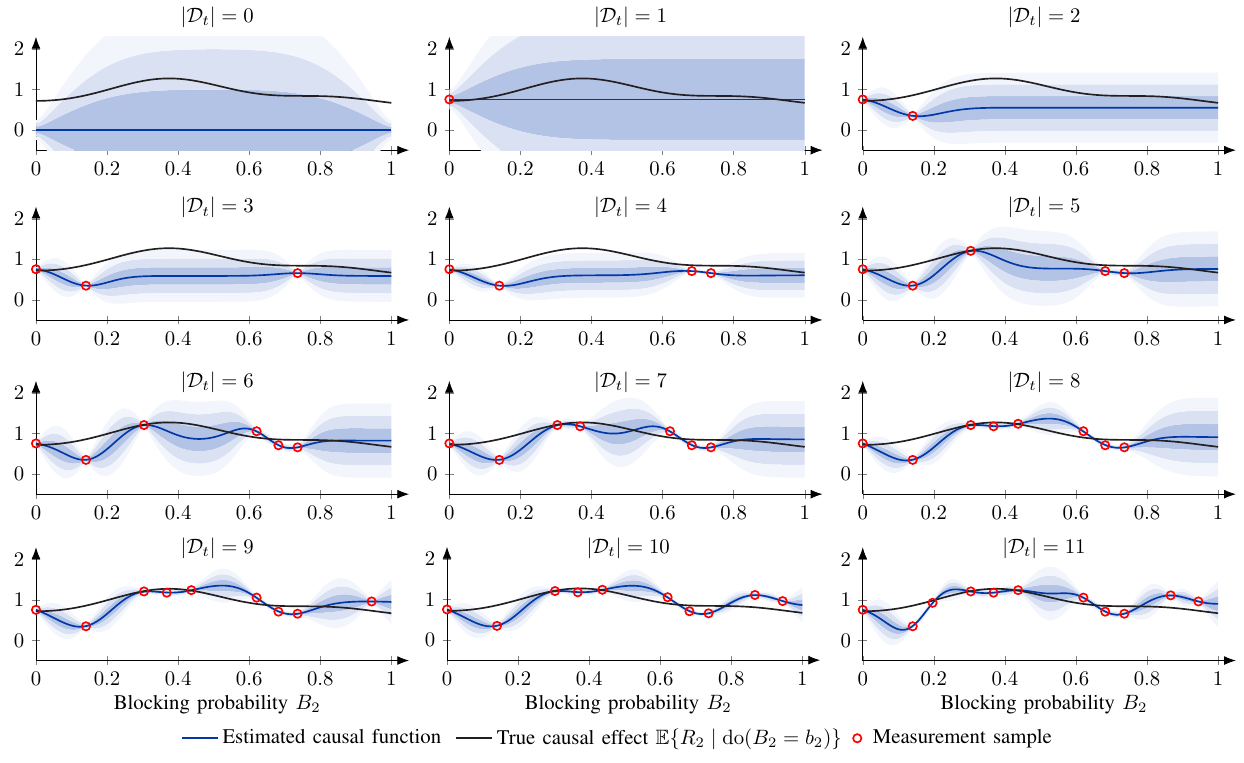}
  }
    \caption{\hyperref[scenario_1]{Scenario 1}: the system is stationary and the causal functions do not change. The figure shows the evolution of the causal effect $\mathbb{E}\{R_2 \mid \mathrm{do}(B_2=b_2)\}$ based on the functions $\hat{\mathbf{F}}_t$ estimated through (\ref{eq:estimator}) as the dataset $\mathcal{D}_t$ [cf.~(\ref{eq:dataset})] is updated using our \textbf{active learning method} with the rollout intervention policy; cf.~(\ref{eq:rollout}). Curves show the mean values; shaded regions indicate one, two, and three standard deviations from the mean (darker to lighter shades of blue).}
    \label{fig:learned_model_causal_evolution}
  \end{figure*}

The evaluation results for the two scenarios are detailed below. We present the results by visually comparing the estimated functions against the causal functions in \figref{fig:system_model} and showing how the loss $\mathscr{L}$ [cf.~(\ref{eq:loss_fun})] evolves as additional measurement samples are collected.

\vspace{2mm}

\noindent\textbf{\textit{\hyperref[scenario_1]{Scenario 1} (Stationary system)}.}
\Figref{fig:learned_model_passive} shows the causal functions estimated using the \textsc{gp} estimator $\varphi(\mathcal{D}_t)$ [cf.~(\ref{eq:estimator})] based on $30$ samples collected through passive learning. Because the data is collected without interventions, the data is confined to the system’s \textit{current} operating region; cf.~\defref{def:cor}. As a result, the estimated causal functions exhibit high uncertainty in unexplored parts of the operating region; cf.~\defref{def:or}.

\Figref{fig:learned_model_causal} shows the causal functions estimated using the \textsc{gp} estimator $\varphi(\mathcal{D}_t)$ [cf.~(\ref{eq:estimator})] based on $30$ samples collected with our active learning method, which uses the rollout intervention policy; cf.~(\ref{eq:rollout}). Unlike the passive learning policy, this policy selects interventions that explore diverse configurations of the system, which enables our method to collect data outside of the current operating region; cf.~\defref{def:cor}. As a result, the estimated functions better capture the system behavior.

\Figref{fig:learned_model_causal_evolution} shows the evolution of the causal effect $\mathbb{E}\{R_2\mid \mathrm{do}(B_2=b_2)\}$ based on the causal functions estimated via our method. Initially, the estimate is uncertain due to a lack of data. However, as new samples are collected, the uncertainty rapidly shrinks and the estimate converges toward the true causal function, as expected from \propref{prop:bayes_optimal} and \propref{prop:consistency}.

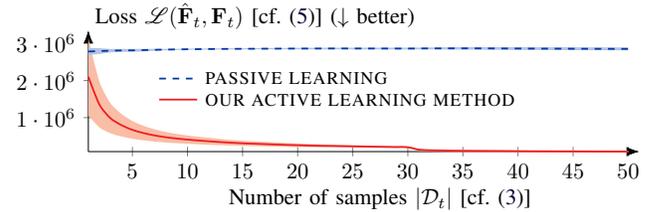
\begin{figure}[H]
  \centering
  \scalebox{0.84}{
    \begin{tikzpicture}

\pgfplotsset{/dummy/workaround/.style={/pgfplots/axis on top}}

\pgfplotstableread{
1 2782647.8179572984 2876343.6141801653 2688952.0217344314
2 2794201.007763573 2872344.386803995 2716057.628723151
3 2804771.304214983 2850993.3979914007 2758549.210438565
4 2811745.66707885 2852514.3646116504 2770976.9695460494
5 2814110.238868273 2852335.7403383874 2775884.7373981583
6 2820755.01462916 2854313.832641119 2787196.196617201
7 2828780.6685610553 2854156.826438289 2803404.5106838215
8 2834639.0526219243 2854109.8169286163 2815168.2883152324
9 2839397.8021350442 2855140.6634800844 2823654.940790004
10 2843754.6479841853 2859138.1283447375 2828371.167623633
11 2846931.740591906 2862425.3500025882 2831438.131181224
12 2849925.496871956 2865544.5365235833 2834306.457220329
13 2852617.6063432246 2867961.1267310884 2837274.085955361
14 2854461.6265042997 2869876.258311995 2839046.9946966046
15 2856352.3093149625 2872242.198968968 2840462.419660957
16 2857923.461757198 2874268.2437470546 2841578.6797673414
17 2859363.325499864 2876233.803033497 2842492.8479662305
18 2860458.060131333 2878279.937486776 2842636.1827758895
19 2861498.7982677436 2880242.2581750494 2842755.3383604377
20 2862139.7046373608 2882212.9780694046 2842066.431205317
21 2862785.602702557 2884016.367366397 2841554.8380387174
22 2863463.903627973 2885267.829486985 2841659.977768961
23 2864312.246836181 2886655.170895021 2841969.3227773407
24 2864436.1265244586 2886937.5756849153 2841934.677364002
25 2864071.2297794386 2886704.8275461504 2841437.6320127267
26 2863940.991094847 2886773.3127897466 2841108.669399948
27 2863773.2803215925 2887192.364571918 2840354.196071267
28 2863616.056644194 2887570.3797213854 2839661.733567002
29 2863022.6734864763 2887517.118986448 2838528.227986505
30 2862709.578479355 2887923.6655162205 2837495.49144249
31 2864950.702941316 2890009.3777129995 2839892.028169633
32 2866206.014375149 2891872.050181948 2840539.9785683495
33 2866861.0814046105 2896025.7449312173 2837696.417878004
34 2867370.606668152 2899170.9971096897 2835570.216226614
35 2868236.378800614 2902434.4720428153 2834038.2855584123
36 2867801.949813845 2903074.8074625237 2832529.0921651665
37 2866717.610470812 2903003.5813796753 2830431.6395619484
38 2865578.1214881875 2902446.0547509165 2828710.1882254584
39 2864463.40286607 2901693.3462616927 2827233.4594704476
40 2863144.2120362534 2900645.489127151 2825642.9349453556
41 2862028.1524573616 2899681.1070786444 2824375.1978360787
42 2860962.777410052 2898763.511118014 2823162.0437020897
43 2859838.62751203 2898060.936838932 2821616.318185128
44 2858999.557985715 2897572.7157273097 2820426.4002441205
45 2858123.1860138825 2896657.450956397 2819588.921071368
46 2857039.4155375334 2895760.184027 2818318.647048067
47 2855919.7647990817 2894978.459600265 2816861.0699978983
48 2855051.7952625593 2894440.844199398 2815662.7463257206
49 2854175.5084554628 2893790.5105767744 2814560.506334151
50 2853464.018924764 2893252.8846049984 2813675.1532445294
}\passive

\pgfplotstableread{
1 2102884.2007799745 3144526.326031795 1061242.0755281541
2 1357647.9451182203 1976926.5126897916 738369.3775466491
3 988278.6552181514 1396068.204544557 580489.1058917458
4 799012.2241835003 1103471.23126978 494553.21709722053
5 670396.305552625 909907.6006105687 430885.0104946813
6 585844.0505823864 781973.6097713433 389714.4913934296
7 521192.751698774 687658.6922013048 354726.81119624316
8 473371.8166492806 616105.1892365837 330638.44406197756
9 437149.98787379346 562064.3926745292 312235.5830730577
10 408155.41891866585 519768.5345800696 296542.3032572621
11 383744.345583769 485224.382674222 282264.30849331606
12 362092.1361611305 456226.2551032511 267958.0172190099
13 341381.6920261391 425416.15349481185 257347.23055746636
14 323810.2736819294 400144.4634752508 247476.08388860797
15 308614.5470354617 378515.3886077542 238713.7054631692
16 297767.48939474457 362457.79089869885 233077.1878907903
17 286732.88526955445 348654.8930312778 224810.8775078311
18 275543.3814658324 332571.9071925224 218514.8557391424
19 265645.2282055373 318193.4345287458 213097.02188232876
20 256785.9867229748 305134.444752382 208437.5286935676
21 248891.45362168824 293294.6015690552 204488.3056743213
22 241792.6956222305 282814.6912044891 200770.70003997194
23 237807.18287759408 272585.96772125637 203028.3980339318
24 232164.1180289412 263857.597445522 200470.63861236032
25 226572.98317690069 255759.61675165631 197386.34960214506
26 221400.90140927135 248236.77552880975 194565.02728973294
27 216641.5390848807 241385.13511851785 191897.94305124355
28 212072.49927356018 235215.9738793964 188929.02466772398
29 207682.90878273925 229525.56833667558 185840.24922880292
30 203486.99473057585 224200.0510178476 182773.9384433041
31 136105.86546918313 165019.64506901742 107192.08586934884
32 118496.40856278117 151552.16605660357 85440.65106895877
33 112862.43888379095 145443.98810447133 80280.88966311057
34 107893.60427774982 140286.620922831 75500.58763266866
35 105442.82171337916 137666.93745538828 73218.70597137004
36 102713.27948184122 134444.599801248 70981.95916243445
37 100997.86402716939 132599.8242916948 69395.90376264398
38 99199.4352475282 130425.64108391933 67973.22941113709
39 97100.64622813866 128035.55101298698 66165.74144329035
40 94822.71462403427 125343.7413589857 64301.68788908284
41 92697.77520485538 123086.8889502187 62308.66145949207
42 91087.0659702865 121722.50448162988 60451.627458943105
43 90568.37802260082 120587.0869461436 60549.66909905805
44 89825.60025067934 119320.14253966585 60331.057961692844
45 89234.9409963558 118638.46304504179 59831.41894766981
46 87225.04589537073 116290.37625591886 58159.71553482259
47 85963.8374205122 115836.65178770744 56091.02305331696
48 85651.60713463466 115129.08395642905 56174.130312840265
49 85235.03578840243 114218.62912291991 56251.44245388494
50 84822.49745233096 113147.51433934708 56497.48056531485
}\activee

\node[scale=1] (kth_cr) at (0,0)
{
\begin{tikzpicture}
  \begin{axis}
[
        xmin=1,
        xmax=51,
        ymax=3350000,
        ymin=84822,
        width=10.3cm,
        height=3.5cm,
        axis y line=center,
        axis x line=bottom,
        scaled y ticks=false,
        xlabel style={below right},
        ylabel style={above left},
        axis line style={-{Latex[length=2mm]}},
        smooth,
        legend style={at={(0.8,0.73)}},
        legend columns=1,
        legend style={
          align=left,
          /tikz/every node/.style={anchor=west},
          draw=none,
            /tikz/column 2/.style={
                column sep=5pt,
              }
              }
              ]
              \addplot[RoyalAzure,name path=l1, thick, dashed] table [x index=0, y index=1, domain=0:1] {\passive};
              \addplot[Red,name path=l1, thick] table [x index=0, y index=1, domain=0:1] {\activee};
\legend{\textsc{passive learning}, \textsc{our active learning method}}

              \addplot[draw=none,Black,mark repeat=2, name path=A, thick, domain=0:1] table [x index=0, y index=2] {\activee};
              \addplot[draw=none,Black,mark repeat=2, name path=B, thick, domain=0:1] table [x index=0, y index=3] {\activee};
              \addplot[Red!30!white] fill between [of=A and B];

              \addplot[draw=none,Black,mark repeat=2, name path=A, thick, domain=0:1] table [x index=0, y index=2] {\passive};
              \addplot[draw=none,Black,mark repeat=2, name path=B, thick, domain=0:1] table [x index=0, y index=3] {\passive};
              \addplot[RoyalAzure!30!white] fill between [of=A and B];
  \end{axis}
\node[inner sep=0pt,align=center, scale=1, rotate=0, opacity=1] (obs) at (4.68,-0.75)
{
  Number of samples $|\mathcal{D}_t|$ [cf. (\ref{eq:dataset})]
};
\node[inner sep=0pt,align=center, scale=1, rotate=0, opacity=1] (obs) at (2.7,2.15)
{
  Loss $\mathscr{L}(\hat{\mathbf{F}}_t, \mathbf{F}_t)$ [cf. (\ref{eq:loss_fun})] ($\downarrow$ better)
};
\end{tikzpicture}
};

\end{tikzpicture}        
  }
  \caption{Loss [cf.~(\ref{eq:loss_fun})] of the functions $\hat{\mathbf{F}}_t$ estimated through passive learning (blue curve) and our active learning method (red curve) when applied to \hyperref[scenario_1]{Scenario 1}. Curves show the mean value from evaluations with $5$ random seeds; shaded areas indicate standard deviations.}
  \label{fig:evaluation_loss}
\end{figure}

Lastly, \figref{fig:evaluation_loss} quantifies the accuracy of the estimated functions through the loss function $\mathscr{L}$; cf.~(\ref{eq:loss_fun}). We see in the figure that the loss of the functions estimated through passive learning plateaus and remains high. In contrast, the loss of the functions estimated through our active learning method reduces with each measurement sample. Specifically, the loss of passive learning (blue curve) decreases from around $2.8\cdot 10^{6}$ to $2.77\cdot 10^{6}$, which is negligible and barely visible in the plot. By comparison, the loss of our method decreases to $84\cdot 10^3$.

\myboxxx{\textbf{Takeaway from \hyperref[scenario_1]{Scenario 1}.}}{Black!5}{Black!2}{
Passive learning provides limited coverage of the system's operating region [cf.~\defref{def:or}], leading to inaccurate model estimates. Our active learning method overcomes this limitation by selecting interventions across the complete operating region.
}

\begin{figure*}
  \centering
  \scalebox{0.85}{
    \includegraphics{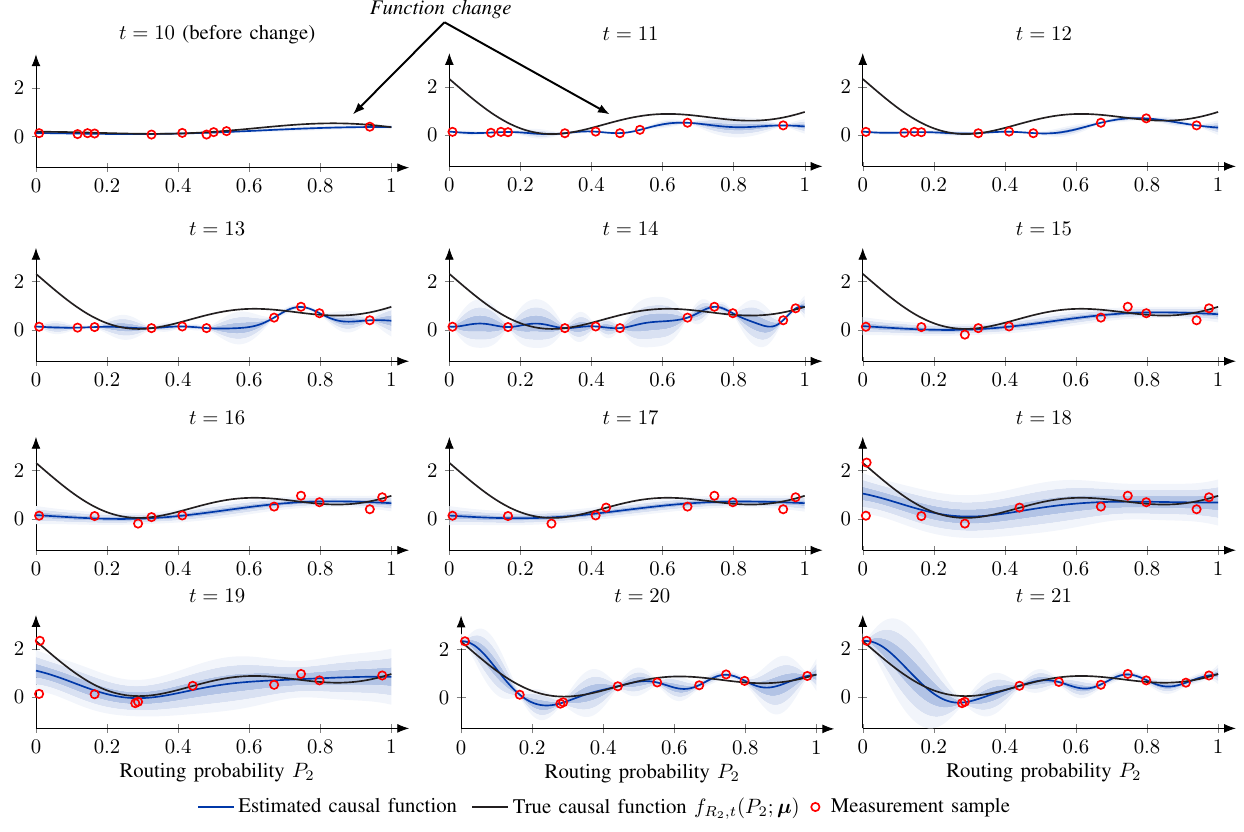}
  }
  \caption{\hyperref[scenario_2]{Scenario 2}: a change in the \textsc{it} system occurs between time steps $t=10$ and $t=11$, which causes the function $f_{R_2,t}$ to change. The dataset $\mathcal{D}_t$ [cf.~(\ref{eq:dataset})] is defined as a first-in-first-out buffer of size $10$ and is updated using our \textbf{active learning method} with the rollout intervention policy; cf.~(\ref{eq:rollout}). Curves show the mean values; shaded regions indicate one, two, and three standard deviations from the mean (darker to lighter shades of blue). The function $f_{R_2,t}(P_2; \bm{\mu})$ denotes $f_{R_2,t}(\Tilde{L}_1, \Tilde{L}_2, P_1, P_2, C_1, C_3, \epsilon_{R_i})$ [cf.~(\ref{eq:causal_functions_it_system})] evaluated with all inputs fixed to their mean values except for $P_2$.}
  \label{fig:adaptation_eval}
\end{figure*}

\vspace{2mm}

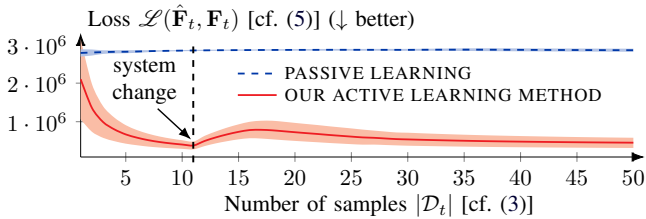
\begin{figure}[H]
  \centering
  \scalebox{0.86}{
    \begin{tikzpicture}

\pgfplotsset{/dummy/workaround/.style={/pgfplots/axis on top}}

\pgfplotstableread{
1 2782647.8179572984 2876343.6141801653 2688952.0217344314
2 2794201.007763573 2872344.386803995 2716057.628723151
3 2804771.304214983 2850993.3979914007 2758549.210438565
4 2811745.66707885 2852514.3646116504 2770976.9695460494
5 2814110.238868273 2852335.7403383874 2775884.7373981583
6 2820755.01462916 2854313.832641119 2787196.196617201
7 2828780.6685610553 2854156.826438289 2803404.5106838215
8 2834639.0526219243 2854109.8169286163 2815168.2883152324
9 2839397.8021350442 2855140.6634800844 2823654.940790004
10 2843754.6479841853 2859138.1283447375 2828371.167623633
11 2846931.740591906 2862425.3500025882 2831438.131181224
12 2849925.496871956 2865544.5365235833 2834306.457220329
13 2852617.6063432246 2867961.1267310884 2837274.085955361
14 2854461.6265042997 2869876.258311995 2839046.9946966046
15 2856352.3093149625 2872242.198968968 2840462.419660957
16 2857923.461757198 2874268.2437470546 2841578.6797673414
17 2859363.325499864 2876233.803033497 2842492.8479662305
18 2860458.060131333 2878279.937486776 2842636.1827758895
19 2861498.7982677436 2880242.2581750494 2842755.3383604377
20 2862139.7046373608 2882212.9780694046 2842066.431205317
21 2862785.602702557 2884016.367366397 2841554.8380387174
22 2863463.903627973 2885267.829486985 2841659.977768961
23 2864312.246836181 2886655.170895021 2841969.3227773407
24 2864436.1265244586 2886937.5756849153 2841934.677364002
25 2864071.2297794386 2886704.8275461504 2841437.6320127267
26 2863940.991094847 2886773.3127897466 2841108.669399948
27 2863773.2803215925 2887192.364571918 2840354.196071267
28 2863616.056644194 2887570.3797213854 2839661.733567002
29 2863022.6734864763 2887517.118986448 2838528.227986505
30 2862709.578479355 2887923.6655162205 2837495.49144249
31 2864950.702941316 2890009.3777129995 2839892.028169633
32 2866206.014375149 2891872.050181948 2840539.9785683495
33 2866861.0814046105 2896025.7449312173 2837696.417878004
34 2867370.606668152 2899170.9971096897 2835570.216226614
35 2868236.378800614 2902434.4720428153 2834038.2855584123
36 2867801.949813845 2903074.8074625237 2832529.0921651665
37 2866717.610470812 2903003.5813796753 2830431.6395619484
38 2865578.1214881875 2902446.0547509165 2828710.1882254584
39 2864463.40286607 2901693.3462616927 2827233.4594704476
40 2863144.2120362534 2900645.489127151 2825642.9349453556
41 2862028.1524573616 2899681.1070786444 2824375.1978360787
42 2860962.777410052 2898763.511118014 2823162.0437020897
43 2859838.62751203 2898060.936838932 2821616.318185128
44 2858999.557985715 2897572.7157273097 2820426.4002441205
45 2858123.1860138825 2896657.450956397 2819588.921071368
46 2857039.4155375334 2895760.184027 2818318.647048067
47 2855919.7647990817 2894978.459600265 2816861.0699978983
48 2855051.7952625593 2894440.844199398 2815662.7463257206
49 2854175.5084554628 2893790.5105767744 2814560.506334151
50 2853464.018924764 2893252.8846049984 2813675.1532445294
}\passive

\pgfplotstableread{
1 2102884.2007799745 3144526.326031795 1061242.0755281541
2 1357647.9451182203 1976926.5126897916 738369.3775466491
3 988278.6552181514 1396068.204544557 580489.1058917458
4 799012.2241835003 1103471.23126978 494553.21709722053
5 670396.305552625 909907.6006105687 430885.0104946813
6 585844.0505823864 781973.6097713433 389714.4913934296
7 521192.751698774 687658.6922013048 354726.81119624316
8 473371.8166492806 616105.1892365837 330638.44406197756
9 437149.98787379346 562064.3926745292 312235.5830730577
10 408155.41891866585 519768.5345800696 296542.3032572621
11 383744.345583769 485224.382674222 282264.30849331606
12 501331.9388036585 634088.3474565308 368636.314173964
13 594581.9090480969 748090.8760098699 439819.34883336636
14 672419.5775689951 848336.2908776737 495862.81142916385
15 737270.7568963566 931158.6675811156 543019.1503355203
16 784070.6684360135 998452.5035805966 572056.0239847635
17 789940.5358804469 1025065.1110555254 563638.9417320819
18 776486.0703226539 1024320.7628992248 542947.0492530925
19 752200.2671837845 999106.3583930313 521437.9633887818
20 727355.0460219153 969005.6097034672 502290.5482927389
21 703872.9969958192 940605.949116571 484160.7964118622
22 681612.2060116959 913298.5147160224 467229.2823092327
23 657038.8172758478 882190.4456516667 449178.3550813167
24 635294.0267380334 853983.3023235284 433656.3934161913
25 613638.2349376526 826051.2365072508 418092.03226986545
26 592838.3840354097 795743.1189169114 405452.19701538904
27 582666.5284663937 769515.0086367275 406840.0793543102
28 565334.3782835208 744280.3566834048 396293.87683303724
29 549158.3017419428 720876.8213097104 386352.9474143817
30 540991.407869695 714430.8271412586 377770.6145287146
31 532991.000000 702482.138000 372560.709000
32 525491.000000 691546.156000 367318.209000
33 518491.000000 681297.174000 362425.209000
34 511991.000000 671732.192000 357881.709000
35 505991.000000 662848.210000 353687.709000
36 500491.000000 654642.228000 349843.209000
37 495491.000000 647111.246000 346348.209000
38 490991.000000 640252.264000 343202.709000
39 486991.000000 634062.282000 340406.709000
40 483191.000000 628148.300000 337750.509000
41 479591.000000 622509.118000 335234.109000
42 476191.000000 617143.536000 332857.509000
43 472991.000000 612050.354000 330620.709000
44 469991.000000 607228.372000 328523.709000
45 467191.000000 602676.390000 326566.509000
46 464591.000000 598393.208000 324749.109000
47 462191.000000 594377.626000 323071.509000
48 459991.000000 590628.444000 321533.709000
49 457991.000000 587144.462000 320135.709000
50 456191.000000 583924.480000 318877.509000

}\activee

\node[scale=1] (kth_cr) at (0,0)
{
\begin{tikzpicture}
  \begin{axis}
[
        xmin=1,
        xmax=51,
        ymax=3300000,
        ymin=84822,
        width=10.3cm,
        height=3.5cm,
        axis y line=center,
        axis x line=bottom,
        scaled y ticks=false,
        xlabel style={below right},
        ylabel style={above left},
        axis line style={-{Latex[length=2mm]}},
        smooth,
        legend style={at={(0.95,0.8)}},
        legend columns=1,
        legend style={
          align=left,
          /tikz/every node/.style={anchor=west},
          draw=none,
            /tikz/column 2/.style={
                column sep=5pt,
              }
              }
              ]
              \addplot[RoyalAzure,name path=l1, thick, dashed] table [x index=0, y index=1, domain=0:1] {\passive};
              \addplot[Red,name path=l1, thick] table [x index=0, y index=1, domain=0:1] {\activee};
\legend{\textsc{passive learning}, \textsc{our active learning method}}

              \addplot[draw=none,Black,mark repeat=2, name path=A, thick, domain=0:1] table [x index=0, y index=2] {\activee};
              \addplot[draw=none,Black,mark repeat=2, name path=B, thick, domain=0:1] table [x index=0, y index=3] {\activee};
              \addplot[Red!30!white] fill between [of=A and B];

              \addplot[draw=none,Black,mark repeat=2, name path=A, thick, domain=0:1] table [x index=0, y index=2] {\passive};
              \addplot[draw=none,Black,mark repeat=2, name path=B, thick, domain=0:1] table [x index=0, y index=3] {\passive};
              \addplot[RoyalAzure!30!white] fill between [of=A and B];
  \end{axis}
\node[inner sep=0pt,align=center, scale=1, rotate=0, opacity=1] (obs) at (4.68,-0.75)
{
  Number of samples $|\mathcal{D}_t|$ [cf. (\ref{eq:dataset})]
};
\node[inner sep=0pt,align=center, scale=1, rotate=0, opacity=1] (obs) at (2.7,2.15)
{
  Loss $\mathscr{L}(\hat{\mathbf{F}}_t, \mathbf{F}_t)$ [cf. (\ref{eq:loss_fun})] ($\downarrow$ better)
};
\end{tikzpicture}
};

\node[inner sep=0pt,align=center, scale=1, rotate=0, opacity=1] (obs) at (-2.8,0.45)
{  
  system\\change
};

\draw[-, black, thick, line width=0.3mm, dashed] (-2.05,-0.8) to (-2.05,1);
\draw[-{Latex[length=2mm]}, line width=0.22mm, color=black] (-2.6,0.03) to (-2.08,-0.45);
\end{tikzpicture}      
  }
  \caption{Loss [cf.~(\ref{eq:loss_fun})] of the functions $\hat{\mathbf{F}}_t$ estimated through passive learning (blue curve) and our active learning method (red curve) when applied to \hyperref[scenario_2]{Scenario 2}. Curves show the mean value from evaluations with $5$ random seeds; shaded areas indicate standard deviations.}
  \label{fig:evaluation_loss_2}
\end{figure}

\noindent\textbf{\textit{\hyperref[scenario_2]{Scenario 2} (Non-stationary system)}.}
Unlike \hyperref[scenario_1]{Scenario 1}, where the system is stationary, this scenario involves a change in the system between time steps $=10$ and $t=11$. \Figref{fig:evaluation_loss_2} shows the model accuracy, as quantified by the loss function $\mathscr{L}$; cf.~(\ref{eq:loss_fun}). As in \hyperref[scenario_1]{Scenario 1}, we find that models estimated based on passive learning have persistently high loss. In contrast, the loss of the model estimated via our active learning method is steadily decreasing as more data is collected.

At time step $t=11$, the loss temporarily increases due to the system change. However, our method adapts by updating the model to reflect the new dynamics. \Figref{fig:adaptation_eval} illustrates this adaptation by showing how the estimate of the causal function $f_{R_2,t}$ [cf.~(\ref{eq:causal_functions_it_system})] evolves from $t=10$ to $t=21$. Initially, the estimate differs substantially from the true function, but by $t=21$, it closely aligns with the updated system behavior.

\myboxxx{\textbf{Takeaway from \hyperref[scenario_2]{Scenario 2}.}}{Black!5}{Black!2}{
Unlike offline identification methods, which use a fixed dataset to estimate the model, our method estimates the model sequentially and selects interventions based on the uncertainty in the current estimate. This approach allows for quick adaptation of the model to system changes.
}

\section{Use Cases of the Identified Causal Model}
Once the causal functions $\mathbf{F}_t$ [cf.~(\ref{eq:scm_def})] have been identified with sufficiently high confidence through our method, they can be used for optimizing various downstream tasks in \textsc{it} systems. We give several examples of such use cases below.

\vspace{2mm}

\noindent\textbf{\textit{Forecasting system metrics}.}
The causal functions can be used to predict how key performance indicators evolve in response to changes in system variables, such as service loads and routing probabilities. For example, the function $f_{R_1,t}$ of the \textsc{scm} in \sectionref{sec:it_system_case} predicts the response time of service $S_1$ based on the load, routing configuration, and \textsc{cpu} allocation.

\vspace{2mm}

\noindent\textbf{\textit{Anomaly detection}.}
The causal functions can be used to identify abnormal system behavior. For instance, in the context of the \textsc{scm} described in \sectionref{sec:it_system_case}, if the response time $R_i$ deviates significantly from what the function $f_{R_i,t}$ predicts given the current load and configuration, this may indicate an ongoing performance anomaly or a potential cyberattack \cite{hammar_stadler_tnsm,hammar_stadler_tnsm_23}.

\vspace{2mm}

\noindent\textbf{\textit{Automatic control}.}
The causal functions enable simulation-based optimization of control policies through reinforcement learning. For instance, the functions described in \sectionref{sec:it_system_case} define how varying the \textsc{cpu} counts $C_1$ and $C_3$ will affect the response times $R_1$ and $R_2$. This cause-and-effect relationship can be used to optimize a resource allocation policy that dynamically scales $C_1$ and $C_3$ to keep the response times below a threshold.

\vspace{2mm}

\noindent\textbf{\textit{Root cause analysis}.}
The causal functions can support the diagnosis of system failures by tracing observed performance degradations or anomalies back to their causal origins.

\vspace{2mm}
\noindent\textbf{\textit{Digital twin}.}
The causal functions can be used to build a digital twin, i.e., a virtual replica of the \textsc{it} system that provides a controlled environment for virtual operations \cite{digital_twins_kim,kim_phd_thesis}. By using our method to periodically update the causal model based on new measurements, the digital twin can simulate the impact of hypothetical changes to the \textsc{it} system (e.g., workload shifts, configuration updates, policy changes, or security interventions) before applying them to the \textsc{it} system.

\section{Discussion of Design Choices in Our Method}
Our method involves two main steps: (\textit{i}) estimation of causal functions through \textsc{gp} regression; and (\textit{ii}) selection of interventions through rollout and lookahead optimization. The main reason for using \textsc{gp} regression is that it quantifies the uncertainty in its estimates, which we use to guide the selection of interventions. Alternatives to \textsc{gp} regression include mixture density networks (\textsc{mdn}s) \cite{9422765} and Bayesian neural networks (\textsc{bnn}s) \cite{10639342}, both of which provide function estimates with uncertainty quantification. Compared to these methods, the main advantage of \textsc{gp}s is that they come with theoretical guarantees: under general conditions, they can approximate any continuous function arbitrarily well; see \propref{prop:consistency}. By contrast, \textsc{mdn}s and \textsc{bnn}s typically require careful tuning of the neural network architecture to obtain accurate estimates.

On the other hand, \textsc{mdn}s and \textsc{bnn}s scale more favorably to large datasets than \textsc{gp}s. In particular, the computational complexity of \textsc{gp} regression is cubic in the size of the dataset \cite{Rasmussen2006Gaussian}. However, this complexity can be addressed using sparse \textsc{gp} approximations \cite{pmlr-v2-snelson07a} or neural \textsc{gp} methods \cite{pmlr-v31-damianou13a}, which reduce complexity while retaining most of the predictive accuracy. Another practical approach is to use a first-in-first-out buffer to bound the dataset size during online learning, as we did in the experiments related to \hyperref[scenario_2]{Scenario~2}.

In addition to \textsc{mdn}s and \textsc{bnn}s, another alternative to \textsc{gp}s is conformal prediction (also known as hedged prediction \cite{8129828}), which can be applied on top of any function estimator to provide uncertainty quantification \cite{10.5555/1062391}. For instance, random forest regressors or feedforward neural networks can serve as base estimators for conformal prediction. Compared to \textsc{gp}s, these models may be easier to scale to large datasets, but they generally lack the convergence guarantees of \textsc{gp}s.

Regarding the selection of interventions, the main alternatives to our rollout method [cf.~(\ref{eq:rollout})] are random selection and heuristic selection. Examples of such methods include $\epsilon$-greedy, Thompson sampling, upper-confidence-bound strategies, and myopic one-step lookahead policies; see textbook \cite{Garnett_2023} for a comprehensive overview of these methods. Compared to these methods, the advantages of our rollout method are a) that it uses the \textsc{gp}'s uncertainty estimates to drive exploration of the system's (complete) operating region [cf.~\defref{def:or}]; and b) that it is couched on well-established theory; see \cite{bertsekas2021rollout,bertsekas2022rolloutalgorithmsapproximatedynamic}. (Note that the computation of an optimal intervention policy is intractable, which is why exact dynamic programming methods are not a viable alternative to rollout.)
\section{Related Work}
System identification has a long history in the control systems community, where the primary goal is to build mathematical models of dynamic systems from observed input-output data \cite{Ljung1998,aastrom1971system,wahlberg2002system,9946382}. In parallel, related ideas have been studied in the artificial intelligence (\textsc{ai}) and reinforcement learning communities, where the task is often referred to as model learning rather than system identification \cite{araya2011active,rl_bible,sutton1991dyna}. However, there are essential differences between these classical approaches and our method. First, most existing work in both control and \textsc{ai} focuses on physical or simulated control systems, whereas our emphasis is on modeling \textsc{it} systems. Second, traditional methods are generally offline and assume access to large batches of data, while our approach is designed for online learning of a causal model. The benefit of our approach is that it allows to quickly adapt the model to system changes, such as service migrations or software updates.

Online system identification has received attention in the context of adaptive control and dual control. In this line of research, the goal is to design methods for simultaneously identifying an unknown dynamical system while controlling it \cite{7051249,1103122,126844,10.5555/546778,1100241,704978,9044339,ASTROM1973185,10.5555/3586589.3586621,1104847}. Adaptive control typically identifies the system based on measurements from the current operating region, whereas dual control explicitly selects controls to explore the complete operating region. The main difference between these works and our paper is that we focus on learning a structural causal model, whereas the referenced works focus on learning a dynamical system model, such as a linear system \cite{9946382} or a Markov decision process \cite{7051249}. Moreover, our problem formulation is different. Our objective is to accurately learn the underlying (causal) system model from system measurements. By contrast, the objective in most of the referenced works is to simultaneously learn a system model and a control policy.

Online causal learning has been studied in the contexts of causal Bayesian optimization and causal discovery. In causal Bayesian optimization, the objective is to identify interventions that maximize a target variable in a causal model with known structure but unknown dynamics \cite{cbo,dcbo,mcbo,Zhang2023,pmlr-v216-gultchin23a}. While methods designed for such problems typically involve learning the causal functions, this is not the main goal. Rather, the main goal is to learn just enough about the causal functions to identify an optimal intervention. Causal discovery, on the other hand, seeks to learn the causal structure from data \cite{JMLR:v9:he08a,NEURIPS2019_5ee56059,10647072,Wang_Rios_Jha_Shanmugam_Bagehorn_Yang_Filepp_Abe_Shwartz_2024}. This line of work differs from our paper in that we aim to learn the causal functions rather than the causal structure.

Prior work that focuses on the same problem as us includes \cite{RubTolHenSch17,pmlr-v139-linzner21a,NEURIPS2020_45c166d6,NEURIPS2022_675e371e} and \cite{NIPS2000_0731460a}. Compared to these papers, the main difference is that our method is designed explicitly for \textsc{it} systems, whereas the referenced papers focus on other types of systems, e.g., healthcare systems \cite{NEURIPS2020_45c166d6}. Moreover, our intervention policy is based on rollout, whereas the referenced papers use (myopic) heuristic intervention policies.

Lastly, we note that a growing body of research applies causal modeling to various decision-making problems that arise in the operation of \textsc{it} systems, particularly in cybersecurity \cite{hammar2024optimaldefenderstrategiescage2,andrew2022developing} and root cause analysis \cite{yagemann2021arcus,NEURIPS2022_c9fcd02e,liu2018towards,zhang2016causality}. However, these approaches assume the existence of a causal model and focus on specific decision-making tasks. In contrast, our method addresses the more general problem of learning the causal system model itself, which can then support a wide range of downstream tasks in \textsc{it} systems.

\section{Conclusion}
A longstanding goal in systems engineering and operation is to automate network and service management tasks. We argue that a key component to achieve such automation is a causal model that explains how changes to system variables affect system behavior. Traditionally, such models have been designed by domain experts, which does not scale with the growing complexity and increasing dynamism of \textsc{it} systems.

This paper presents a method for online, data-driven identification of a causal system model. The main idea is to learn the system dynamics through \textit{active causal learning}, where a rollout policy selects targeted interventions that generate data to update the model via Gaussian process regression. We show that this method is Bayes-optimal (\propref{prop:bayes_optimal}), asymptotically consistent (\propref{prop:consistency}), and that the intervention policy has the cost improvement property (\propref{prop:bertsekas}). Testbed experiments demonstrate that our method quickly identifies accurate models of dynamic systems at a low operational cost.

\vspace{2mm}
\noindent\textbf{\textit{Future work}.}
From a practical point of view, an important direction for future work is to investigate further use cases of causal models identified through our method, such as real-time control, diagnosis, and forecasting in \textsc{it} systems. We have not presented a thorough study of such tasks in this paper to keep the focus on presenting the core method.

From a theoretical perspective, a natural extension of this work is to generalize our method by relaxing certain assumptions. In particular, the current problem formulation assumes that the causal graph and the distribution of exogenous variables are fixed and known. The first assumption can be relaxed by incorporating causal discovery methods for identifying the graph from data. The second assumption is purely for ease of exposition; removing it has minimal effect on implementation and theory. In particular, if the distribution of exogenous variables is unknown and time varying, it can be learned with the same method we use for learning the causal functions. The only notable change is that the loss function $\mathscr{L}$ [cf.~(\ref{eq:loss_fun})] and the surrogate loss function $\mathcal{L}$ [cf.~(\ref{eq:surrogate_loss})] must be updated to include a term that quantifies the accuracy of the estimated distribution, e.g., based on the Kullback-Leibler divergence.

\section*{Acknowledgments}
This research is supported by the Swedish Research Council under contract 2024-06436. The authors would like to thank Forough Shahab Samani for her help in setting up the testbed.

\appendices

\section{Notation}
Our notation is summarized in \tableref{tab:notation}.
\begin{table}
  \centering
  \scalebox{0.8}{
    \begin{tabular}{ll} \toprule
\rowcolor{lightgray}
    {\textit{Notation(s)}} & {\textit{Description}} \\ \midrule
      $\mathscr{M}_t, \mathcal{G}$ & Structural causal model, causal graph; cf.~(\ref{eq:scm_def}).\\
      $\mathbf{U}, \mathbf{V}, \mathbf{F}_t$ & Exo/endo-genous variables and functions of \textsc{scm} $\mathscr{M}_t$; cf.~(\ref{eq:scm_def}).\\
      $\mathbf{X}, \mathbf{N}$ & Controllable/non-controllable variables of \textsc{scm} $\mathscr{M}_t$; cf.~(\ref{eq:scm_def}).\\
      $\varphi, \pi$ & Estimator [cf.~(\ref{eq:estimator})] and intervention policy; cf.~(\ref{eq:objective}). \\
      $\mathscr{L}, \mathcal{D}_t$ & Loss function [cf.~(\ref{eq:loss_fun})] and dataset [cf.~(\ref{eq:dataset})]. \\
      $c, \gamma$ & Cost function and discount factor; cf.~(\ref{eq:objective}). \\
      $\mathcal{L}, g$ & Expected loss and cost functions; cf.~(\ref{eq:cost_dynamics}).\\
      $\ell, m$ & Rollout and lookahead horizons; cf.~(\ref{eq:rollout}).\\
      $L$ & Number of rollouts cf.~(\ref{eq:rollout_approximation}).\\
      $\pi, \tilde{\pi}$ & Base and rollout policies; cf.~(\ref{eq:rollout}).\\      
      $\mathbf{b}_t, \mathbf{z}_t$ & Belief state and measurement; cf.~(\ref{eq:belief_dynamics}).\\
      $u_t$ & Rollout control (intervention) at time $t$; cf.~(\ref{eq:belief_dynamics}).\\
      $J_{\pi}$ & Cost-to-go function of the policy $\pi$; cf.~(\ref{eq:bellman_2}).\\
      $\bm{\mu}, \mathbf{\Sigma}$ & Mean vector and covariance matrix; cf.~\appendixref{appendix:gps}.\\      
      $\mathcal{N}(\bm{\mu}, \mathbf{\Sigma})$ & Multivariate Gaussian distribution; cf.~\appendixref{appendix:gps}.\\
      $m_i,k_i$ & Mean and covariance functions of a Gaussian process; cf.~\sectionref{sec:gps_reg}.\\          
      $r \sim \mathcal{GP}(m_i,k_i)$ & Sampling a function $r$ from a Gaussian process; cf.~\sectionref{sec:gps_reg}.\\
      $\mathcal{GP}(m_{i|\mathcal{D}},k_{i|\mathcal{D}})$ & Posterior Gaussian process after observing $\mathcal{D}$; cf.~\sectionref{sec:gps_reg}.\\
      $\mathrm{do}(\mathbf{X}=\mathbf{x})$ & Intervention assigning the values $\mathbf{x}$ to the variables in $\mathbf{X}$; cf.~\sectionref{sec:online_prob_def}.\\
      $\mathrm{do}(\emptyset)$ & The passive intervention; cf.~\sectionref{sec:online_prob_def}.\\
      $M$ & The number of samples obtained after an intervention; cf.~\sectionref{sec:online_prob_def}.\\
      $\mathcal{O}$ & The (complete) operating region; cf.~\defref{def:or}.\\
      $\mathcal{O}_t$ & The current operating region; cf.~\defref{def:cor}.\\            
    \bottomrule\\
  \end{tabular}}
  \caption{Notation.}\label{tab:notation}
\end{table}

\section{Proof of \Propref{prop:bayes_optimal}}\label{appendix:proof_prop1}
For ease of notation we write $\hat{\mathbf{F}}, f_{V_i}, \mathbf{F}, \hat{f}_{V_i}$ instead of $\hat{\mathbf{F}}_t, f_{V_i,t}, \mathbf{F}_t, \hat{f}_{V_i,t}$. Moreover, we write $\mathbf{x}$ instead of $\mathbf{x}_i$. We seek to find the estimator $\hat{\mathbf{F}}$ that minimizes the expected loss $\mathbb{E}_{\mathbf{F} \sim \varphi(\mathcal{D}_t)}\{\mathscr{L}(\hat{\mathbf{F}}, \mathbf{F})\}$. We start by expanding this loss using the definition of $\mathscr{L}$ [cf.~(\ref{eq:loss_fun})], which gives
\begin{align*}
&\mathbb{E}_{\mathbf{F} \sim \varphi(\mathcal{D}_t)}\{\mathscr{L}(\hat{\mathbf{F}}, \mathbf{F})\}\\
&=\mathbb{E}_{\mathbf{F} \sim \varphi(\mathcal{D}_t)}\left\{\sum_{V_i \in \mathbf{V}}\int_{\mathcal{R}(\mathrm{pa}_{\mathcal{G}}(V_i))}\left(f_{V_i}(\mathbf{x}) - \hat{f}_{V_i}(\mathbf{x})\right)^2\mathbb{P}[\mathrm{d}\mathbf{x}]\right\}\\
&=\sum_{V_i \in \mathbf{V}}\int_{\mathcal{R}(\mathrm{pa}_{\mathcal{G}}(V_i))}\mathbb{E}_{\mathbf{F} \sim \varphi(\mathcal{D}_t)}\left\{\left(f_{V_i}(\mathbf{x}) - \hat{f}_{V_i}(\mathbf{x})\right)^2\right\}  \mathbb{P}[\mathrm{d}\mathbf{x}]\\
&=\sum_{V_i \in \mathbf{V}}\int_{\mathcal{R}(\mathrm{pa}_{\mathcal{G}}(V_i))}\mathbb{E}_{f_{V_i}\sim \mathcal{GP}(m_{i|\mathcal{D}_t}, k_{i|\mathcal{D}_t})}\bigg\{\\
  &\quad\quad\quad\quad\quad\quad\quad\quad\quad\quad\quad\quad\left(f_{V_i}(\mathbf{x}) - \hat{f}_{V_i}(\mathbf{x})\right)^2\bigg\}\mathbb{P}[\mathrm{d}\mathbf{x}].
\end{align*}
Applying standard \textsc{gp} properties, we decompose $f_{V_i}|\mathcal{D}_t$ as
\begin{align*}
(f_{V_i}|\mathcal{D}_t)(\mathbf{x}) = m_{i|\mathcal{D}_t}(\mathbf{x}) + h_{i}(\mathbf{x}) \quad \text{for all }\mathbf{x}\text{ and }V_i,
\end{align*}
where $h_i \sim \mathcal{GP}(m_0, k_{i|\mathcal{D}_t})$ and $m_0(\mathbf{x})=0$ for all $\mathbf{x}$. Leveraging this decomposition, we can rewrite the loss as
\begin{align*}
&\mathbb{E}_{\mathbf{F} \sim \varphi(\mathcal{D}_t)}\{\mathscr{L}(\hat{\mathbf{F}}, \mathbf{F})\} =\sum_{V_i \in \mathbf{V}}\int_{\mathcal{R}(\mathrm{pa}_{\mathcal{G}}(V_i))}\bigg(\\
&\mathbb{E}_{h_i\sim \mathcal{GP}(m_0, k_{i|\mathcal{D}_t})}\left\{\left(m_{i|\mathcal{D}_t}(\mathbf{x}) + h_{i}(\mathbf{x}) - \hat{f}_{V_i}(\mathbf{x})\right)^2\right\}\bigg)\mathbb{P}[\mathrm{d}\mathbf{x}]\\
&=\sum_{V_i \in \mathbf{V}}\int_{\mathcal{R}(\mathrm{pa}_{\mathcal{G}}(V_i))}\mathbb{E}_{h_i\sim \mathcal{GP}(m_0, k_{i|\mathcal{D}_t})}\bigg\{\left(m_{i|\mathcal{D}_t}(\mathbf{x})-\hat{f}_{V_i}(\mathbf{x})\right)^2\\
& + 2h_{i}(\mathbf{x})\left(m_{i|\mathcal{D}_t}(\mathbf{x})-\hat{f}_{V_i}(\mathbf{x})\right) + \left(h_{i}(\mathbf{x})\right)^2\bigg\} \mathbb{P}[\mathrm{d}\mathbf{x}]\\
&=\sum_{V_i \in \mathbf{V}}\int_{\mathcal{R}(\mathrm{pa}_{\mathcal{G}}(V_i))}\Bigg(\left(m_{i|\mathcal{D}_t}(\mathbf{x})-\hat{f}_{V_i}(\mathbf{x})\right)^2\\
&\quad\quad + 2\mathbb{E}_{h_i\sim \mathcal{GP}(m_0, k_{i|\mathcal{D}_t})}\{h_{i}(\mathbf{x})\}\left(m_{i|\mathcal{D}_t}(\mathbf{x})-\hat{f}_{V_i}(\mathbf{x})\right) + \\
&\quad\quad\quad\quad\quad \mathbb{E}_{h_i\sim \mathcal{GP}(m_0, k_{i|\mathcal{D}_t})}\left\{\left(h_{i}(\mathbf{x})\right)^2\right\}\Bigg)\mathbb{P}[\mathrm{d}\mathbf{x}]\\
&=\sum_{V_i \in \mathbf{V}}\int_{\mathcal{R}(\mathrm{pa}_{\mathcal{G}}(V_i))}\Bigg(\left(m_{i|\mathcal{D}_t}(\mathbf{x})-\hat{f}_{V_i}(\mathbf{x})\right)^2\\
& \quad\quad\quad\quad\quad\quad\quad\quad\quad\quad + k_{i|\mathcal{D}_t}(\mathbf{x}, \mathbf{x})\Bigg)\mathbb{P}[\mathrm{d}\mathbf{x}].
\end{align*}
Clearly, the minimizer of this expression is obtained by setting $\hat{f}_{V_i}=m_{i|\mathcal{D}_t}(\mathbf{x})$ for all $V_i$ and $\mathbf{x}$. As a consequence, we have
\begin{align*}
\mathbb{E}_{\mathbf{F} \sim \varphi(\mathcal{D}_t)}\{\mathbf{F}\} \in \argmin_{\hat{\mathbf{F}}}\mathbb{E}_{\mathbf{F} \sim \varphi(\mathcal{D}_t)}\left\{\mathscr{L}(\hat{\mathbf{F}}, \mathbf{F})\right\}.
\end{align*}
\qed
\section{Experimental Setup}\label{appendix:hyperparameters}
All computations are performed using an \textsc{m4 pro} chip. We approximate the integrals in (\ref{eq:cost_dynamics}) using $100$ Monte-Carlo samples. The hyperparameters we use for the evaluation are listed in \tableref{tab:hyperparams} and are selected based on random search.

\begin{table}[H]
  \centering
  \scalebox{0.9}{
    \begin{tabular}{ll} \toprule
\rowcolor{lightgray}
    {\textit{Parameter}} & {\textit{Value}} \\ \midrule
      Monte-carlo samples $L$ [cf.~(\ref{eq:rollout_approximation})]  & 10\\
      Minimizer of (\ref{eq:rollout}) & Differential evolution (\textsc{de}) \cite[Fig. 3]{differential_evolution}\\
      Population size of \textsc{de} & $10$\\ 
      Iterations of \textsc{de} & $30$\\      
    \bottomrule\\
  \end{tabular}}
  \caption{Hyperparameters.}\label{tab:hyperparams}
\end{table}

\section{Gaussian Process Formulas}\label{appendix:gps}
A Gaussian process (\textsc{gp}) is a collection of random variables, any finite number of which have a joint Gaussian distribution; see \cite[Def. 2.1]{Rasmussen2006Gaussian} for a formal definition. It generalizes the Gaussian distribution and is specified by its mean function $m(\mathbf{x})$ and covariance function $k(\mathbf{x},\mathbf{x}^{\prime})$. Suppose that we want to estimate the output of the causal function $f_{V_i,t} \in \mathbf{F}_t$ [cf.~(\ref{eq:scm_def})] at inputs $\mathbf{x}_{*}\triangleq (\mathbf{x}_1,\hdots,\mathbf{x}_n)$. The mean and covariance functions can be used to define the Gaussian distribution $\mathbf{f}_{*}\triangleq (f_{V_i,t}(\mathbf{x}_1),\hdots,f_{V_i,t}(\mathbf{x}_n)) \sim \mathcal{N}(\bm{\mu}, \bm{\Sigma})$, where $\bm{\mu} \triangleq (m(\mathbf{x}_1), \hdots, m(\mathbf{x}_{n}))$ and
\begin{align*}
\bm{\Sigma} \triangleq K(\mathbf{x}_{*},\mathbf{x}_{*})\triangleq
\begin{bmatrix}
k(\mathbf{x}_1, \mathbf{x}_1) & k(\mathbf{x}_1, \mathbf{x}_2) & \cdots & k(\mathbf{x}_1, \mathbf{x}_n) \\
k(\mathbf{x}_2, \mathbf{x}_1) & k(\mathbf{x}_2, \mathbf{x}_2) & \cdots & k(\mathbf{x}_2, \mathbf{x}_n) \\
\vdots & \vdots & \ddots & \vdots \\
k(\mathbf{x}_n, \mathbf{x}_1) & k(\mathbf{x}_n, \mathbf{x}_2) & \cdots & k(\mathbf{x}_n, \mathbf{x}_n)
\end{bmatrix}.
\end{align*}
Since this Gaussian construction is feasible for an arbitrary (finite) number of input values $n$, the \textsc{gp} effectively defines a probability distribution over functions. We denote this distribution as $f_{V_i,t} \sim \mathcal{GP}(m, k)$.

Suppose that we observe the function values $\hat{\mathbf{f}}\triangleq (f_{V_i,t}(\hat{\mathbf{x}}_1)+\epsilon_1, \hdots, f_{V_i,t}(\hat{\mathbf{x}}_M)+\epsilon_M)$ and the inputs $\hat{\mathbf{x}}\triangleq (\hat{\mathbf{x}}_1,\hdots,\hat{\mathbf{x}}_M)$, where each $\epsilon_i$ is an i.i.d. Gaussian noise term with variance $\sigma_{\epsilon}^2$. Let $\mathcal{D}_t=\{\hat{\mathbf{f}}, \hat{\mathbf{x}}\}$ denote the dataset of these samples. We can then construct the posterior Gaussian distribution $\mathbb{P}[\mathbf{f}_{*} \mid \mathcal{D}_t]=\mathcal{N}(\bm{\mu}_{\mathcal{D}_t}, \bm{\Sigma}_{\mathcal{D}_t})$ via the calculations
\begin{align*}
  &\bm{\mu}_{\mathcal{D}_t} = \mathbb{E}\{\mathbf{f}_{*} \mid \mathbf{x}_{*}, \hat{\mathbf{f}},\hat{\mathbf{x}}\}\\
  &\quad\text{ }\text{ }= m(\mathbf{x}_{*}) + K(\mathbf{x}_{*},\hat{\mathbf{x}})(K(\hat{\mathbf{x}}, \hat{\mathbf{x}}) + \sigma_{\epsilon}^2\mathbf{I}_{M})^{-1}(\hat{\mathbf{f}}-m(\hat{\mathbf{x}})),\\
&\bm{\Sigma}_{\mathcal{D}_t} = K(\mathbf{x}_{*}, \mathbf{x}_{*}) - K(\mathbf{x}_{*}, \hat{\mathbf{x}})(K(\hat{\mathbf{x}}, \hat{\mathbf{x}}) + \sigma_{\epsilon}^2\mathbf{I}_{M})^{-1}K(\hat{\mathbf{x}}, \mathbf{x}_{*}),
\end{align*}
where $\mathbf{I}_M$ is the $M \times M$ identity matrix and $m(\hat{\mathbf{x}})$ is a shorthand for $(m(\hat{\mathbf{x}}_1), \hdots, m(\hat{\mathbf{x}}_M))$ \cite{Rasmussen2006Gaussian}.

This posterior allows us to predict $f_{V_i,t}(\mathbf{x}_j)$ using the mean $(\bm{\mu}_{\mathcal{D}_t})_{j}$, whose uncertainty is quantified by the variance $(\bm{\Sigma}_{\mathcal{D}_t})_{jj}$. Since we can make such predictions for any input $\mathbf{x}_j$, the posterior is also a \textsc{gp}, which we denote as
\begin{align*}
f_{V_i,t} \mid \mathcal{D}_t \sim \mathcal{GP}(m_{\mid \mathcal{D}_t}, k_{\mid \mathcal{D}_t}).
\end{align*}
Here $m_{\mid \mathcal{D}_t}(\mathbf{x})$ and $k_{\mid\mathcal{D}_t}(\mathbf{x}, \mathbf{x}^{\prime})$ denote the posterior mean and covariance functions given the dataset $\mathcal{D}_t=\{\hat{\mathbf{f}}, \hat{\mathbf{x}}\}$, i.e.,
\begin{align*}
&m_{\mid \mathcal{D}_t}(\mathbf{x})\triangleq m(\mathbf{x}) + k(\mathbf{x}, \hat{\mathbf{x}})\left(K(\hat{\mathbf{x}}, \hat{\mathbf{x}}) + \sigma^2_{\epsilon}\mathbf{I}_M\right)^{-1}(\hat{\mathbf{f}}-m(\hat{\mathbf{x}})),\\
&k_{\mid\mathcal{D}_t}(\mathbf{x},\mathbf{x}^{\prime}) \triangleq k(\mathbf{x}, \mathbf{x}^{\prime}) - k(\mathbf{x}, \hat{\mathbf{x}})\left(K(\hat{\mathbf{x}}, \hat{\mathbf{x}}) + \sigma^2_{\epsilon}\mathbf{I}_M\right)^{-1}k(\hat{\mathbf{x}}, \mathbf{x}^{\prime}).
\end{align*}
Here, $k(\mathbf{x}, \hat{\mathbf{x}})$ is a shorthand for the vector $(k(\mathbf{x},\hat{\mathbf{x}}_1), \dots, k(\mathbf{x},\hat{\mathbf{x}}_M))$.
\bibliographystyle{IEEEtran}
\bibliography{references,url}
\end{document}